\RequirePackage{fix-cm}
\documentclass[twocolumn]{svjour3}          % twocolumn

\smartqed
\usepackage{graphicx}
\usepackage[utf8]{inputenc}
\usepackage[T1]{fontenc}
\usepackage{hyperref}
\usepackage{url}
\usepackage{booktabs}
\usepackage{nicefrac}
\usepackage{microtype}
\usepackage{latexsym}
\usepackage{soul}
\usepackage{caption}
\usepackage[list=true]{subcaption}
\usepackage{amsmath}
\usepackage{amssymb}
\usepackage[normalem]{ulem}
\usepackage{multirow}
\usepackage{wrapfig}
\usepackage{dsfont}
\usepackage[bottom]{footmisc}
\usepackage{colortbl}

\journalname{International Journal on Document Analysis and Recognition (IJDAR)}

\begin{document}

\title{Benchmarking Online Sequence-to-Sequence and Character-based Handwriting Recognition from IMU-Enhanced Pens}
\titlerunning{Benchmarking Online Handwriting Recognition from IMU-Enhanced Pens}

\author{Felix Ott \and
        David Rügamer \and
        Lucas Heublein \and
        Tim Hamann \and
        Jens~Barth \and
        Bernd Bischl \and
        Christopher Mutschler
}

\authorrunning{Ott et al.}

\institute{Felix Ott \at
              Fraunhofer IIS, Fraunhofer Institute for Integrated Circuits \\
              LMU Munich, Germany \\
              \email{felix.ott@iis.fraunhofer.de} \\
              orcid: \url{ https://orcid.org/0000-0002-4392-0830}
           \and
           David Rügamer \at
              LMU Munich, Germany \\
              RWTH Aachen, Germany \\
              \email{david.ruegamer@stat.uni-muenchen.de}
           \and
           Lucas Heublein \at
              Fraunhofer IIS, Fraunhofer Institute for Integrated Circuits \\
              \email{heublels@iis.fraunhofer.de}
           \and
           Tim Hamann \at
              STABILO International GmbH, Heroldsberg, Germany \\
              \email{tim.hamann@stabilo.com}
           \and
           Jens Barth \at
              STABILO International GmbH, Heroldsberg, Germany \\
              \email{jens.barth@stabilo.com}
           \and
           Bernd Bischl \at
              LMU Munich, Germany \\
              \email{bernd.bischl@stat.uni-muenchen.de}
           \and
           Christopher Mutschler \at
              Fraunhofer IIS, Fraunhofer Institute for Integrated Circuits \\
              \email{christopher.mutschler@iis.fraunhofer.de}
}

\date{Received: 31 March 2022 / Revised: 2 June 2022 / Accepted: 1 September 2022}

\maketitle

\begin{abstract}
\textbf{Purpose.} Handwriting is one of the most frequently occurring patterns in everyday life and with it come challenging applications such as handwriting recognition (HWR), writer identification, and signature verification. In contrast to offline HWR that only uses spatial information (i.e., images), online HWR (OnHWR) uses richer spatio-temporal information (i.e., trajectory data or inertial data). While there exist many offline HWR datasets, there is only little data available for the development of OnHWR methods on paper as it requires hardware-integrated pens. \textbf{Methods.} This paper presents data and benchmark models for real-time sequence-to-sequence (seq2seq) learning and single character-based recognition. Our data is recorded by a sensor-enhanced ballpoint pen, yielding sensor data streams from triaxial accelerometers, a gyroscope, a magnetometer and a force sensor at 100\,\textit{Hz}. We propose a variety of datasets including equations and words for both the writer-dependent and writer-independent tasks. Our datasets allow a comparison between classical OnHWR on tablets and on paper with sensor-enhanced pens. We provide an evaluation benchmark for seq2seq and single character-based HWR using recurrent and temporal convolutional networks and Transformers combined with a connectionist temporal classification (CTC) loss and cross-entropy (CE) losses. \textbf{Results.} Our convolutional network combined with BiLSTMs outperforms Transformer-based architectures, is on par with InceptionTime for sequence-based classification tasks, and yields better results compared to 28 state-of-the-art techniques. Time-series augmentation methods improve the sequence-based task, and we show that CE variants can improve the single classification task. \textbf{Conclusion.} Our implementations together with the large benchmark of state-of-the-art techniques of novel OnHWR datasets serve as a baseline for future research in the area of OnHWR on paper.
\end{abstract}


\section{Introduction}
\label{chap_introduction}

\newcommand\tabrotate[1]{\rotatebox{90}{#1\hspace{\tabcolsep}}}
\begin{table*}
\begin{center}
\setlength{\tabcolsep}{2.5pt}
    \caption{Overview of state-of-the-art trajectory-based and our inertial-based online handwriting datasets.}
    \label{table_datasets}
    \vspace{-0.1cm}
    \small \begin{tabular}{ p{0.5cm} | p{0.5cm} | p{0.5cm} | p{0.5cm} | p{0.5cm} | p{0.5cm} }
    & \multicolumn{1}{c|}{\textbf{Dataset}} & \multicolumn{1}{c|}{\textbf{Content}} & \multicolumn{1}{c|}{\textbf{Device}} & \multicolumn{1}{c|}{\textbf{Writers}} & \multicolumn{1}{c}{\textbf{Statistics}} \\ \hline
    & \multicolumn{1}{l|}{Kuchibue~\cite{nakagawa,nakagawa_matsumoto}} & \multicolumn{1}{l|}{Japanese characters} & \multicolumn{1}{l|}{Tablet} & \multicolumn{1}{r|}{120} & \multicolumn{1}{l}{$10,154 \times 120$ char. patterns} \\
    \multirow{3}{*}{\tabrotate{Characters}} & \multicolumn{1}{l|}{MRG-OHTC~\cite{ma_liu}} & \multicolumn{1}{l|}{Tibetan characters} & \multicolumn{1}{l|}{Tablet} & \multicolumn{1}{r|}{130} & \multicolumn{1}{l}{910 character classes} \\
    & \multicolumn{1}{l|}{CASIA~\cite{wang_liu}} & \multicolumn{1}{l|}{Chinese characters} & \multicolumn{1}{l|}{Anoto pen on paper} & \multicolumn{1}{r|}{1,020} & \multicolumn{1}{l}{3.5m characters} \\
    & \multicolumn{1}{l|}{OnHW-chars~\cite{ott}} & \multicolumn{1}{l|}{English characters} & \multicolumn{1}{l|}{Sensor pen} & \multicolumn{1}{r|}{119} & \multicolumn{1}{l}{31,275 characters, 52 classes} \\
    & \multicolumn{1}{l|}{UNIPEN~\cite{guyon_schomaker}} & \multicolumn{1}{l|}{Sentences, words, characters} & \multicolumn{1}{l|}{Pen-based computer} & \multicolumn{1}{c|}{-} & \multicolumn{1}{l}{>12,000 chars. per writer} \\
    & \multicolumn{1}{l|}{CROHME~\cite{mouchere_gaudin}} & \multicolumn{1}{l|}{Mathematical expressions} & \multicolumn{1}{l|}{White-board, tablet} & \multicolumn{1}{c|}{>100} & \multicolumn{1}{l}{9,507 expressions} \\ \hline
    & \multicolumn{1}{l|}{IRONOFF \cite{viard_gaudin}} & \multicolumn{1}{l|}{French words, chars., digits} & \multicolumn{1}{l|}{Trajectory, images} & \multicolumn{1}{c|}{-} & \multicolumn{1}{l}{50,000 words, 32,000 chars.} \\
    \multirow{5}{*}{\tabrotate{Sequence}} & \multicolumn{1}{l|}{ICROW~\cite{icrow}} & \multicolumn{1}{l|}{Dutch, Irish, Italian words} & \multicolumn{1}{c|}{-} & \multicolumn{1}{r|}{67} & \multicolumn{1}{l}{13,119 words} \\
    & \multicolumn{1}{l|}{IAM-OnDB~\cite{liwicki}} & \multicolumn{1}{l|}{English sentences} & \multicolumn{1}{l|}{Whiteboard} & \multicolumn{1}{r|}{197} & \multicolumn{1}{l}{82,272 words} \\
    & \multicolumn{1}{l|}{LMCA~\cite{kherallah}} & \multicolumn{1}{l|}{Arabic words, chars., digits} & \multicolumn{1}{l|}{Tablet} & \multicolumn{1}{r|}{55} & \multicolumn{1}{l}{30k digits, 100k chars., 500 w.} \\
    & \multicolumn{1}{l|}{ADAB~\cite{abed}} & \multicolumn{1}{l|}{Arabic words} & \multicolumn{1}{l|}{Tablet} & \multicolumn{1}{r|}{170} & \multicolumn{1}{l}{20,000+ words} \\
    & \multicolumn{1}{l|}{IBM\_UB\_1~\cite{shivram}} & \multicolumn{1}{l|}{English words} & \multicolumn{1}{l|}{Notepad} & \multicolumn{1}{r|}{43} & \multicolumn{1}{l}{6,654 pages} \\
    & \multicolumn{1}{l|}{VNOnDB~\cite{nguyen_nguyen,nguyen}} & \multicolumn{1}{l|}{Vietnamese words, lines, paragr.} & \multicolumn{1}{l|}{Tablet} & \multicolumn{1}{r|}{200} & \multicolumn{1}{l}{110,746 words} \\ \hline
    & \multicolumn{1}{l|}{OnHW-equations} & \multicolumn{1}{l|}{Equations written on paper} & \multicolumn{1}{l|}{Sensor pen} & \multicolumn{1}{r|}{55} & \multicolumn{1}{l}{10,720 equations, 15 classes} \\
    \multirow{1}{*}{\tabrotate{\textbf{Ours}}} & \multicolumn{1}{l|}{OnHW-words500} & \multicolumn{1}{l|}{Repeated 500 words on paper} & \multicolumn{1}{l|}{Sensor pen} & \multicolumn{1}{r|}{53} & \multicolumn{1}{l}{25,218 words, 59 classes} \\
    & \multicolumn{1}{l|}{OnHW-wordsRandom} & \multicolumn{1}{l|}{Random words written on paper} & \multicolumn{1}{l|}{Sensor pen} & \multicolumn{1}{r|}{54} & \multicolumn{1}{l}{14,645 words, 59 classes} \\
    & \multicolumn{1}{l|}{OnHW-wordsTraj} & \multicolumn{1}{l|}{Words written on a tablet} & \multicolumn{1}{l|}{Sensor pen on tablet} & \multicolumn{1}{r|}{2} & \multicolumn{1}{l}{16,752 words, 52 classes} \\
    & \multicolumn{1}{l|}{OnHW-symbols} & \multicolumn{1}{l|}{Numbers, symbols on paper} & \multicolumn{1}{l|}{Sensor pen} & \multicolumn{1}{r|}{27} & \multicolumn{1}{l}{2,326 characters, 15 classes} \\
    \end{tabular}
    \vspace{-0.4cm}
\end{center}
\end{table*}

Handwriting provides language information based on structured symbols and is used for communication or documentation of speech. HWR refers to the digitalization of written text and can be categorized into offline and online HWR. Research on offline HWR systems is very advanced and has almost reached a human-level performance, but cannot be applied for real-time recognition applications (as they induce an unacceptable delay) \cite{fahmy} as the written text has first to be digitalized. Optical character recognition (OCR), one of the dominant approaches in offline HWR, deals with the analysis of the visual representation of handwriting only. Its application and accuracy are limited as it cannot make use of temporal information such as writing direction and speed, or the pressure applied to the paper \cite{ott,plamondon}.

In contrast, online HWR typically works on different types of spatio-temporal signals such as the positions of the pen tip (in 2D), its temporal context or the movement on the writing surface. These handwriting signals can then, e.g., be partitioned into (indexed) strokes \cite{vinciarelli}. Compared to offline HWR, OnHWR has its own merits, e.g., the difficult segmentation of cursive written sequences into single characters. Many highly relevant handwriting problems in everyday life require both an informative representation of the writing and well-working classification algorithms \cite{hussain}. Examples include the verification of signatures, the indentification of writers, or the recognition of handwriting. 

The representation of written text crucially depends on the way it has been recorded. Many recording systems make use of a stylus pen (a touch pen with a sensible magnetic mesh tip) together with a touch screen surface \cite{alimoglu}. Systems for writing on paper are only prototypical, such as the ones used in \cite{bu_xie,schrapel,wang_hsu} or the GyroPen \cite{deselaers} that provides a pen-like interaction from standard built-in sensors in modern smartphones. An advanced system for recording online HWR data was proposed by Ott et al.~\cite{ott} who use a sensor-enhanced ballpoint pen that is extended with inertial measurement units (IMUs). The hand movement and velocity patterns with such a pen are different to air-writing \cite{yana_onoye}. In this paper, we propose a novel dataset collection of equations and words recorded with an IMU-enhanced pen. Using this pen allows an online representation of the accelerations, orientations and the pressure applied to the pen. Writing styles can thereby be characterized by an information-rich multivariate time-series (MTS). These datasets lay the foundation for HWR from pens with integrated sensors \cite{deselaers,klass_lorenz,ott_ijcai,ott_wacv,ott,schrapel,wang_hsu,wehbi}, a so far unsolved challenge in machine learning. %cite: xu_xue_zhang, xu_xue2,xu_xue,chen_alregib

For machine learning tasks derived from online handwriting data, we distinguish between single label prediction tasks (i.e., classifying characters, digits and symbols) and tasks to predict sequences of labels (i.e., words, sentences and equations). We here focus on the online seq2seq prediction task for writer-dependent (WD) and writer-independent (WI) classification, but also consider the single label classification task. Seq2seq models in natural language processing (NLP) and speech recognition \cite{synnaeve} are used to convert sequences of \textsc{Type A} to sequences of \textsc{Type B} (e.g., sentences from English to German). Many real-world datasets take the form of sequences, e.g., written texts, numbers, audio or video frame sequences. While many approaches build on language models or lexica \cite{bluche3,quiniou,seni,synnaeve} that outperform model-free approaches for certain datasets (e.g., sentences), these approaches require additional efforts to properly deal with the data at hand. They cannot handle dialects and informal words off-the-shelf, do not recognize wrongly written words, and require a large corpus volume with large training times to achieve an acceptable accuracy \cite{kaity}. Even with additional pre-processing, language models and lexica cannot (or only with high effort \cite{zhang_du}) be applied to certain types of sequences, e.g., equations, as in our case. For our benchmark baselines we therefore resort to language- and lexicon-free approaches without token passing. More specifically, we provide an evaluation benchmark with CNNs combined with (bidirectional) LSTMs and TCNs, and an attention-based model for the seq2seq OnHWR, as well as several Transformers for the single character-based classification task. 

The remainder of the paper is organized as follows. We discuss related work in Section~\ref{chap_related_work}. Section~\ref{chap_evaluation} presents our novel collection of online handwriting datasets on sequence level. Section~\ref{chap_method} introduces the suggested benchmark models, in particular, we propose several CNN architectures. In Section~\ref{chap_results} we provide experimental results before we end with a conclusion in Section~\ref{chap_summary}.
\section{Background \& Related Work}
\label{chap_related_work}

We will first provide an overview of available datasets of online handwriting datasets and explain the particularities for each one. Next, we discuss related methodological approaches to model such data. For a detailed overview of text classification methods we refer to \cite{kowsari,li_peng}.

\subsection{Datasets}
\label{sec_rw_datasets}

While there are many offline datasets, online data is rare \cite{hussain}. To properly evaluate OnHWR methods, we need a multi-label online dataset that allows for the evaluation of tasks for both the writer-dependent and the writer-independent case. Table~\ref{table_datasets} gives an overview of available online datasets. For the single character prediction task, the Kuchibue~\cite{nakagawa,nakagawa_matsumoto}, MRG-OHTC~\cite{ma_liu}, CASIA~\cite{wang_liu}, and OnHW-chars~\cite{ott} datasets are available. While the OnHW-chars dataset is rather small, we provide single character-based datasets from a larger database. For our sequence-based method (i.e., a technique that predicts a whole sequence of characters), the IRONOFF~\cite{viard_gaudin}, ICROW~\cite{icrow}, IAM-OnDB~\cite{liwicki}, LMCA~\cite{kherallah}, ADAB~\cite{abed}, IBM-UB~\cite{shivram} and VNOnDB~\cite{nguyen,nguyen_nguyen} word and sentence datasets can be used.

The commonly used IAM-OnDB~\cite{liwicki} and VNOnDB \cite{nguyen} datasets only include online trajectory data written on a tablet. However, writing on even and smooth surfaces influences the writing style of the user \cite{gerth}. To circumvent this disadvantage, we initially recorded a small character-only dataset with a sensor-enhanced pen on usual paper in previous work \cite{ott}. In this paper we make use of this novel pen and record sequence-based samples for a comparison and evaluation benchmark with the trajectory-based IAM-OnDB (line level) and VNOnDB-words datasets. Hence, our datasets allow a broad research on sequence-based classification from sensor-enhanced pens, and allows the connection between classical OnHW recognition on tablets with sensor-enhanced pens.

\subsection{Methods}
\label{sec_rw_methods}

While hidden markov models (HMMs) \cite{almaza,bertolami,doetsch,dreuw2,espana} have initially been applied to offline HWR, more recently, deep learning models became predominant, including convolutional neural networks (CNNs) \cite{yousef}, temporal convolutional networks (TCNs) \cite{sharma,sharma2}, recurrent neural networks (RNNs) \cite{dutta,graves,pham,puigcerver,sudholt,wigington} including long short-term memorys (LSTMs), bidirectional LSTMs (BiLSTMs) \cite{carbune,tian}, and multidimensional RNNs \cite{graves3,voigtlaender}. More recently, attention models further improved the classification accuracy of RNNs \cite{bluche}, but did not outperform previous approaches for OnHWR. Despite Transformers \cite{vaswani} and its variants \cite{choromanski,jaegle,kang,kitaev,tay_bahri,wang_li} got very popular for NLP \cite{rijhwani} and image processing, these have so far only been applied to offline HWR \cite{kang}. The Transformer by \cite{peng_xie} is based on a language model and is used for Chinese text recognition. Similarly, variational autoencoders (VAEs), RNNs \cite{graves_gen} and generative adversarial networks (GANs) \cite{fogel} have been successfully applied for synthetic offline handwriting generation, but not for the online case so far. For the time-series classification task, standard convolutional architectures \cite{Fawaz,he_zhang,rahimian,wang_2016,zou}, spatio-temporal methods \cite{bai,chung,elsayed,karim,mlstm_fcn} and variants \cite{fauvel,dempster,tang_long,wang_2018} as well as Transformers \cite{zerveas} have been employed. In \cite{ott}, we evaluated machine learning techniques, while in this paper we provide a broad evaluation benchmark on classical and novel time-series classification methods previously mentioned. While many approaches predict one class after the other, \cite{chowdhury,michael} predicted sequences similar to our approach. This is necessary to construct a suitable loss function described in the following. See Appendix~\ref{sec_appendix_related_work} for a more detailed overview of related work.

\begin{figure*}[t!]
	\centering
	\begin{minipage}[h]{0.195\linewidth}
        \centering
    	\includegraphics[trim=10 10 10 10, clip, width=1.0\linewidth]{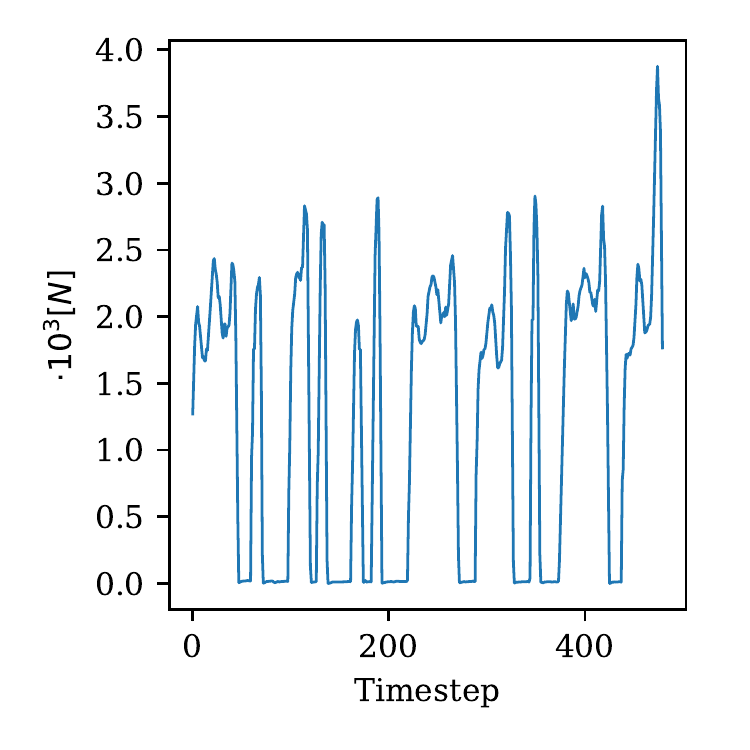}
    	\vspace{-0.1cm}
    	\subcaption{Force.}
    	\label{figure_pen_signal_1}
    \end{minipage}
    \hfill
	\begin{minipage}[h]{0.195\linewidth}
        \centering
    	\includegraphics[trim=10 10 10 10, clip, width=1.0\linewidth]{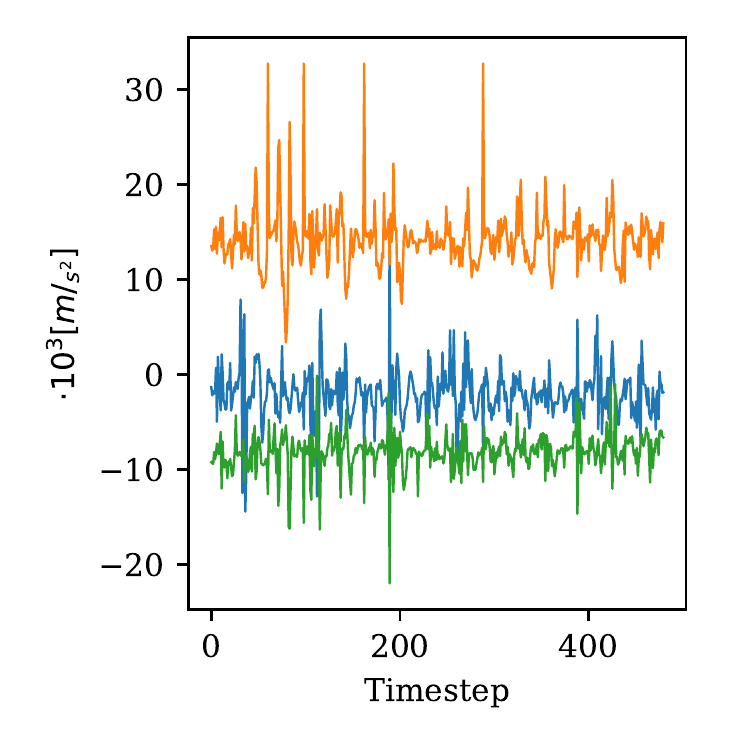}
    	\vspace{-0.1cm}
    	\subcaption{Front acc.}
    	\label{figure_pen_signal_2}
    \end{minipage}
    \hfill
	\begin{minipage}[h]{0.195\linewidth}
        \centering
    	\includegraphics[trim=10 10 10 10, clip, width=1.0\linewidth]{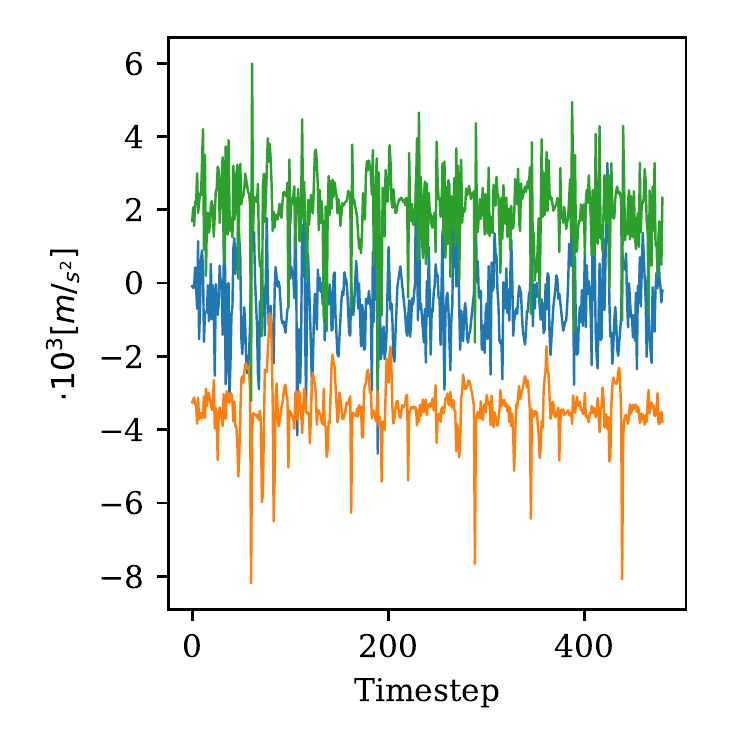}
    	\vspace{-0.1cm}
    	\subcaption{Rear acc.}
    	\label{figure_pen_signal_3}
    \end{minipage}
    \hfill
	\begin{minipage}[h]{0.195\linewidth}
        \centering
    	\includegraphics[trim=10 10 10 10, clip, width=1.0\linewidth]{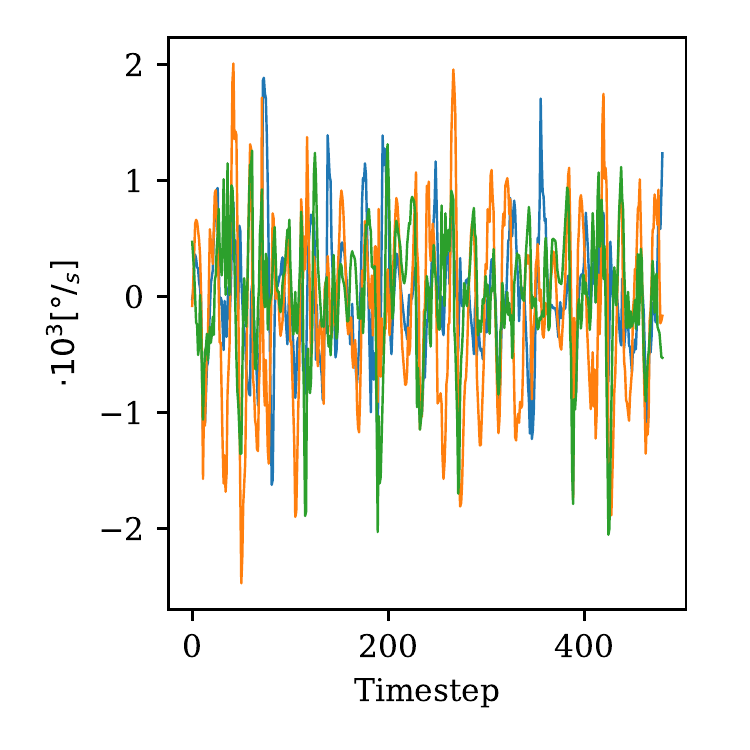}
    	\vspace{-0.1cm}
    	\subcaption{Gyroscope.}
    	\label{figure_pen_signal_4}
    \end{minipage}
    \hfill
	\begin{minipage}[h]{0.195\linewidth}
        \centering
    	\includegraphics[trim=10 10 10 10, clip, width=1.0\linewidth]{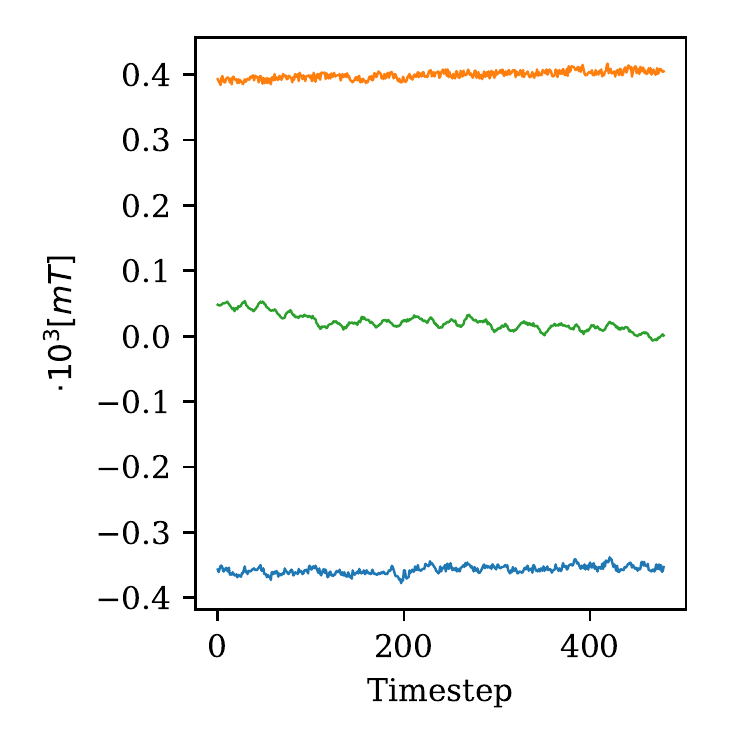}
    	\vspace{-0.1cm}
    	\subcaption{Magnetometer.}
    	\label{figure_pen_signal_5}
    \end{minipage}
    \caption{Exemplary sensor data for the x-, y- and z-axis of the equation: 20583:70.}
    \label{figure_exemplary_signal}
\end{figure*}

\paragraph{Loss Functions.} For sequence prediction the connectionist temporal classification (CTC) \cite{graves2,graves,kim_hori} loss combined with beam search \cite{scheidl} has extensively been used. The Edit distance (ED) \cite{levenshtein} quantifies how dissimilar two strings are to one another by counting the minimum number of operations required to transform one string into the other. The ED allows deletion, insertion and substitution. However, the ED is a discrete function that is known to be hard to optimize. Ofitserov et al.~\cite{ofitserov} proposed a soft ED, which is a smooth approximation of ED that is differentiable. Seni et al.~\cite{seni2} used the ED for HWR. We use the CTC loss for sequence prediction (see Section~\ref{chap_method}).

\section{Datasets and Evaluation Methodology}
\label{chap_evaluation}

\begin{figure}[t!]
	\centering
	\includegraphics[width=1.0\linewidth]{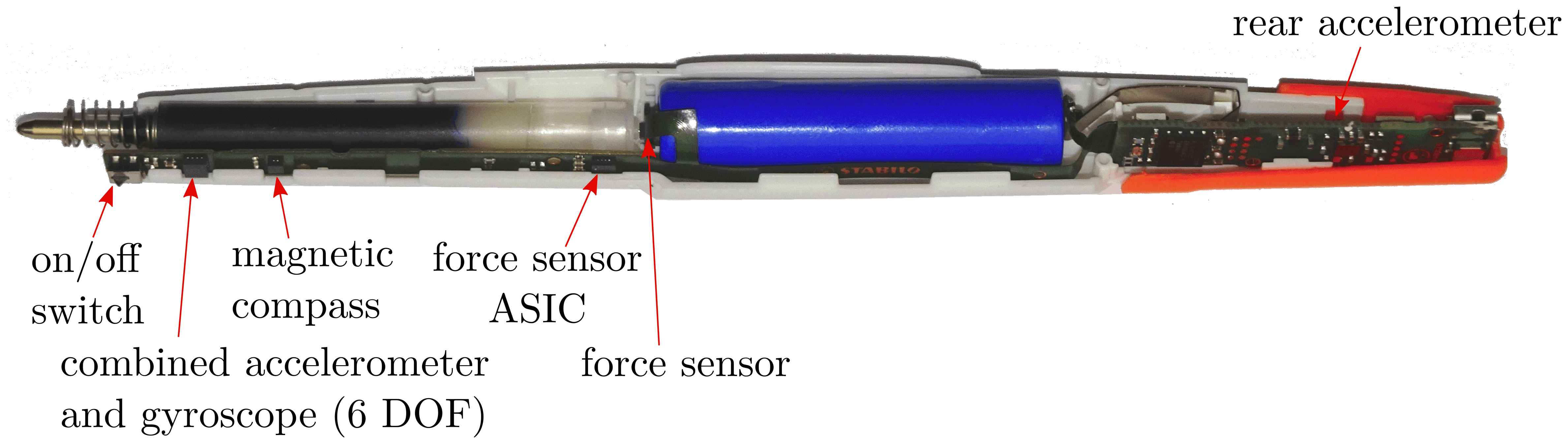}
    \caption{Sensor-enhanced pen.}
    \label{figure_digipen}
\end{figure}

\begin{table*}
\begin{center}
\setlength{\tabcolsep}{3.4pt}
    \caption{Overview of our recordings from right-handed writers and state-of-the-art online handwriting datasets for writer-dependent (WD) and writer-independent (WI) tasks.}
    \label{table_datasets2}
    \small \begin{tabular}{ p{0.5cm} | p{0.5cm} | p{0.5cm} | p{0.5cm} | p{0.5cm} | p{0.5cm} | p{0.5cm} | p{0.5cm} | p{0.5cm} | p{0.5cm} }
    \multicolumn{1}{c|}{} & \multicolumn{1}{c|}{\textbf{Number}} & \multicolumn{1}{c|}{\textbf{Number}} & \multicolumn{1}{c|}{\textbf{Maximal}} & \multicolumn{5}{c|}{\textbf{Number Samples}} & \multicolumn{1}{c}{\textbf{Total}} \\
    \multicolumn{1}{c|}{\textbf{Dataset}} & \multicolumn{1}{c|}{\textbf{Writers}} & \multicolumn{1}{c|}{\textbf{Classes}} & \multicolumn{1}{c|}{\textbf{Length}}  & \multicolumn{1}{c|}{\textbf{Total}} & \multicolumn{2}{c|}{\textbf{WD}} & \multicolumn{2}{c|}{\textbf{WI}} & \multicolumn{1}{c}{\textbf{Chars.}} \\ \hline
    \multicolumn{1}{l|}{\textbf{OnHW-equations}} & \multicolumn{1}{r|}{55} & \multicolumn{1}{r|}{15} & \multicolumn{1}{r|}{15} & \multicolumn{1}{r|}{10,713} & \multicolumn{1}{r}{8,595} & \multicolumn{1}{r|}{2,118} & \multicolumn{1}{r}{8,610} & \multicolumn{1}{r|}{2,103} & \multicolumn{1}{r}{106,968} \\
    \multicolumn{1}{l|}{\textbf{OnHW-words500(R)}} & \multicolumn{1}{r|}{53} & \multicolumn{1}{r|}{59} & \multicolumn{1}{r|}{19} & \multicolumn{1}{r|}{25,218} & \multicolumn{1}{r}{20,176} & \multicolumn{1}{r|}{5,042} & \multicolumn{1}{r}{19,918} & \multicolumn{1}{r|}{5,300} & \multicolumn{1}{r}{137,219} \\
    \multicolumn{1}{l|}{\textbf{OnHW-wordsRandom}} & \multicolumn{1}{r|}{54} & \multicolumn{1}{r|}{59} & \multicolumn{1}{r|}{27} & \multicolumn{1}{r|}{14,641} & \multicolumn{1}{r}{11,744} & \multicolumn{1}{r|}{2,897} & \multicolumn{1}{r}{11,716} & \multicolumn{1}{r|}{2,925} & \multicolumn{1}{r}{146,350} \\
    \multicolumn{1}{l|}{\textbf{OnHW-wordsTraj}} & \multicolumn{1}{r|}{2} & \multicolumn{1}{r|}{59} & \multicolumn{1}{r|}{10} & \multicolumn{1}{r|}{16,752} & \multicolumn{1}{r}{13,250} & \multicolumn{1}{r|}{3,502} & \multicolumn{1}{c}{-} & \multicolumn{1}{c|}{-} & \multicolumn{1}{r}{146,512} \\
    \multicolumn{1}{l|}{\textbf{OnHW-symbols}} & \multicolumn{1}{r|}{27} & \multicolumn{1}{r|}{15} & \multicolumn{1}{c|}{single} & \multicolumn{1}{r|}{2,326} & \multicolumn{1}{r}{1,853} & \multicolumn{1}{r|}{473} & \multicolumn{1}{r}{1,715} & \multicolumn{1}{r|}{611} & \multicolumn{1}{r}{2,326} \\ \hline
    \multicolumn{1}{l|}{\textbf{ICROW}~\cite{icrow}} & \multicolumn{1}{r|}{67} & \multicolumn{1}{r|}{53} & \multicolumn{1}{r|}{15} & \multicolumn{1}{r|}{13,119} & \multicolumn{1}{r}{10,500} & \multicolumn{1}{r|}{2,619} & \multicolumn{1}{r}{10,524} & \multicolumn{1}{r|}{2,595} & \multicolumn{1}{r}{90,138} \\
    \multicolumn{1}{l|}{\textbf{IAM-OnDB}~\cite{liwicki}} & \multicolumn{1}{r|}{197} & \multicolumn{1}{r|}{81} & \multicolumn{1}{r|}{64} & \multicolumn{1}{r|}{10,773} & \multicolumn{1}{r}{8,702} & \multicolumn{1}{r|}{2,071} & \multicolumn{1}{r}{8,624} & \multicolumn{1}{r|}{2,149} & \multicolumn{1}{r}{265,477} \\
    \multicolumn{1}{l|}{\textbf{VNOnDB-words}~\cite{nguyen}} & \multicolumn{1}{r|}{201} & \multicolumn{1}{r|}{147} & \multicolumn{1}{r|}{11} & \multicolumn{1}{r|}{110,746} & \multicolumn{1}{r}{88,677} & \multicolumn{1}{r|}{22,069} & \multicolumn{1}{r}{88,486} & \multicolumn{1}{r|}{22,260} & \multicolumn{1}{r}{368,455} \\
    \multicolumn{1}{l|}{\textbf{OnHW-chars}~\cite{ott}} & \multicolumn{1}{r|}{119} & \multicolumn{1}{r|}{52} & \multicolumn{1}{c|}{single} & \multicolumn{1}{r|}{31,275} & \multicolumn{1}{r}{23,059} & \multicolumn{1}{r|}{8,216} & \multicolumn{1}{r}{23,059} & \multicolumn{1}{r|}{8,216} & \multicolumn{1}{r}{31,275} \\
    \end{tabular}
\end{center}
\end{table*}

Our datasets are a collection of existing and newly generated online handwriting recordings. Section~\ref{chap_record_setup} first describes our recording setup to create novel and information-rich datasets. The following Section~\ref{chap_eval_data} gives details about the properties of our different OnHW datasets and compares them to existing datasets. Section~\ref{chap_eval_metrics} proposes a set of evaluation metrics.

\subsection{Recording Setup}
\label{chap_record_setup}

Our datasets are recorded with a sensor-enhanced pen developed by STABILO International GmbH that contains two accelerometers at the front and the back (3 axes each), one gyroscope (3 axes), one magnetometer (3 axes), and one force sensor at 100\,\textit{Hz} (see Figure~\ref{figure_digipen}). The data recordings contain 14 measurements provided by the sensors: four sensor data signals (each in x, y and z direction), the force with which the pen tip touches the surface, and the timestep at which the tablet receives the data from the pen. Figure~\ref{figure_exemplary_signal} shows an exemplary sensor signal from a written equation. Using the force sensor the sensor data allows to separate strokes well as the writer lifts the pen between every character (this is not possible for cursive writing, e.g., for words). In total, we let 131 adult writers participate in our data collection. For more information on the sensor pen and data acquisition, see Appendices~\ref{sec_appendix_pen} and~\ref{sec_appendix_data_format}.

\subsection{Datasets}
\label{chap_eval_data}

\begin{table*}
\begin{center}
\setlength{\tabcolsep}{3.4pt}
    \caption{Overview of our datasets from left-handed writers for writer-dependent (WD) and writer-independent (WI) tasks. For WD tasks a 80/20 train/validation split is used, for WI a dataset-specific split is used.}
    \label{table_datasets2_left}
    \small \begin{tabular}{ p{0.5cm} | p{0.5cm} | p{0.5cm} | p{0.5cm} | p{0.5cm} | p{0.5cm} | p{0.5cm} | p{0.5cm} | p{0.5cm} }
    \multicolumn{1}{c|}{} & \multicolumn{1}{c|}{\textbf{Number}} & \multicolumn{1}{c|}{\textbf{Maximal}} & \multicolumn{5}{c|}{\textbf{Number Samples}} & \multicolumn{1}{c}{\textbf{Total}} \\
    \multicolumn{1}{c|}{\textbf{Dataset}} & \multicolumn{1}{c|}{\textbf{Writers}} & \multicolumn{1}{c|}{\textbf{Length}} & \multicolumn{1}{c|}{\textbf{Total}} & \multicolumn{2}{c|}{\textbf{WD}} & \multicolumn{2}{c|}{\textbf{WI}} & \multicolumn{1}{c}{\textbf{Chars.}} \\ \hline
    \multicolumn{1}{l|}{\textbf{OnHW-equations-L}} & \multicolumn{1}{r|}{4} & \multicolumn{1}{r|}{15} & \multicolumn{1}{r|}{843} & \multicolumn{1}{r}{677} & \multicolumn{1}{r|}{166} & \multicolumn{1}{r}{543} & \multicolumn{1}{r|}{300} & \multicolumn{1}{r}{8,438} \\
    \multicolumn{1}{l|}{\textbf{OnHW-words500-L}} & \multicolumn{1}{r|}{2} & \multicolumn{1}{r|}{19} & \multicolumn{1}{r|}{1,000} & \multicolumn{1}{r}{800} & \multicolumn{1}{r|}{200} & \multicolumn{1}{r}{500} & \multicolumn{1}{r|}{500} & \multicolumn{1}{r}{5,438} \\
    \multicolumn{1}{l|}{\textbf{OnHW-wordsRandom-L}} & \multicolumn{1}{r|}{2} & \multicolumn{1}{r|}{26} & \multicolumn{1}{r|}{996} & \multicolumn{1}{r}{798} & \multicolumn{1}{r|}{198} & \multicolumn{1}{r}{497} & \multicolumn{1}{r|}{499} & \multicolumn{1}{r}{10,029} \\
    \multicolumn{1}{l|}{\textbf{OnHW-symbols-L}} & \multicolumn{1}{r|}{4} & \multicolumn{1}{c|}{single} & \multicolumn{1}{r|}{361} & \multicolumn{1}{r}{289} & \multicolumn{1}{r|}{72} & \multicolumn{1}{r}{271} & \multicolumn{1}{r|}{90} & \multicolumn{1}{r}{361} \\ \hline
    \multicolumn{1}{l|}{\textbf{OnHW-chars-L}~\cite{ott}} & \multicolumn{1}{r|}{9} & \multicolumn{1}{c|}{single} & \multicolumn{1}{r|}{2,270} & \multicolumn{1}{r}{1,816} & \multicolumn{1}{r|}{454} & \multicolumn{1}{c}{-} & \multicolumn{1}{c|}{-} & \multicolumn{1}{r}{2,270} \\
    \end{tabular}
\end{center}
\end{table*}

\begin{figure*}[t!]
	\centering
    \includegraphics[width=0.7\linewidth]{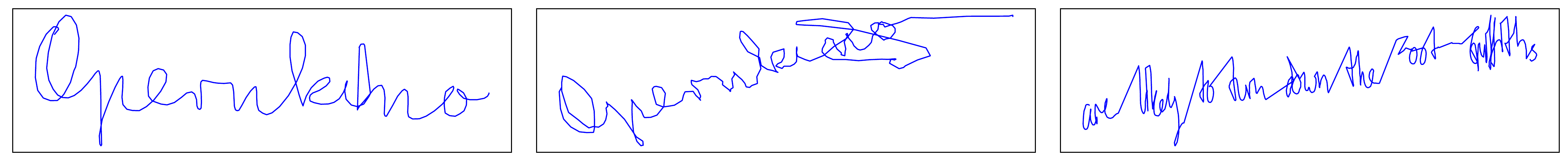}
    \caption{Exemplary online samples of our \textit{OnHW-wordsTraj} dataset (left: tablet, middle: camera) and the IAM-OnDB~\cite{liwicki} (line level) dataset (right).}
    \label{figure_exemplary_online}
\end{figure*}

We propose a large set of four different sequence-based datasets (see the first four entries in Table~\ref{table_datasets2}): The \textbf{OnHW-equations} dataset was part of the \textit{UbiComp 2021 challenge}\footnote{UbiComp challenge: \url{https://stabilodigital.com/ubicomp-2021-challenge/}} and is written by 55 writers and consists of 10 number classes and 5 operator classes (\texttt{+}, \texttt{-}, $\cdot$, \texttt{:}, \texttt{=}). The dataset consists of a total of 10,713 samples. While in the \textbf{OnHW-words500} dataset only the same 500 words per each writer appear, in the \textbf{OnHW-wordsRandom} dataset every sample is randomly chosen from a large German and English word list. This allows the comparison of indirectly learning a lexicon of 500 words or, alternatively, completely lexicon-free learning. The OnHW-wordsRandom dataset (14,641 samples) is smaller than the OnHW-words500 dataset (25,218 samples), but contains longer words with a maximal length of 27 labels (19 labels for OnHW-words500). The train/validation-split for the OnHW-words500 dataset is based on words for the WD task such that the same 400 words per writer are in the train set and the same 100 words per writer are in the validation set. For the WI task, the split is done by writer such that all 500 words of a single writer are either in the train or validation set. As it is more likely to overfit on the same words, the WD task of OnHW-words500 is more challenging compared to the OnHW-wordsRandom dataset. The \textbf{OnHW-words500R} dataset is a random split of OnHW-words500.

Additionally, we record the \textbf{OnHW-wordsTraj} dataset that consists of four different data sources. We replace the ink refill with a Wacom EMR module and record online trajectories at 30\,\textit{Hz} on a Samsung Galaxy Tab S4 tablet along with the sensor data. Four cameras pointed on the pen to record the movement of the pen tip at 60\,\textit{Hz}. We manually label the pixels of 100 random images of the recorded videos in the classes ``pen``, ``pen tip`` and ``background``, and train U-Net~\cite{unet} to predict the pen tip pixels from all images. From this we derive the pen tip trajectory in camera coordinates. Two persons wrote 4,257 words in total that results in 16,752 camera samples. With this dataset it is possible to compare results from traditional online trajectory datasets (written on a tablet) with our online sensor pen datasets. Figure~\ref{figure_exemplary_online} exemplarily compares the trajectory and camera data of the OnHW-wordsTraj dataset with the IAM-OnDB~\cite{liwicki} dataset. Table~\ref{table_datasets2_left} gives a dataset overview of left-handed writers. Sample sizes are smaller and range between around 3\% to 13.4\% of the sample sizes of right-handed datasets. For our benchmark, we consider right- and left-handed writers separately, and will publish right- as well as left-handed datasets for future research.\footnote{\href{https://www.iis.fraunhofer.de/de/ff/lv/dataanalytics/anwproj/schreibtrainer/onhw-dataset.html}{www.iis.fraunhofer.de/de/ff/lv/dataanalytics/anwproj/ schreibtrainer/onhw-dataset.html}}

\begin{figure*}[t!]
	\centering
	\begin{minipage}[h]{0.32\linewidth}
        \centering
    	\includegraphics[width=1.0\linewidth]{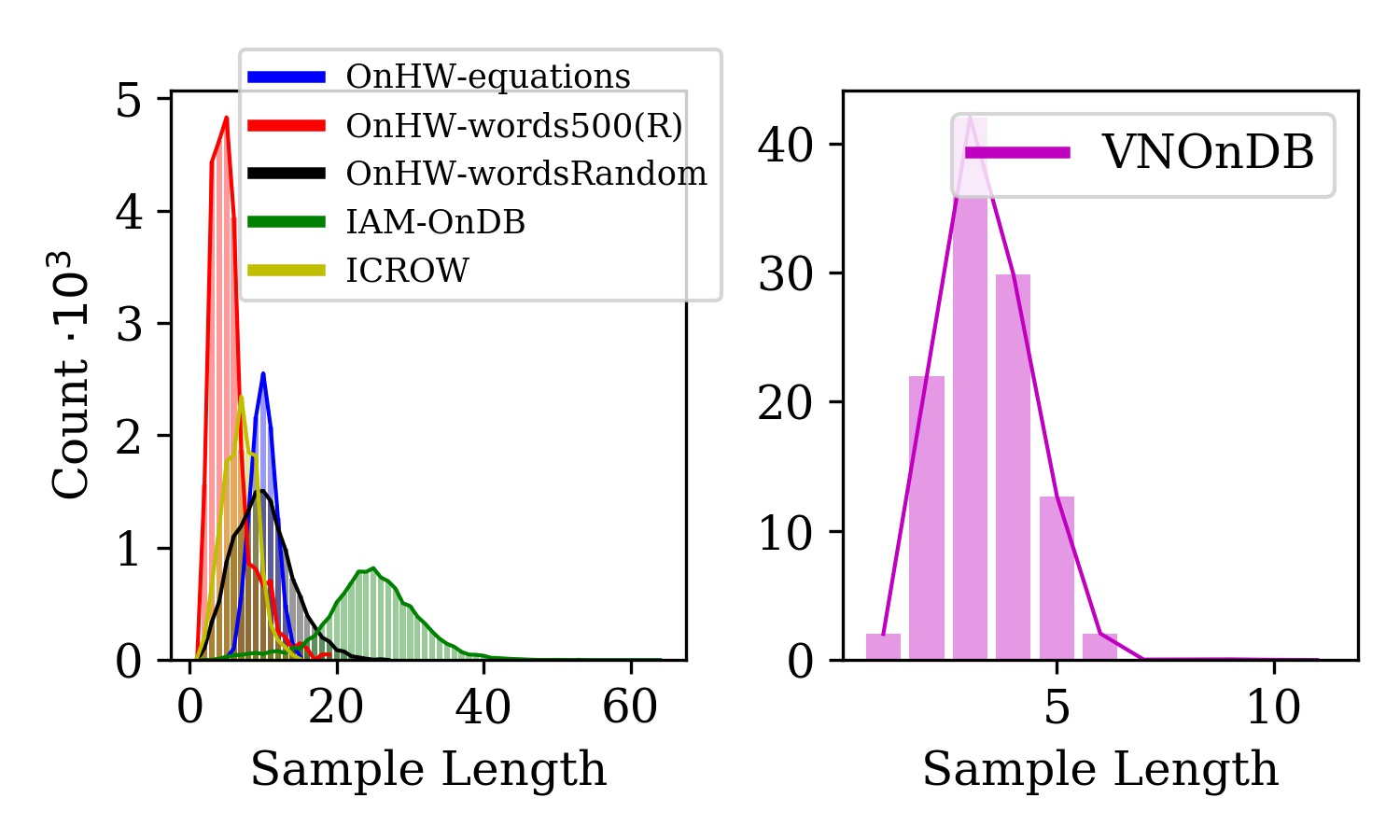}
        \vspace{-0.2cm}
    	\subcaption{Distribution of sample lengths for all datasets.}
    	\label{figure_statistics_dataset1}
    \end{minipage}
    \hfill
	\begin{minipage}[h]{0.32\linewidth}
        \centering
    	\includegraphics[width=1.0\linewidth]{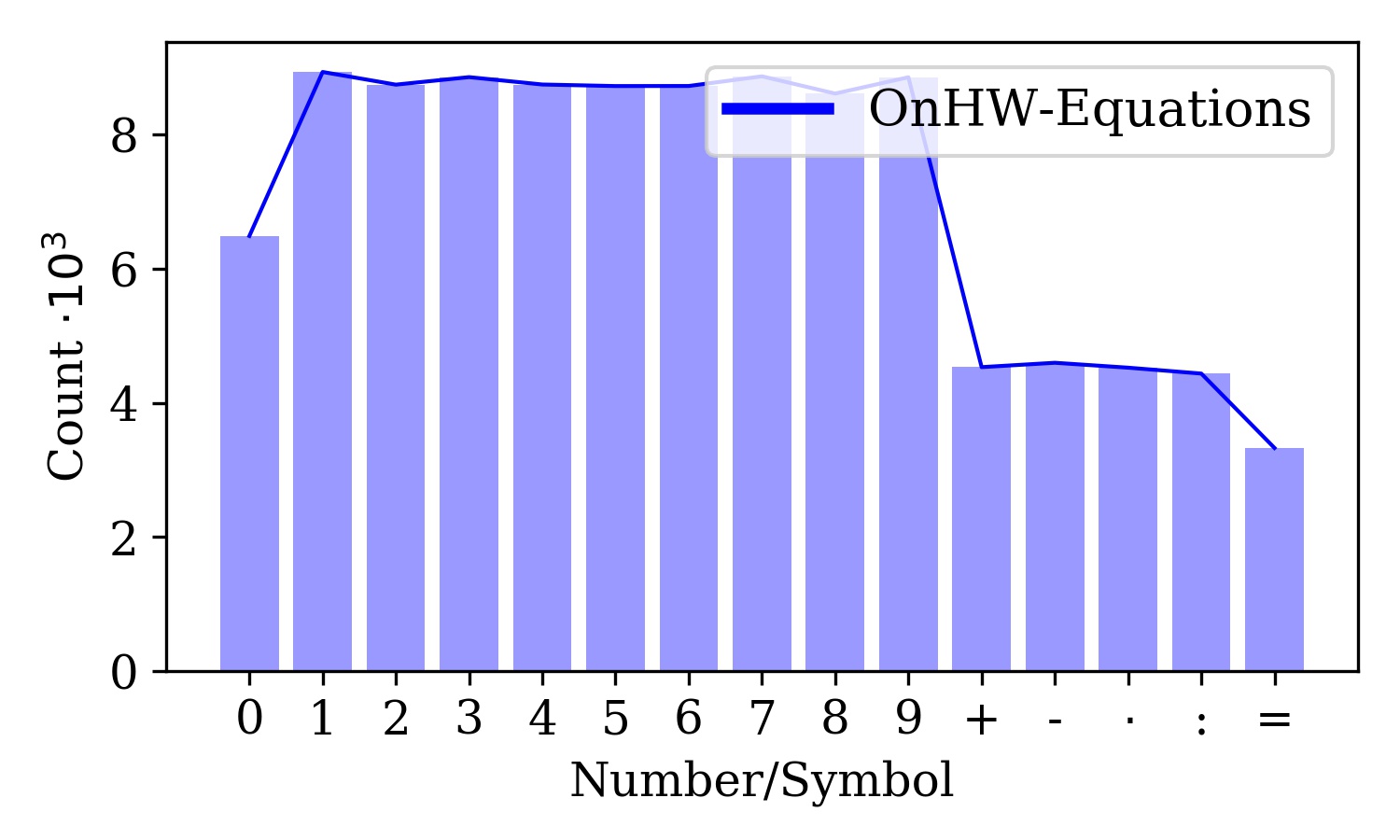}
        \vspace{-0.2cm}
    	\subcaption{Distribution of characters in equations dataset.}
    	\label{figure_statistics_dataset2}
    \end{minipage}
    \hfill
	\begin{minipage}[h]{0.32\linewidth}
        \centering
    	\includegraphics[width=1.0\linewidth]{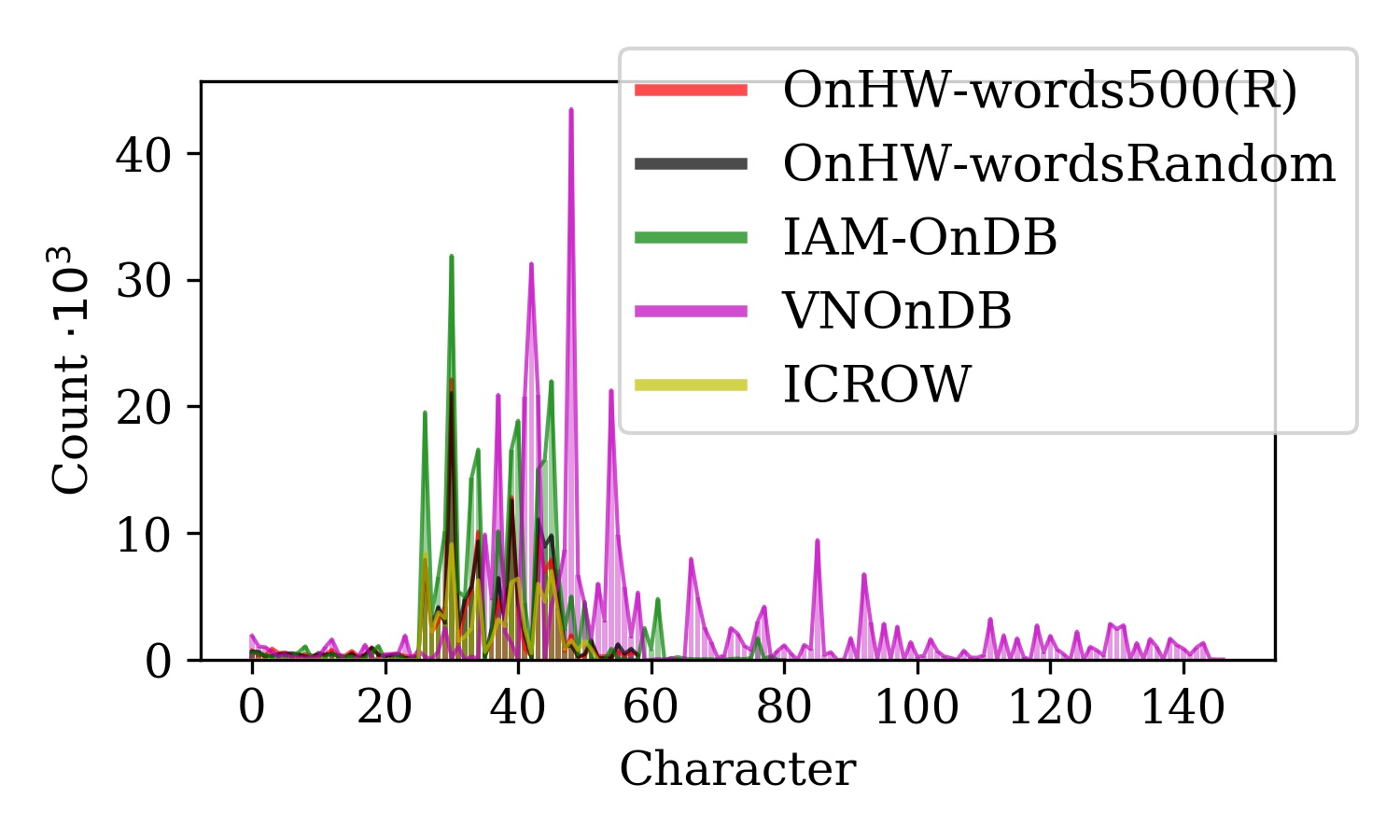}
        \vspace{-0.2cm}
    	\subcaption{Distribution of characters in words and sentences datasets.}
    	\label{figure_statistics_dataset3}
    \end{minipage}
    \caption{Properties of our and state-of-the-art datasets.}
    \label{figure_dataset_properties}
\end{figure*}

Figure~\ref{figure_dataset_properties} compares statistical properties, i.e., the number of samples, sample lengths and character distributions, between our dataset and the state-of-the-art datasets. The IAM-OnDB (line level) and VNOnDB-words datasets consist of more samples and total number of characters compared to our OnHW datasets, but at the same time use a higher number of classes (81 and 147). The IAM-OnDB samples have higher lengths (up to 64) and the VNOnDB samples have smaller lengths (up to 11) (see Figure~\ref{figure_statistics_dataset1}). The VNOnDB dataset is equally distributed compared to other words datasets (see Figure~\ref{figure_statistics_dataset3}), while numbers appear more often than operators in our OnHW-equations dataset (see Figure~\ref{figure_statistics_dataset2}). See Appendices~\ref{sec_appendix_sensor} and~\ref{sec_appendix_statistics} for more details on our datasets.

\begin{figure*}[t!]
	\centering
    \includegraphics[width=0.95\linewidth]{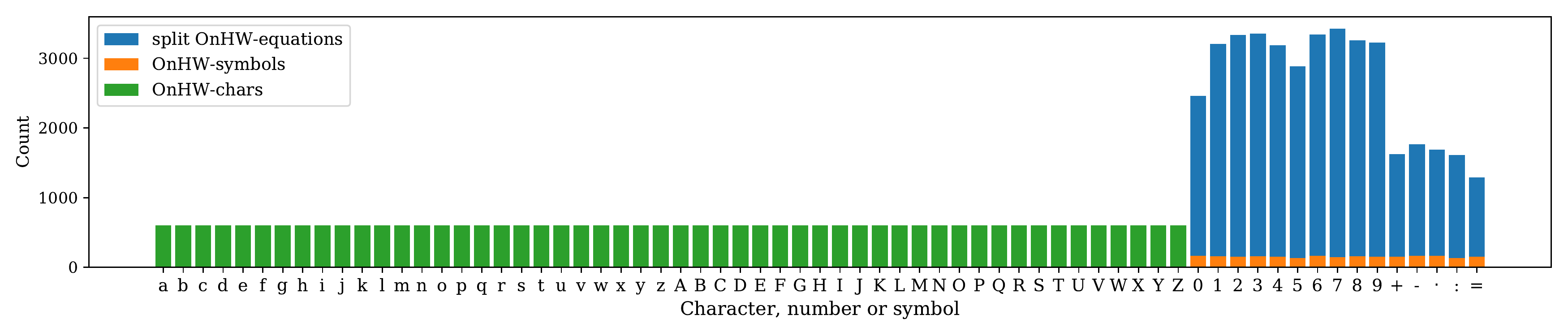}
    \caption{Distribution of samples for the OnHW-chars, OnHW-symbols and split OnHW-equations datasets.}
    \label{figure_distribution_single}
\end{figure*}

\paragraph{Datasets for Single Character Classification.} For the OnHW-equations dataset, it is possible to split the sensor sequence based on the force sensor as the pen is lifted between every single character. This approach provides another useful dataset for a single character classification task. We set split constraints for long tip lifts and recursively split these sequences by assigning a possible number of strokes per character. This results in a total of 39,643 single characters. Furthermore, we recorded the \textbf{OnHW-symbols} dataset with the same labels (numbers \texttt{0} to \texttt{9} and operators \texttt{+}, \texttt{-}, $\cdot$, \texttt{:}, \texttt{=}), written by 27 writers and a total of 2,326 single characters. Figure~\ref{figure_distribution_single} compares the distribution of sample numbers for the OnHW-chars~\cite{ott} (characters), and OnHW-symbols as well as split OnHW-equations (numbers, symbols) datasets. While the samples are equally distributed for small and capital characters ($\approx$ 600 per character), the numbers and symbols are unevenly distributed for the split OnHW-equations dataset (similar to Figure~\ref{figure_statistics_dataset2}).

\subsection{Evaluation Metrics}
\label{chap_eval_metrics}

We define a set of task-specific seq2seq and single character-based evaluation metrics that are commonly used in the community. Metrics for seq2seq evaluation are the \textbf{character error rate} (CER) and \textbf{word error rate} (WER) that are based on the \textbf{Edit distance} (ED). The ED is the minimum number of substitutions $S$, insertions $I$ and deletions $D$ required to change the sequences $\mathbf{f} = (f_1, \ldots, f_r)$ into $\mathbf{g} = (g_1, \ldots, g_n)$ with lengths $r$ and $n$, respectively. The ED is defined by
\[\label{equ_edit_distance}
\resizebox{.9\linewidth}{!}{$
    ED_{i,j} =
    \begin{cases}
        ED_{i-1,j-1} \hspace{3.14cm} \text{for} \,\, f_i = g_j \\
        \text{min}
        \begin{cases}
            ED_{i-1,j} + D(f_i) \\
            ED_{i,j-1} + I(g_j) \hspace{1.34cm} \text{for} \,\, f_i \neq g_j \\
            ED_{i-1,j-1} + S(f_i, g_j) \\
        \end{cases}
    \end{cases}
$}
\tag{1}\]
for $1 \leq i \leq r, 1 \leq j \leq n$, $ED_{i,0} = \sum_{k=1}^i D(f_k)$ for $1 \leq i \leq r$, and $ED_{0,j} = \sum_{k=1}^j I(g_k)$ for $1 \leq j \leq n$ \cite{damerau}. We define the CER $=\frac{S_c + I_c + D_c}{N_c}$ as the ED, the sum of character substitutions $S_c$, insertions $I_c$ and deletions $D_c$, divided by the total number of characters in the set $N_c$. Similarly, the WER $=\frac{S_w + I_w + D_w}{N_w}$ is computed with word operations $S_w$, $I_w$ and $D_w$, and number of words in the set $N_w$ \cite{kang}. For single character evaluation, we use the \textbf{character recognition rate} (CRR) that is the number of correctly classified characters divided by the total number of characters in the test set.

\section{Benchmark Methods}
\label{chap_method}

This section formally defines the seq2seq classification task and our loss functions. We propose our architecture for HWR from IMU-enhanced pens and describe our data augmentation techniques.

\paragraph{Sequence-based Classification Task.} An MTS $\mathbf{U} = \{\mathbf{u}_1,\\ \ldots,\mathbf{u}_m\} \in \mathbb{R}^{m \times l}$ is an ordered sequence of $l \in \mathbb{N}$ streams with $\mathbf{u}_i = (u_{i,1},\ldots, u_{i,l}), i \in \{1,\ldots,m\}$, where $m$ is the length of the time-series that is variable and $l$ is the number of dimensions. Each MTS is associated with $\mathbf{v}$, a sequence of $L$ class labels from a pre-defined label set $\Omega$ with K classes. For our classification task, $\mathbf{v} \in \Omega^L$ describes words and equations. The training set is a subset of the array $\mathcal{U} = \{\mathbf{U}_1, \ldots,\mathbf{U}_n\} \in \mathbb{R}^{n \times m \times l}$, where $n$ is the number of time-series, and the corresponding labels $\mathcal{V} = \{\mathbf{v}_1,\ldots, \mathbf{v}_n\} \in \Omega^{n \times L}$. The aim of the MTS classification task is to predict an unknown class label for a given MTS. We train the classifier using the loss $\mathcal{L}_{CTC}(\mathcal{U}, \mathcal{V})$ \cite{graves2}.

\paragraph{Character-based Classification Task.} In contrast to the sequence-based classification task, corresponding labels $\mathcal{V}$ for the character-based classification task are of length $L=1$. We define $p(i|\mathbf{u})$ to be the predicted probability for the $i^{th}$ class and $q(i|\mathbf{u})$ to be the true class distribution. We train the classifier using the cross entropy loss and variants against overconfidence and class imbalance \cite{lin_goyal,pereyra,reed,tanaka,wang_ma,zhang_sabuncu}.

\paragraph{Sequence-based Loss.} The CTC loss is a solution to avoid pre-segmentation of the training samples. The key idea of CTC is to transform the network outputs into a conditional probability distribution over label sequences. An intermediate label representation allows repetitions of labels and occurrences of blank labels to identify no output label. Hence, the network with the CTC loss has a softmax output layer with one more unit than there are labels. These outputs define the probabilities of all possible ways to align all label sequences with the input sequence. \cite{graves2}

\begin{figure*}[t!]
	\centering
    \includegraphics[width=0.85\linewidth]{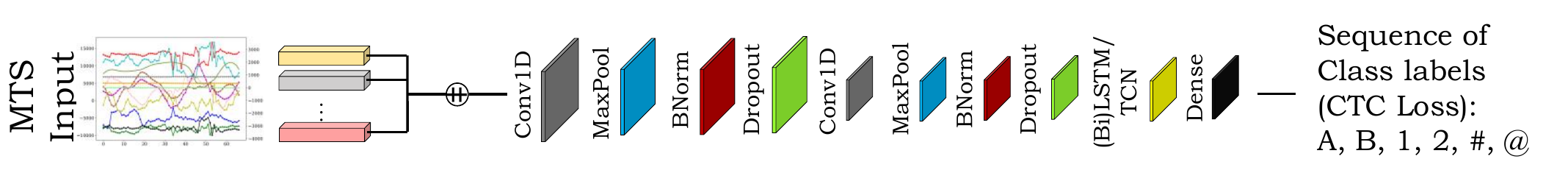}
    \vspace{-0.2cm}
    \caption{Overview of our CNN architecture in combination with a (Bi)LSTM or a TCN.}
    \label{figure_architecture1}
\end{figure*}
\begin{figure*}[t!]
	\centering
    \includegraphics[width=1.0\linewidth]{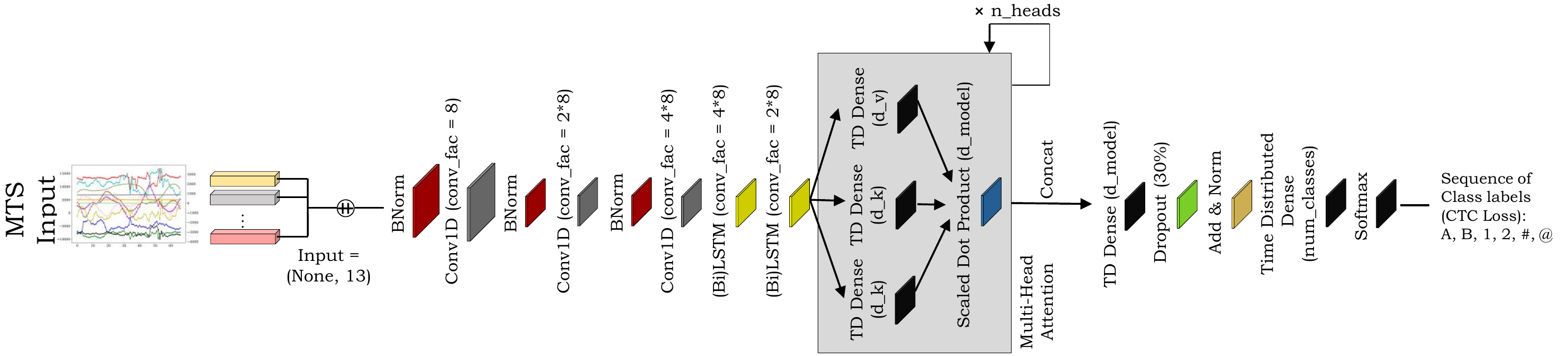}
    \vspace{-0.3cm}
    \caption{Overview of our architecture with multi-head attention.}
    \label{figure_architecture2}
\end{figure*}

\paragraph{Character-based Losses.} We use the categorical cross entropy (CCE) loss defined by
\begin{equation}
    \small \mathcal{L}_{CCE}(\mathcal{U}, \mathcal{V}) = - \frac{1}{K}\sum_{i=1}^{K} q(i|\mathbf{u}) \log p(i|\mathbf{u})
\tag{2}\end{equation}
for model training. Samples with softmax outputs that are less congruent with provided labels are implicitly weighted more than confident sample predictions (more emphasis is put on difficult samples with CCE). Hence, more emphasis is put on difficult samples, which can cause overfitting to noisy labels \cite{wang_ma,zhang_sabuncu}. To account for this imbalance, we modify the CCE loss such that it down-weights the loss assigned to well-classified examples. We use the Focal loss (FL) \cite{lin_goyal} defined by
\[
\resizebox{.98\linewidth}{!}{$
    \mathcal{L}_{FL}(\mathcal{U}, \mathcal{V}, \alpha, \gamma) = - \frac{1}{K}\sum_{i=1}^{K} \alpha_{t} \big(1 - p(i|\mathbf{u})\big)^{\gamma} q(i|\mathbf{u}) \log p(i|\mathbf{u}),
$}
\tag{3}\]
with class balance factor $\alpha \in [0,1]$, and the modulating factor $\big(1 - p(i|\mathbf{u})\big)^{\gamma}$ with focusing parameter $\gamma \geq 0$. As alternative, we apply label smoothing (LSR) \cite{pereyra} that prevents overconfidence by applying a confidence penalty through a regularization term, yielding
\begin{equation}
\begin{aligned}
\begin{split}
    \small \mathcal{L}_{LSR}(\mathcal{U}, \mathcal{V}, \beta) = &-\frac{1}{K}\sum_{i=1}^{K} \log p(i|\mathbf{u}) - \beta H\big(p(i|\mathbf{u})\big) = \\
    &-\frac{1}{K}\sum_{i=1}^{K} \log p(i|\mathbf{u}) - D_{KL}\big(x||p(i|\mathbf{u})\big),
\end{split}
\end{aligned}
\tag{4}\end{equation}
with $\beta$ the strength control of the confidence penalty. Label smoothing is equivalent to an additional Kullback-Leibler (KL) divergence term between a uniformly distributed random variable $x$ and the network's predicted distribution $p$. The bootstrapping approach \cite{reed} is another alternative for each mini-batch. The soft bootstrapping loss (SBS) is
\[
\resizebox{.99\linewidth}{!}{$
    \mathcal{L}_{SBS}(\mathcal{U}, \mathcal{V}, \beta) = - \frac{1}{K}\sum_{i=1}^{K} \big[\beta q(i|\mathbf{u}) + (1-\beta) p(i|\mathbf{u})\big] \log p(i|\mathbf{u}),
$}
\tag{5}\]
for predicted class probabilities $p$ with weighting parameter $\beta$, while the hard bootstrapping loss (HBS)
\begin{equation}
    \small \mathcal{L}_{HBS}(\mathcal{U}, \mathcal{V}, \beta) = - \frac{1}{K}\sum_{i=1}^{K} \big[\beta q(i|\mathbf{u}) + (1-\beta) z_i\big] \log p(i|\mathbf{u})
\tag{6}\end{equation}
uses the maximum a posteriori (MAP) estimation of $p$ given $\mathbf{u}$, with $z_i := \mathds{1}[i = \arg\max q_l, l=1, \ldots, K]$. MAP treats every sample equally for a higher robustness against noisy labels. This can lead to longer training times to reach convergence and can make optimization more difficult \cite{zhang_sabuncu}. The generalized cross entropy (GCE) \cite{zhang_sabuncu} loss 
\begin{equation}
    \small \mathcal{L}_{GCE}(\mathcal{U}, \mathcal{V}, \alpha) = - \frac{1}{K}\sum_{i=1}^{K} \frac{1-p(i|\mathbf{u})^\alpha}{\alpha}
\tag{7}\end{equation}
with $\alpha \in (0, 1]$ uses a negative Box-Cox transformation to combine benefits of the MAE and the CCE. The symmetric cross entropy (SCE) \cite{wang_ma} is
\begin{equation}
    \small \mathcal{L}_{SCE}(\mathcal{U}, \mathcal{V}, \alpha, \beta) = \alpha \mathcal{L}_{CCE}(\mathcal{U}, \mathcal{V}) + \beta \mathcal{L}_{RCE}(\mathcal{U}, \mathcal{V})
\tag{8}\end{equation}
based on the reverse cross entropy (RCE) loss
\begin{equation}
    \small \mathcal{L}_{RCE}(\mathcal{U}, \mathcal{V}) = - \frac{1}{K}\sum_{i=1}^{K} p(i|\mathbf{u}) \log q(i|\mathbf{u}),
\tag{9}\end{equation}
aims for a more effective and robust learning, where $\alpha$ mitigates the overfitting of CCE and $\beta$ allows for flexible exploration of the RCE. Furthermore, we make use of the joint optimization (JO) \cite{tanaka}, which overcomes the noisy labels problem by learning network parameters and labels jointly. The loss is defined by 
\begin{equation}
\begin{aligned}
\begin{split}
    \small \mathcal{L}_{JO}(\Theta, \mathcal{V}|\mathcal{U}, \alpha, \beta) = & \mathcal{L}_{CCE}(\Theta, \mathcal{V}|\mathcal{U}) + \\
    &\alpha \mathcal{L}_{p}(\Theta|\mathcal{U}) + \beta \mathcal{L}_{e}(\Theta|\mathcal{U})
\end{split}
\end{aligned}
\tag{10}\end{equation}
with regularization losses $\mathcal{L}_{p}$ and $\mathcal{L}_{e}$, and network parameters $\Theta$.

\begin{figure*}[t!]
	\centering
	\begin{minipage}[h]{0.195\linewidth}
        \centering
    	\includegraphics[width=1.0\linewidth]{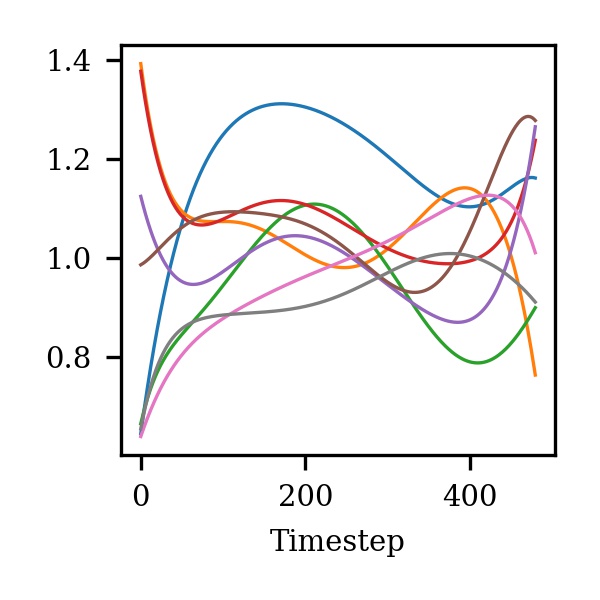}
        \vspace{-0.4cm}
        \subcaption{Bezier curve.}
    	\label{image_data_augm_a}
    \end{minipage}
    \hfill
	\begin{minipage}[h]{0.195\linewidth}
        \centering
    	\includegraphics[width=1.0\linewidth]{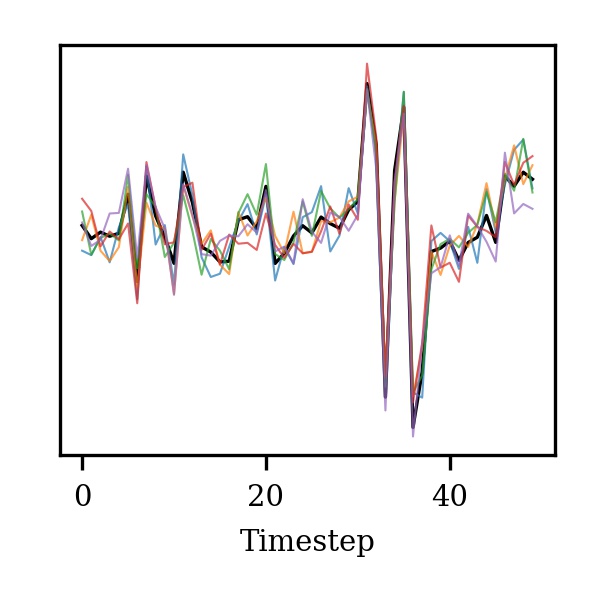}
        \vspace{-0.4cm}
        \subcaption{Jittering.}
    	\label{image_data_augm_b}
    \end{minipage}
    \hfill
	\begin{minipage}[h]{0.195\linewidth}
        \centering
    	\includegraphics[width=1.0\linewidth]{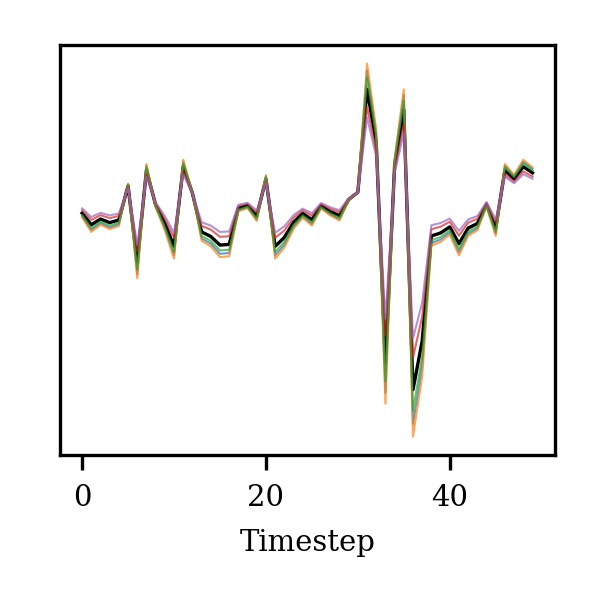}
        \vspace{-0.4cm}
        \subcaption{Scaling.}
    	\label{image_data_augm_c}
    \end{minipage}
    \hfill
	\begin{minipage}[h]{0.195\linewidth}
        \centering
    	\includegraphics[width=1.0\linewidth]{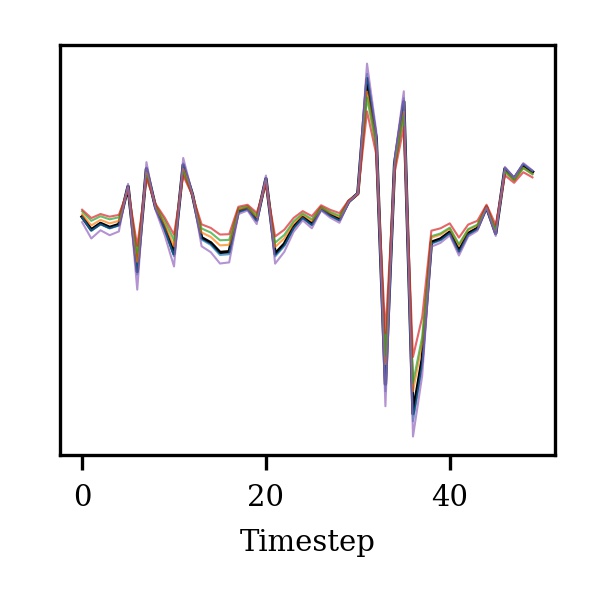}
        \vspace{-0.4cm}
        \subcaption{Magnitude warping.}
    	\label{image_data_augm_d}
    \end{minipage}
    \hfill
	\begin{minipage}[h]{0.195\linewidth}
        \centering
    	\includegraphics[width=1.0\linewidth]{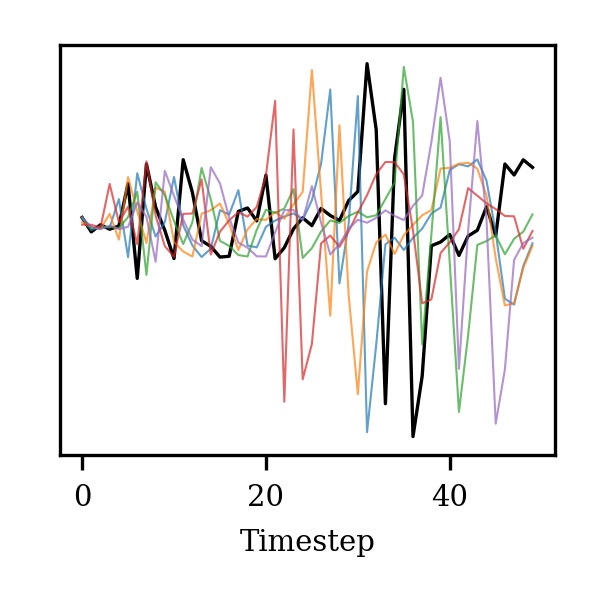}
        \vspace{-0.4cm}
        \subcaption{Time warping.}
    	\label{image_data_augm_e}
    \end{minipage}
    \vspace{-0.2cm}
    \caption{Data augmentation methods of the original sensor sample (black).}
    \label{image_data_augm}
\end{figure*}

\paragraph{Architectures.} We propose two different architectures for seq2seq sensor signal classification. For the first method (see Figure~\ref{figure_architecture1}), a convolution block consisting of 1D convolutions (200 filter, kernel size 4), max pooling (pool size 2), batch normalization and dropout (with rate 0.2) layers is used. One TCN layer of 100 units, one LSTM layer of 100 units, or two BiLSTM layers, each with 60 units, follow to extract the temporal context \cite{reimers}. While we use \texttt{tanh} activations for BiLSTM layers, we choose \texttt{ReLU} for the TCN and LSTM layers. A dense layer with 100 units with the CTC loss predicts a sequence of class labels. Second, we implement an attention-based network (see Figure~\ref{figure_architecture2}) that consists of an encoder with batch normalization, 1D convolutional and (Bi)LSTM layers. These map the input sequence $\mathbf{U} \in \mathbb{R}^{m \times l}$ to a sequence of continuous representations $\mathbf{z}$. A Transformer transforms $\mathbf{z}$ using $n_{\text{head}}$ stacked multi-head self-attention $\text{MultiHead}(Q,K,V) = \text{Concat}(\text{head}_{1}, \ldots, \text{head}_{h}) W^{O}$ with $W^{O} \in \mathbb{R}^{hd_{v} \times d_\text{model}}$. The attention consists of point-wise, fully connected time-distributed layers followed by a scaled dot product layer and layer normalization~\cite{ba} with $d_{\text{model}}$ output dimension \cite{vaswani}. $\text{head}_i = \text{Attention}(QW_{i}^{Q}, KW_{i}^{K}, VW_{i}^{V})$, where $W_{i}^{Q}, W_{i}^{K} \in \mathbb{R}^{d_{\text{model}} \times d_k}$, and $W_{i}^{V} \in \mathbb{R}^{d_{\text{model}} \times d_v}$. The attention can be described as mapping a set of key-value pairs of dimension $d_v$ and a query of dimension $d_k$ to an output, and is computed by $\text{Attention}(Q,K,V) = \text{softmax}\Big(\frac{Q K^{T}}{\sqrt{d_k}}\Big)V$. The matrices $Q$, $K$ and $V$ are a set of queries, keys and values.

\paragraph{Data Augmentation.} As the size of the datasets is limited, data augmentation is a critical pre-processing step for networks to prevent overfitting and improve generalization. However, it is not obvious how to carry out label-preserving augmentation in some domains, i.e., scaling of acceleration signals \cite{um}. We apply the following different data augmentation methods for wearable sensor data on each sensor channel at 50\% probability. \textbf{Time warping} perturbs the temporal location by smoothly distorting the time intervals between samples that, e.g., simulates different sampling rates through time shifts of the connection between device and tablet. \textbf{Scaling} changes the magnitude of the data in a window by multiplying by a random scalar $\sigma = \pm 0.1$ that augments drifts in the signals. \textbf{Shifting} adds a random value $\alpha = \pm 200$ to the force data and $\alpha = \pm 20$ to the other sensor data. While \textbf{jittering} is a way of simulating additive sensor noise by adding a multiple $\sigma = \pm 0.1$ of the standard deviation to all sensor channels, \textbf{magnitude warping} changes the magnitude by convolving the data window with a smooth curve varying around $[0.7, 1.3]$ (only for the accelerometer data). For time and magnitude warping, the data is augmented by B\'{e}zier curves in the interval $[1-\sigma, 1+\sigma]$ that are generated based on 10 random points. As one sample is represented by a sequence of characters and a sample cannot be split into sub-sequences, applying cropping and permutation augmentation is not possible. Figure~\ref{image_data_augm} zooms into the augmented sensor data of the x-axis signal from Figure~\ref{figure_exemplary_signal}.

\section{Experiments}
\label{chap_results}

This section provides evaluation results for the seq2seq (Section~\ref{chap_eval_seqseq}) and the single character-based classification task (Section~\ref{chap_eval_character}), and evaluates left-handed datasets (Section~\ref{chap_eval_left}). We propose a writer-dependent evaluation in Section~\ref{chap_eval_ed_writer}.

\begin{table*}[t!]
\begin{flushleft}
    \caption{Evaluation results (WER, CER) in \% (mean and standard deviation) for our OnHW-equations, OnHW-words500(R), OnHW-wordsRandom and OnHW-wordsTraj writer-dependent (WD) and writer-independent (WI) datasets, and the publicly available IAM-OnDB~\cite{liwicki} (line level) and VNOnDB-words~\cite{nguyen} datasets.}
    \label{table_results1a}
    \setlength{\tabcolsep}{2.3pt}
    \small
\begin{tabular}{ p{0.5cm} | p{0.5cm} | p{0.5cm} | p{0.5cm} | p{0.5cm} | p{0.5cm} | p{0.5cm} | p{0.5cm} | p{0.5cm} | p{0.5cm} | p{0.5cm} | p{0.5cm} | p{0.5cm} | p{0.5cm} | p{0.5cm} | p{0.5cm} | p{0.5cm} }
    \multicolumn{2}{c|}{\textbf{Method}} & \multicolumn{1}{c|}{\textbf{Met-}} & \multicolumn{4}{c|}{\textbf{OnHW-equations}} & \multicolumn{6}{c|}{\textbf{OnHW-words500(R)}} & \multicolumn{4}{c}{\textbf{OnHW-wordsRandom}}  \\
    \multicolumn{2}{c|}{} & \multicolumn{1}{c|}{\textbf{ric}} & \multicolumn{2}{c|}{\textbf{WD}} & \multicolumn{2}{c|}{\textbf{WI}} & \multicolumn{2}{c|}{\textbf{WD}} & \multicolumn{2}{c|}{\textbf{WI}} & \multicolumn{2}{c|}{\textbf{Random}} & \multicolumn{2}{c|}{\textbf{WD}} & \multicolumn{2}{c}{\textbf{WI}} \\
    \multicolumn{2}{c|}{} & \multicolumn{1}{c|}{} & \multicolumn{1}{c}{WER} & \multicolumn{1}{c|}{CER} & \multicolumn{1}{c}{WER} & \multicolumn{1}{c|}{CER} & \multicolumn{1}{c}{WER} & \multicolumn{1}{c|}{CER} & \multicolumn{1}{c}{WER} & \multicolumn{1}{c|}{CER} & \multicolumn{1}{c}{WER} & \multicolumn{1}{c|}{CER} & \multicolumn{1}{c}{WER} & \multicolumn{1}{c|}{CER} & \multicolumn{1}{c}{WER} & \multicolumn{1}{c}{CER} \\ \hline
    \multicolumn{2}{l|}{CNN+LSTM} & \multicolumn{1}{c|}{Mean} & \multicolumn{1}{r}{22.96} & \multicolumn{1}{r|}{3.50} & \multicolumn{1}{r}{69.22} & \multicolumn{1}{r|}{18.11} & \multicolumn{1}{r}{80.70} & \multicolumn{1}{r|}{28.41} & \multicolumn{1}{r}{93.30} & \multicolumn{1}{r|}{48.24} & \multicolumn{1}{r}{76.80} & \multicolumn{1}{r|}{23.73} & \multicolumn{1}{r}{82.29} & \multicolumn{1}{r|}{17.90} & \multicolumn{1}{r}{93.90} & \multicolumn{1}{r}{46.92} \\
\end{tabular}
    \setlength{\tabcolsep}{2.7pt}
    \vspace{-0.1cm}
    \small \begin{tabular}{ p{2.67cm} | p{0.5cm} | p{0.5cm} | p{0.5cm} | p{0.5cm} | p{0.5cm} | p{0.5cm} | p{0.5cm} | p{0.5cm} | p{0.5cm} | p{0.5cm} | p{0.5cm} | p{0.5cm} | p{0.5cm} | p{0.5cm} | p{0.5cm} | p{0.5cm} }
    \multicolumn{1}{c|}{\textbf{Method}} & \multicolumn{1}{c|}{\textbf{Met-}} & \multicolumn{6}{c|}{\textbf{OnHW-wordsTraj}$^1$} & \multicolumn{4}{c|}{\textbf{IAM-OnDB}~\cite{liwicki}} & \multicolumn{4}{c}{\textbf{VNOnDB-words}~\cite{nguyen}} \\
    & \multicolumn{1}{c|}{\textbf{ric}} & \multicolumn{2}{c|}{\textbf{Camera}$^2$} & \multicolumn{2}{c|}{\textbf{IMU}} & \multicolumn{2}{c|}{\textbf{Trajectory}} & \multicolumn{2}{c|}{\textbf{WD}} & \multicolumn{2}{c|}{\textbf{WI}} & \multicolumn{2}{c|}{\textbf{WD}} & \multicolumn{2}{c}{\textbf{WI}} \\
    \multicolumn{1}{r|}{} & \multicolumn{1}{r|}{} & \multicolumn{1}{c}{WER} & \multicolumn{1}{c|}{CER} & \multicolumn{1}{c}{WER} & \multicolumn{1}{c|}{CER} & \multicolumn{1}{c}{WER} & \multicolumn{1}{c|}{CER} & \multicolumn{1}{c}{WER} & \multicolumn{1}{c|}{CER} & \multicolumn{1}{c}{WER} & \multicolumn{1}{c|}{CER} & \multicolumn{1}{c}{WER} & \multicolumn{1}{c|}{CER} & \multicolumn{1}{c}{WER} & \multicolumn{1}{c}{CER} \\ \hline
    \multicolumn{1}{l|}{CNN+LSTM} & \multicolumn{1}{c|}{Mean} & \multicolumn{1}{r}{60.50} & \multicolumn{1}{r|}{14.93} & \multicolumn{1}{r}{57.00} & \multicolumn{1}{r|}{8.95} & \multicolumn{1}{r}{61.10} & \multicolumn{1}{r|}{10.66} & \multicolumn{1}{r}{83.14} & \multicolumn{1}{r|}{11.23} & \multicolumn{1}{r}{84.56} & \multicolumn{1}{r|}{12.96} & \multicolumn{1}{r}{60.54} & \multicolumn{1}{r|}{26.12} & \multicolumn{1}{r}{66.17} & \multicolumn{1}{r}{29.13} \\
\end{tabular}
\end{flushleft}
\end{table*}

\paragraph{Hardware and Training Setup.} For all experiments we use Nvidia Tesla V100-SXM2 GPUs with 32 GB VRAM equipped with Core Xeon CPUs and 192 GB RAM. We use the Adam optimizer with a learning rate of $10^{-4}$. We run each experiment for 1,000 epochs with a batch size of 50 (unless stated differently) and report results for the best epoch. We split each dataset into five approx. 80/20 train/test splits, and report the mean and standard deviation of the WER and CER. We use our OnHW-equations, OnHW-words500(R), OnHW-wordsRandom and OnHW-wordsTraj as well as the IAM-OnDB~\cite{liwicki} and VNOnDB-words~\cite{nguyen} datasets for the sequence-based classification task, and the OnHW-symbols, split OnHW-equations and OnHW-chars~\cite{ott} datasets for the single character-based classification task. Each model is trained from scratch for every dataset. We make use of the time-series classification toolbox tsai~\cite{tsai} that contains a variety of state-of-the-art techniques \cite{bai,chung,elsayed,fauvel,Fawaz,he_zhang,karim,rahimian,dempster,tang_long,wang_2018,wang_2016,zerveas,zou}.

\subsection{Seq2seq Task Evaluation}
\label{chap_eval_seqseq}

\paragraph{Method and Architecture Evaluation.} We first evaluate our CNN and attention-based models for the seq2seq classification task. A summary of results is given in Table~\ref{table_results1a}. For all datasets our CNN+BiLSTM model significantly outperforms the CNN+LSTM and CNN+TCN models. The attention-based model performs poorly on large datasets (OnHW-[equations, words500(R), wordsRandom]), but yields better results than the CNN+ TCN on our OnHW-wordsTraj camera-based dataset and outperforms the CNN+LSTM and CNN+TCN models on the trajectory-based dataset. The CNN+BiLSTM model achieves a very good CER of 1.78\% on the OnHW-equations WD dataset that increases to 12.98\% for the WI task. For the words, IAM-OnDB and VNOnDB datasets, the WI classification task is more difficult. While we achieve very low CERs, the WERs are higher as no lexicon or language model is used. While for the OnHW-wordsRandom dataset the CER of 7.87\% for the WD task increases notably to 35.22\% for the WI task, the difference for the OnHW-words500 dataset is smaller (17.16\% CER for the WD task and 27.80\% for the WI task) as words in the validation set do not appear in the training set (WD task). For the OnHW-words500R dataset, the CER decreases to 5.20\% as the split is randomly shuffled. Our OnHW-wordsTraj dataset allows a comparison of three recording devices (i.e., trajectory, IMU and camera). From the CNN+BiLSTM model we see that the spatio-temporal trajectory-based classification task is easier than OnHWR from IMU-enhanced pens. Furthermore, it is challenging to learn the transformation from camera to paper plane.

\begin{figure}[t!]
	\centering
    \includegraphics[trim=0 0 0 11, clip, width=0.72\linewidth]{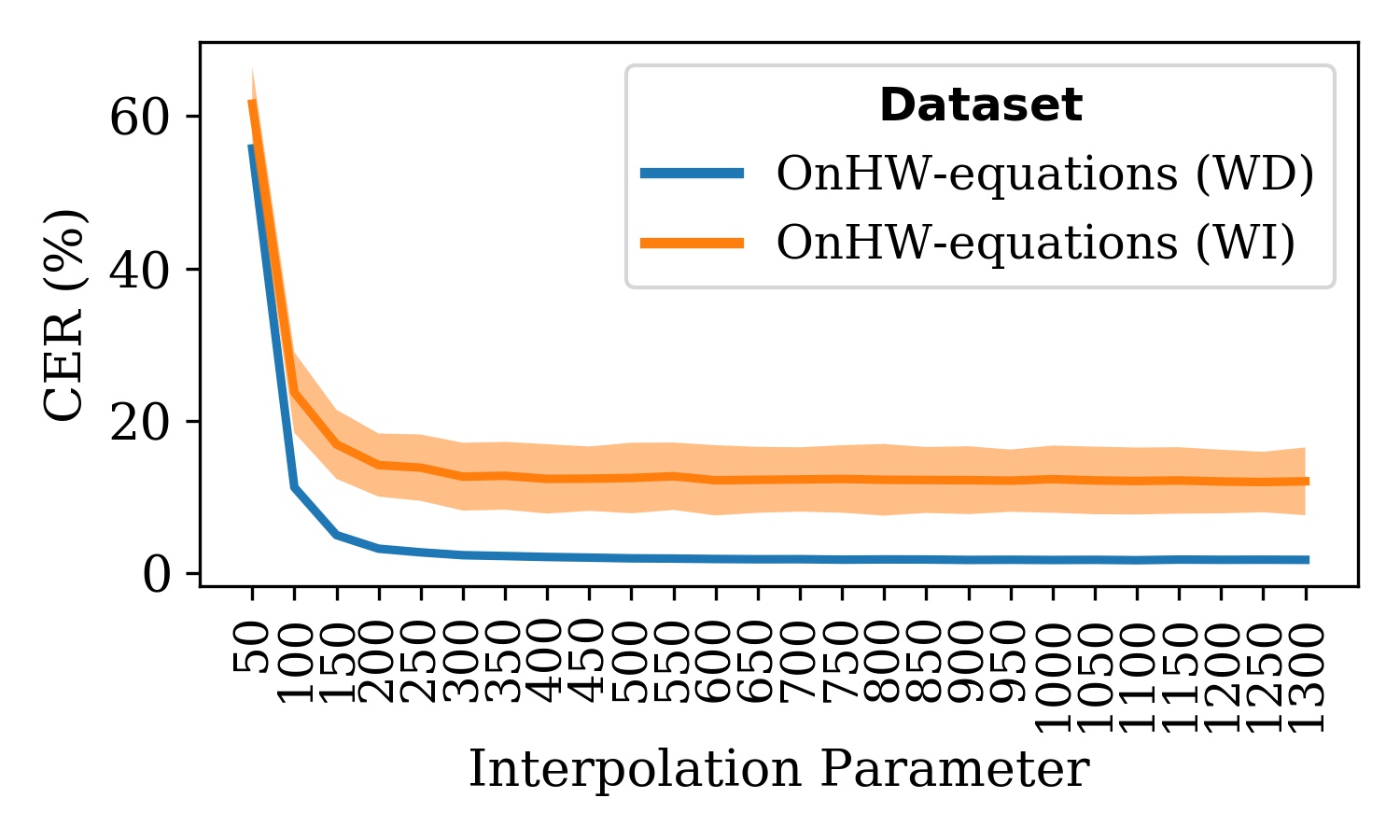}
    \vspace{-0.2cm}
    \caption{CER of InceptionTime~\cite{Fawaz} for different interpolated time-series lengths on OnHW-equations dataset.}
    \label{figure_interpolation}
    \vspace{-0.1cm}
\end{figure}

\begin{figure*}[t!]
	\centering
	\begin{minipage}[b]{0.245\linewidth}
        \centering
        \includegraphics[width=1.0\linewidth]{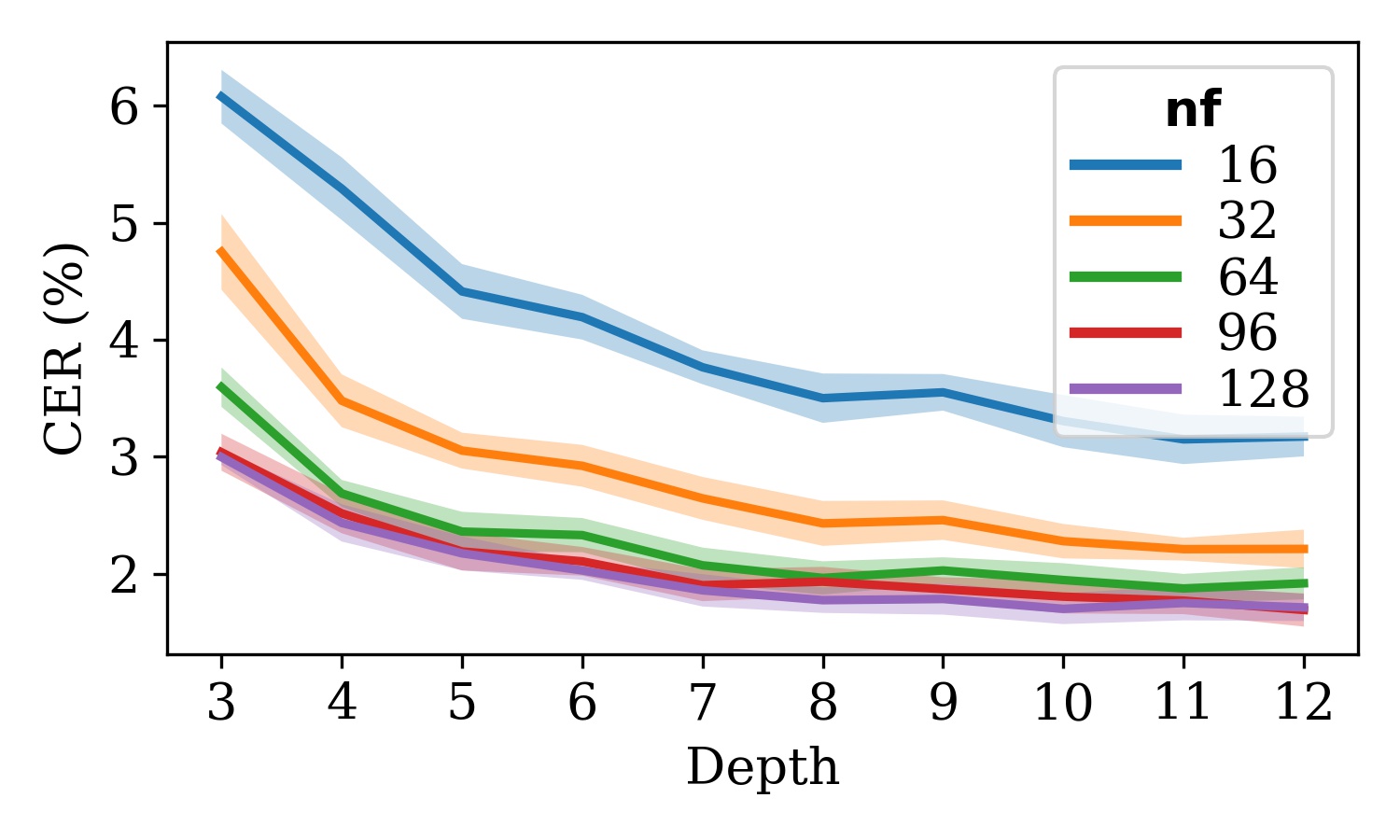}
        \vspace{-0.4cm}
        \subcaption{WD.}
    \end{minipage}
    \hfill
	\begin{minipage}[b]{0.245\linewidth}
        \centering
        \includegraphics[width=1.0\linewidth]{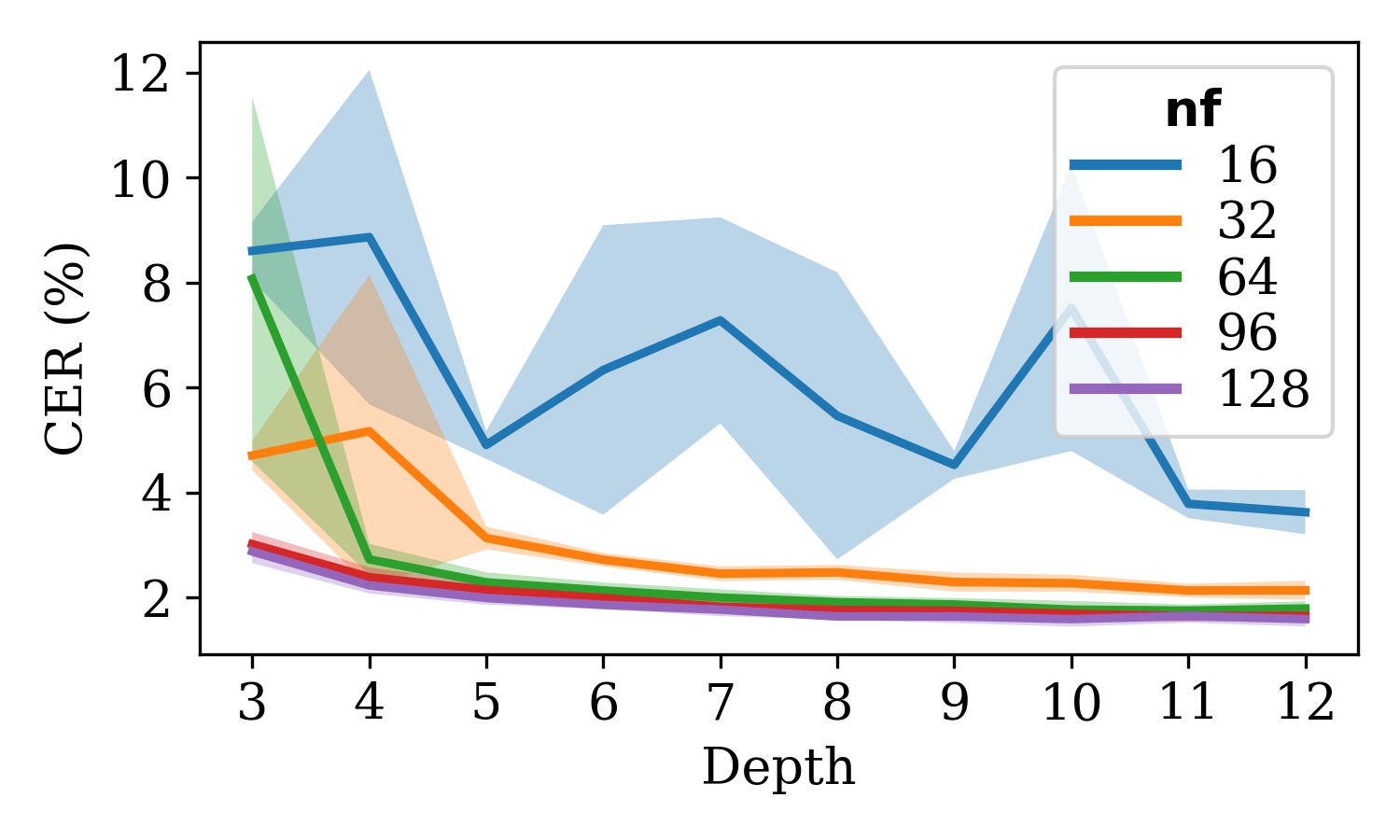}
        \vspace{-0.4cm}
        \subcaption{WD: BiLSTM.}
    \end{minipage}
    \hfill
	\begin{minipage}[b]{0.245\linewidth}
        \centering
        \includegraphics[width=1.0\linewidth]{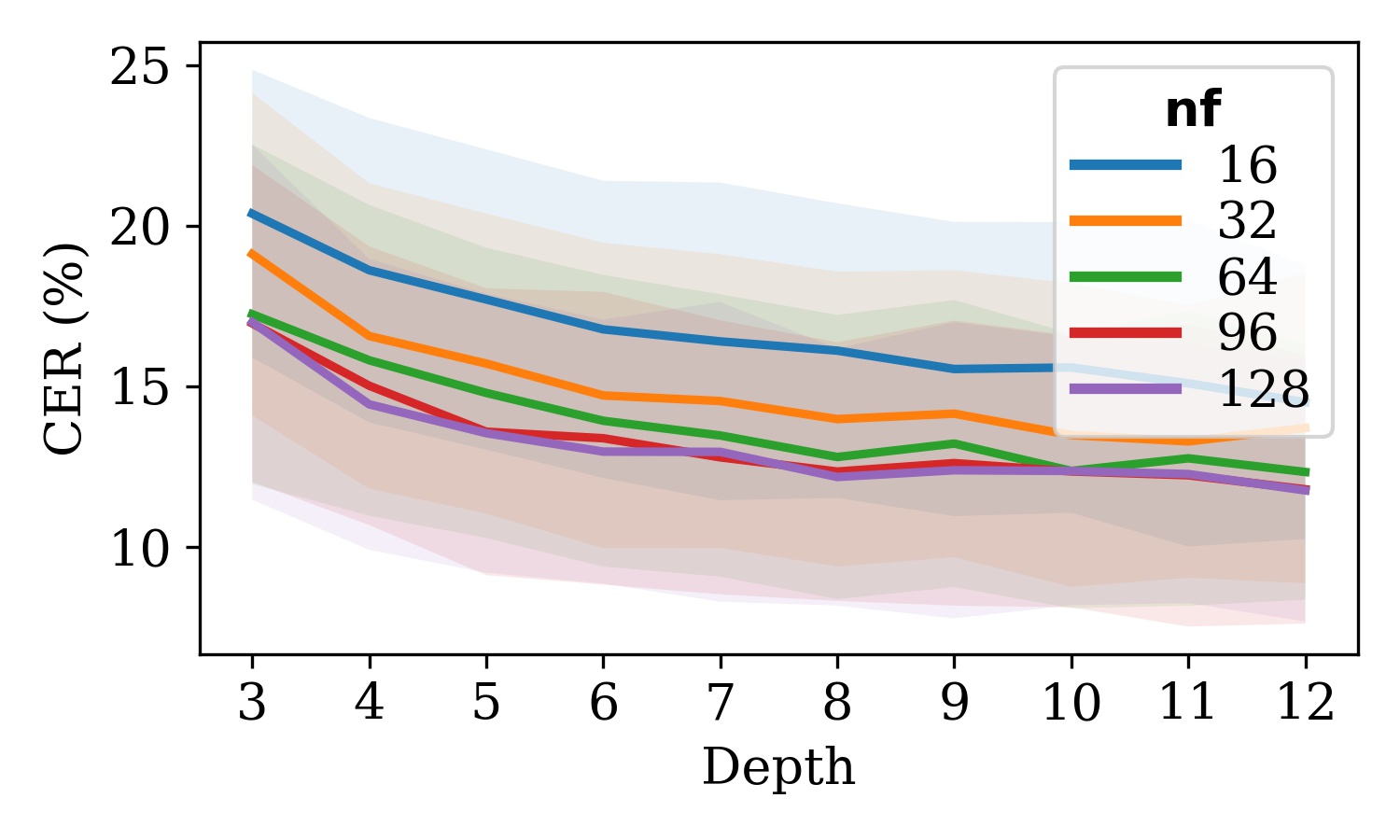}
        \vspace{-0.4cm}
        \subcaption{WI.}
    \end{minipage}
    \hfill
	\begin{minipage}[b]{0.245\linewidth}
        \centering
        \includegraphics[width=1.0\linewidth]{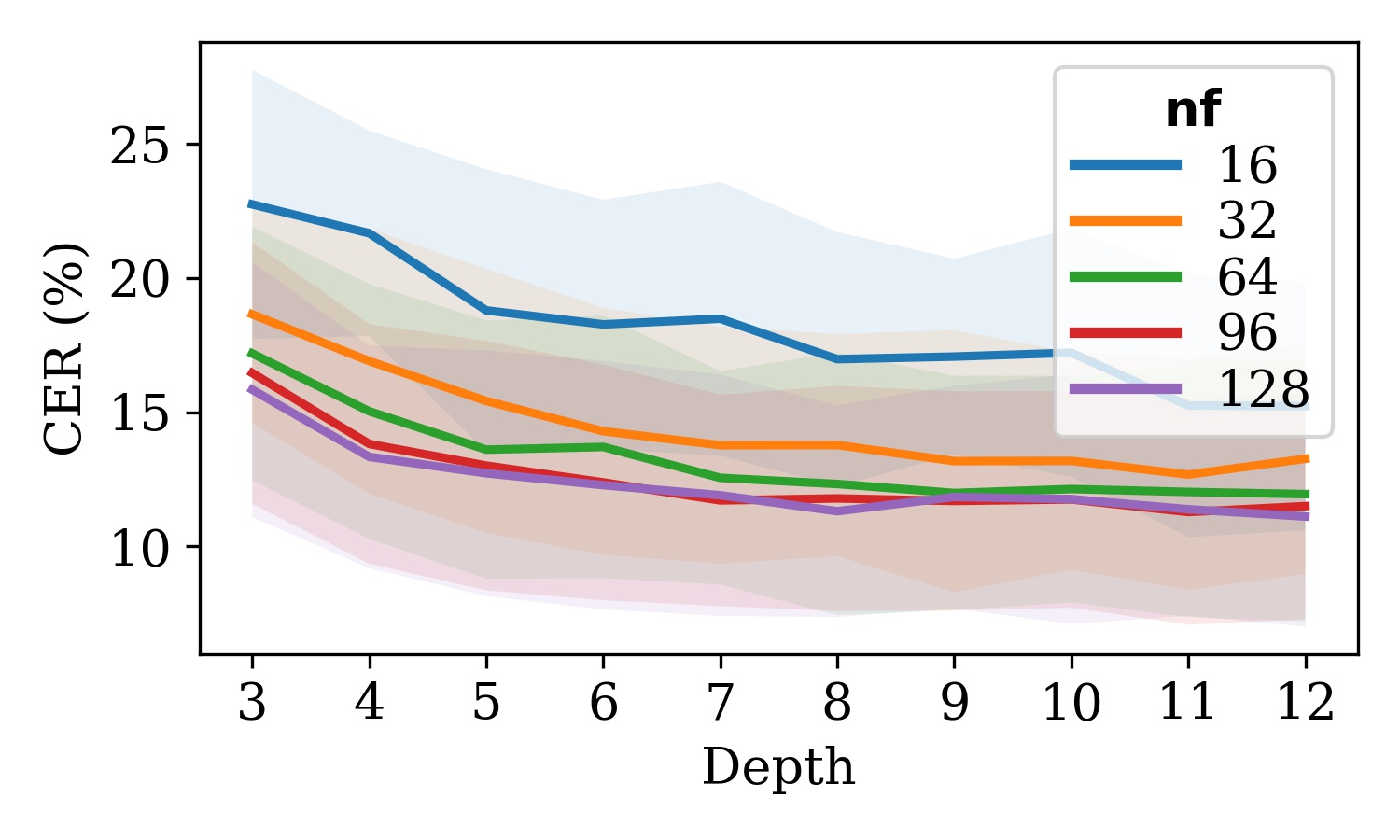}
        \vspace{-0.4cm}
        \subcaption{WI: BiLSTM.}
    \end{minipage}
    \vspace{-0.6cm}
    \caption{Hyperparameter search of $depth$ and $nf$ for InceptionTime~\cite{Fawaz} with and without BiLSTM on the OnHW-equations datasets averaged over 5 splits.}
    \label{image_hps}
\end{figure*}

\begin{figure*}[t!]
	\centering
	\begin{minipage}[b]{0.245\linewidth}
        \centering
        \includegraphics[width=1.0\linewidth]{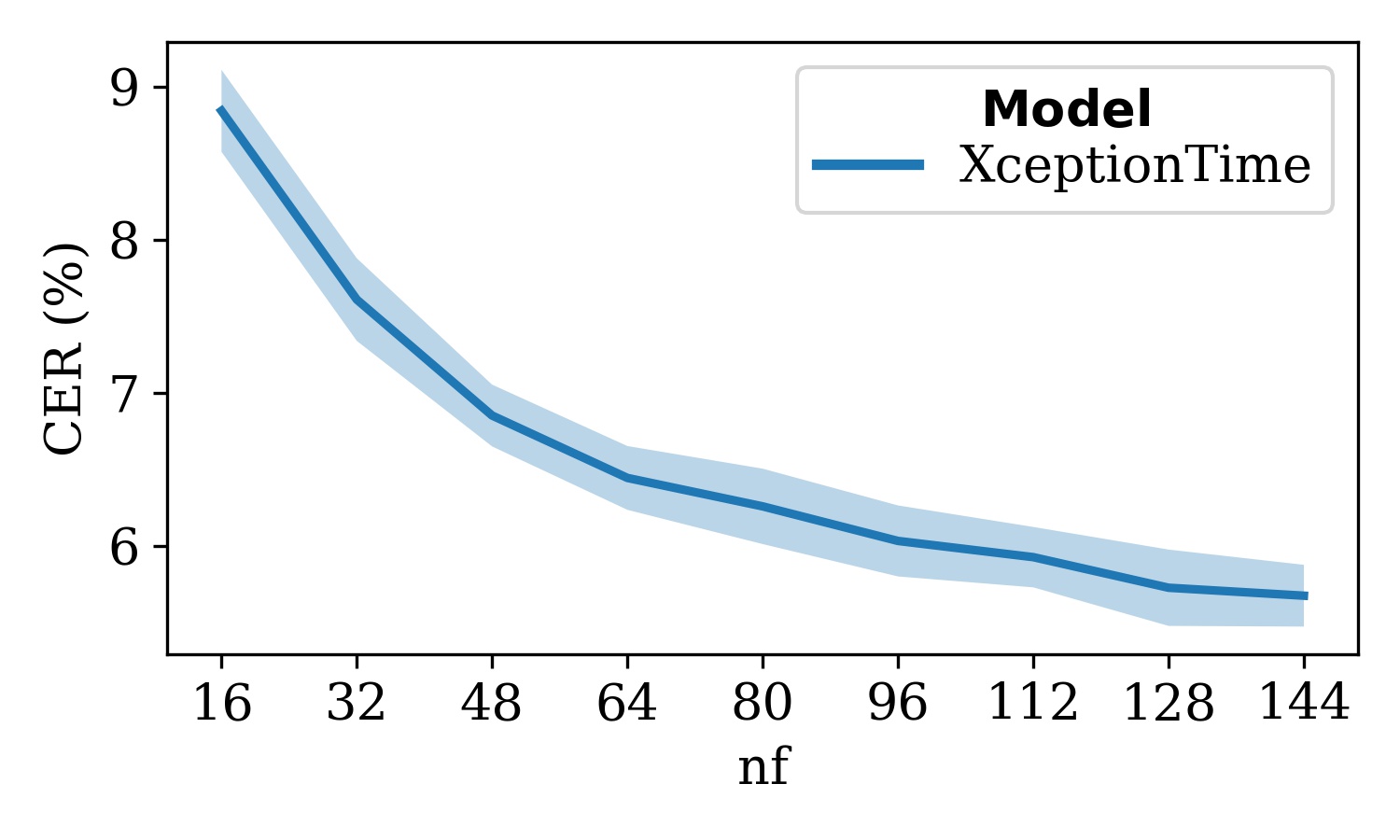}
        \vspace{-0.4cm}
        \subcaption{WD.}
    \end{minipage}
    \hfill
	\begin{minipage}[b]{0.245\linewidth}
        \centering
        \includegraphics[width=1.0\linewidth]{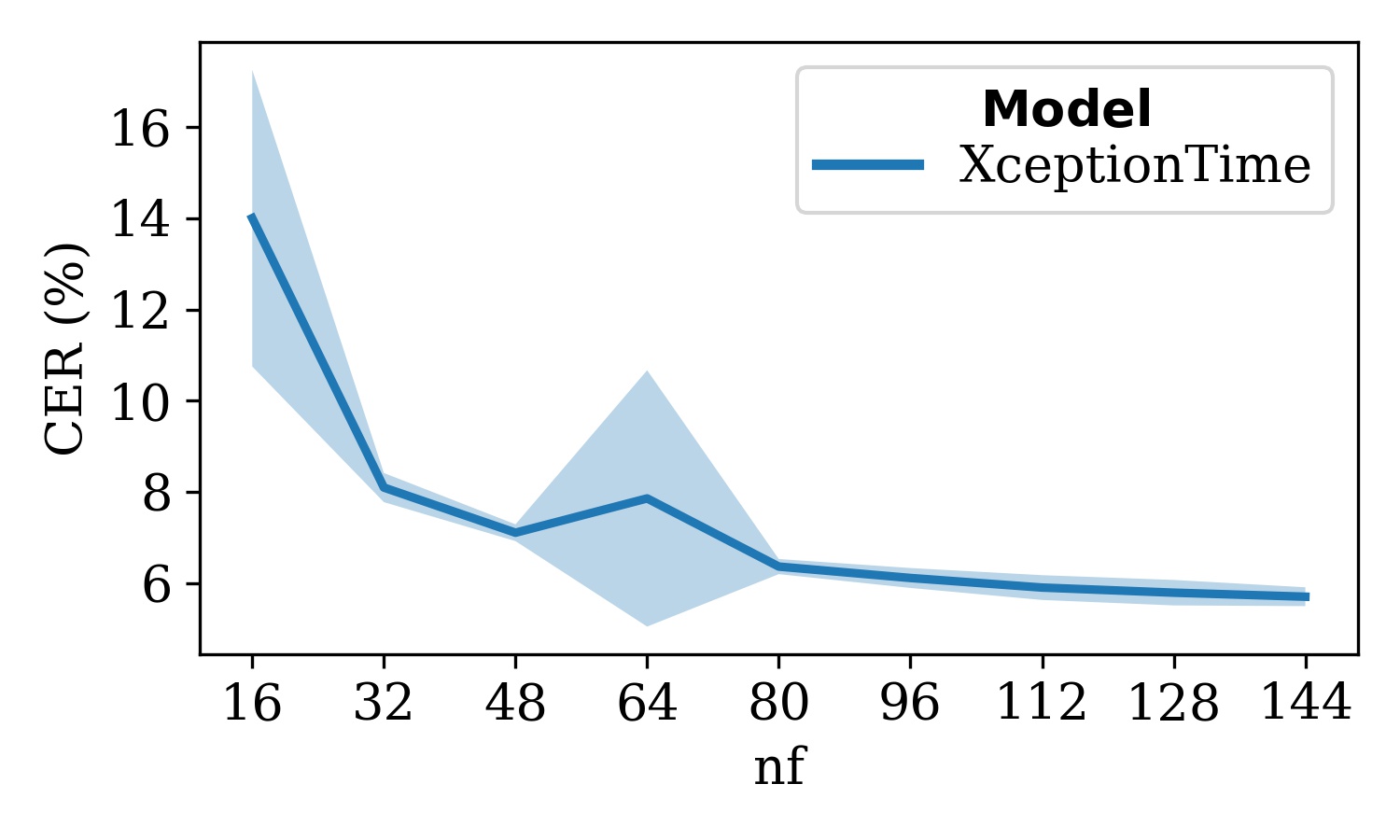}
        \vspace{-0.4cm}
        \subcaption{WD: BiLSTM.}
    \end{minipage}
    \hfill
	\begin{minipage}[b]{0.245\linewidth}
        \centering
        \includegraphics[width=1.0\linewidth]{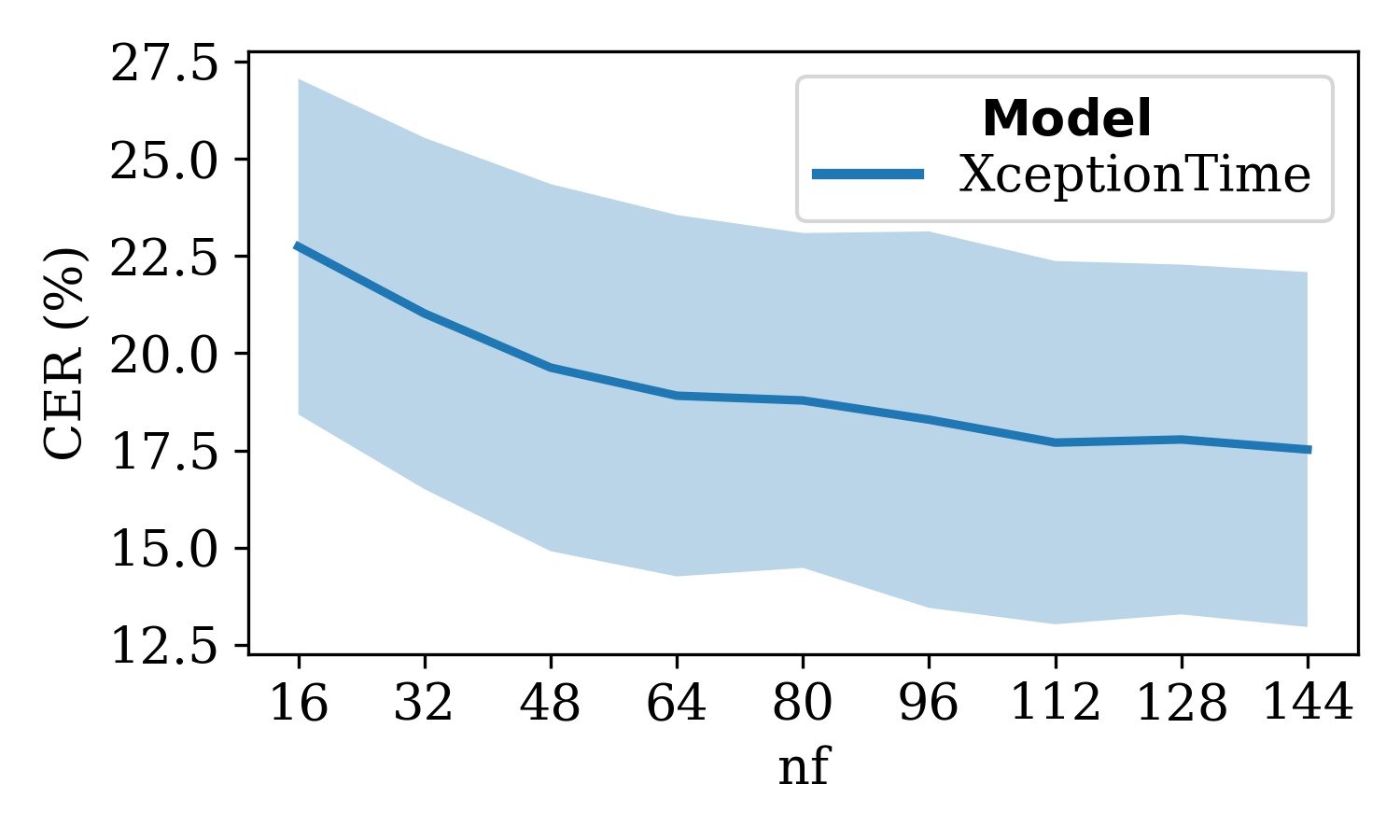}
        \vspace{-0.4cm}
        \subcaption{WI.}
    \end{minipage}
    \hfill
	\begin{minipage}[b]{0.245\linewidth}
        \centering
        \includegraphics[width=1.0\linewidth]{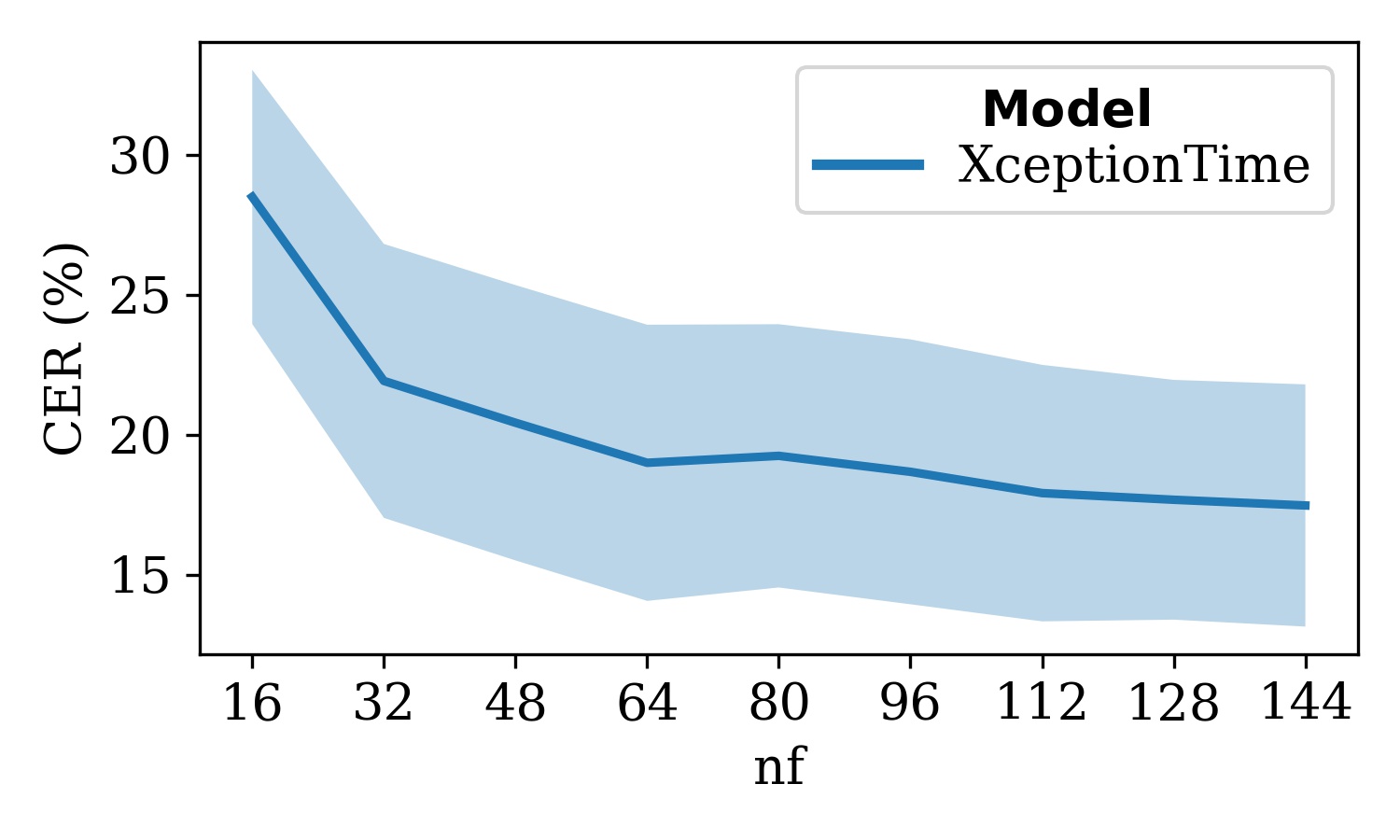}
        \vspace{-0.4cm}
        \subcaption{WI: BiLSTM.}
    \end{minipage}
    \vspace{-0.6cm}
    \caption{Hyperparameter search of $nf$ for XceptionTime~\cite{rahimian} with and without BiLSTM on the OnHW-equations datasets averaged over 5 splits.}
    \label{image_hps_xceptionTime}
\end{figure*}

\begin{figure*}[t!]
	\centering
	\begin{minipage}[b]{0.245\linewidth}
        \centering
        \includegraphics[width=1.0\linewidth]{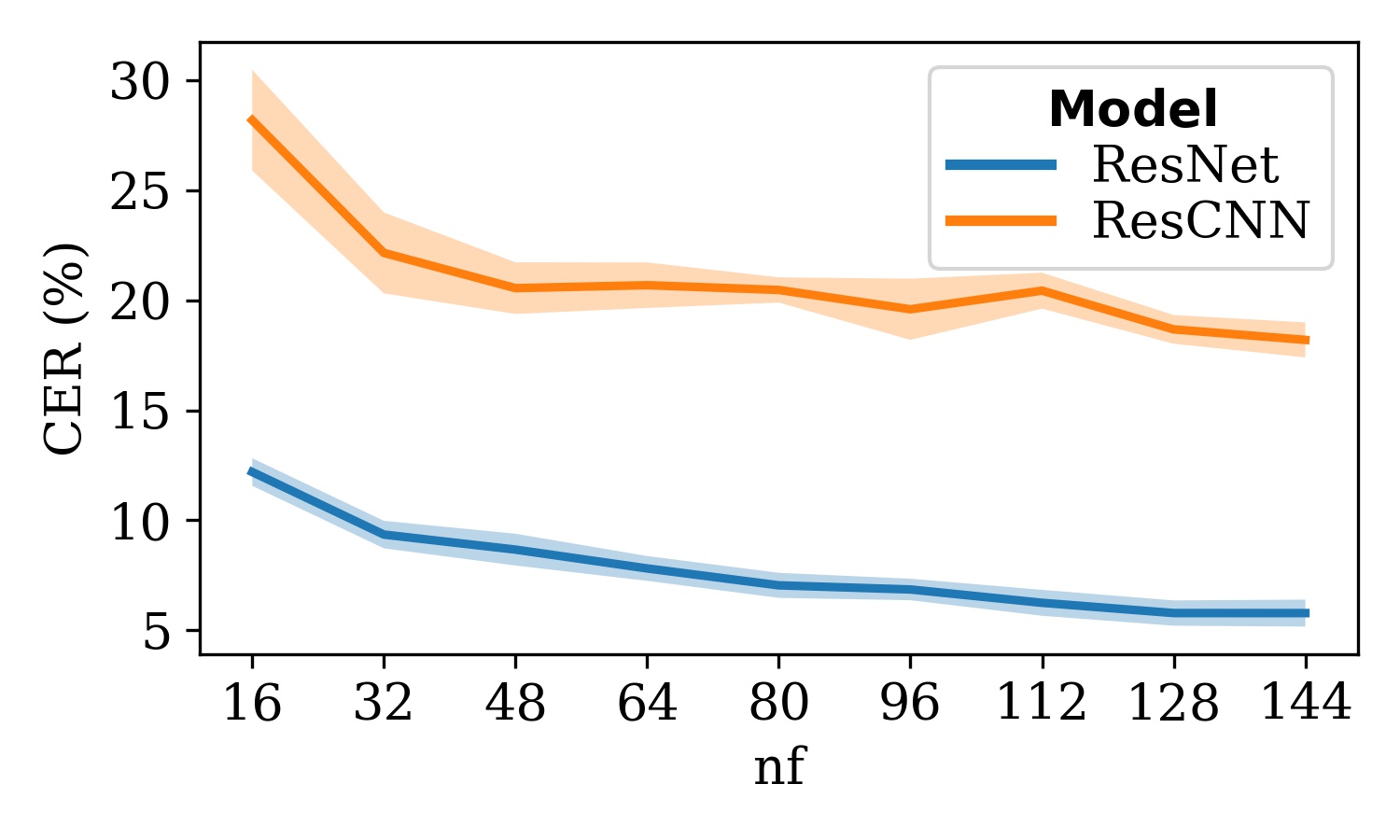}
        \vspace{-0.4cm}
        \subcaption{WD.}
    \end{minipage}
    \hfill
	\begin{minipage}[b]{0.245\linewidth}
        \centering
        \includegraphics[width=1.0\linewidth]{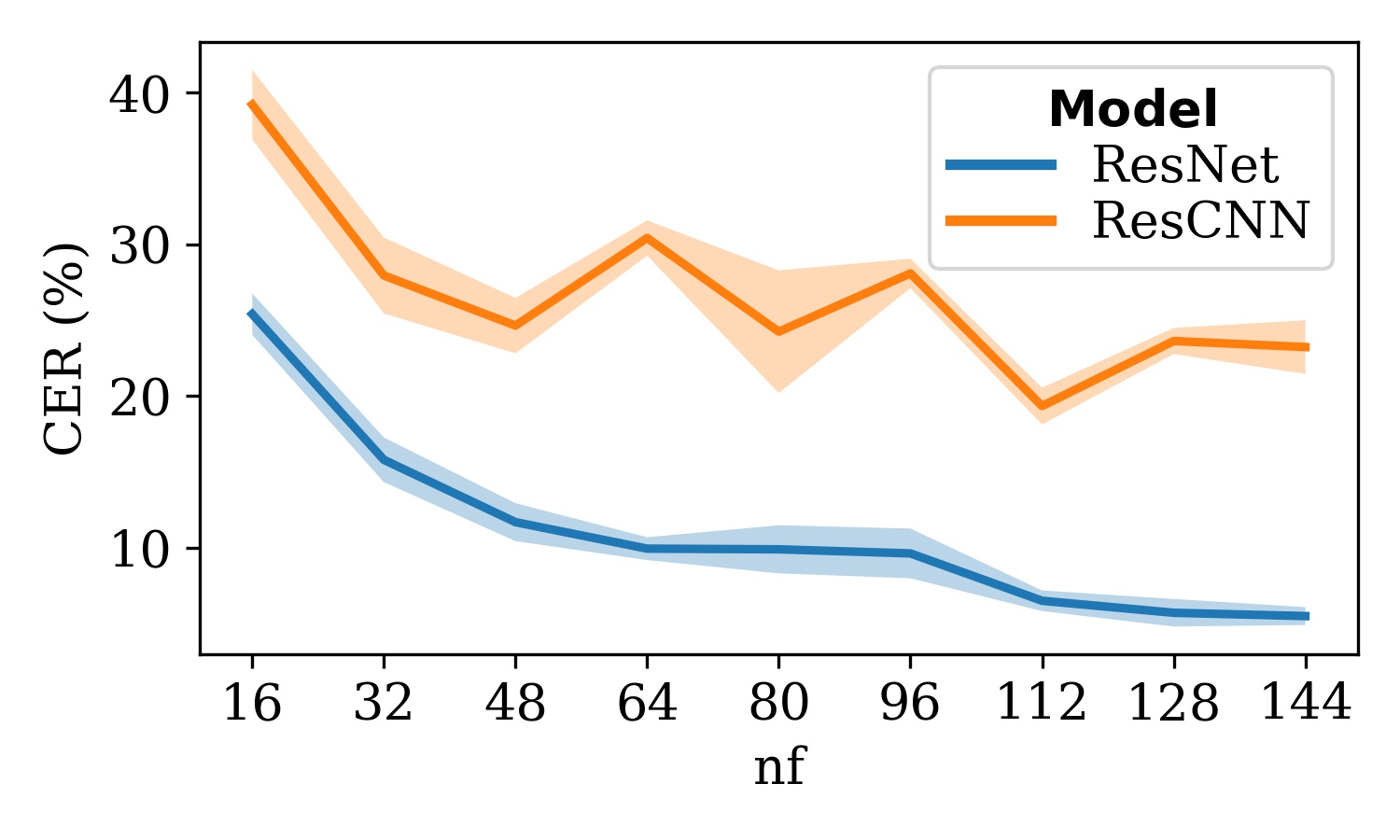}
        \vspace{-0.4cm}
        \subcaption{WD: BiLSTM.}
    \end{minipage}
    \hfill
	\begin{minipage}[b]{0.245\linewidth}
        \centering
        \includegraphics[width=1.0\linewidth]{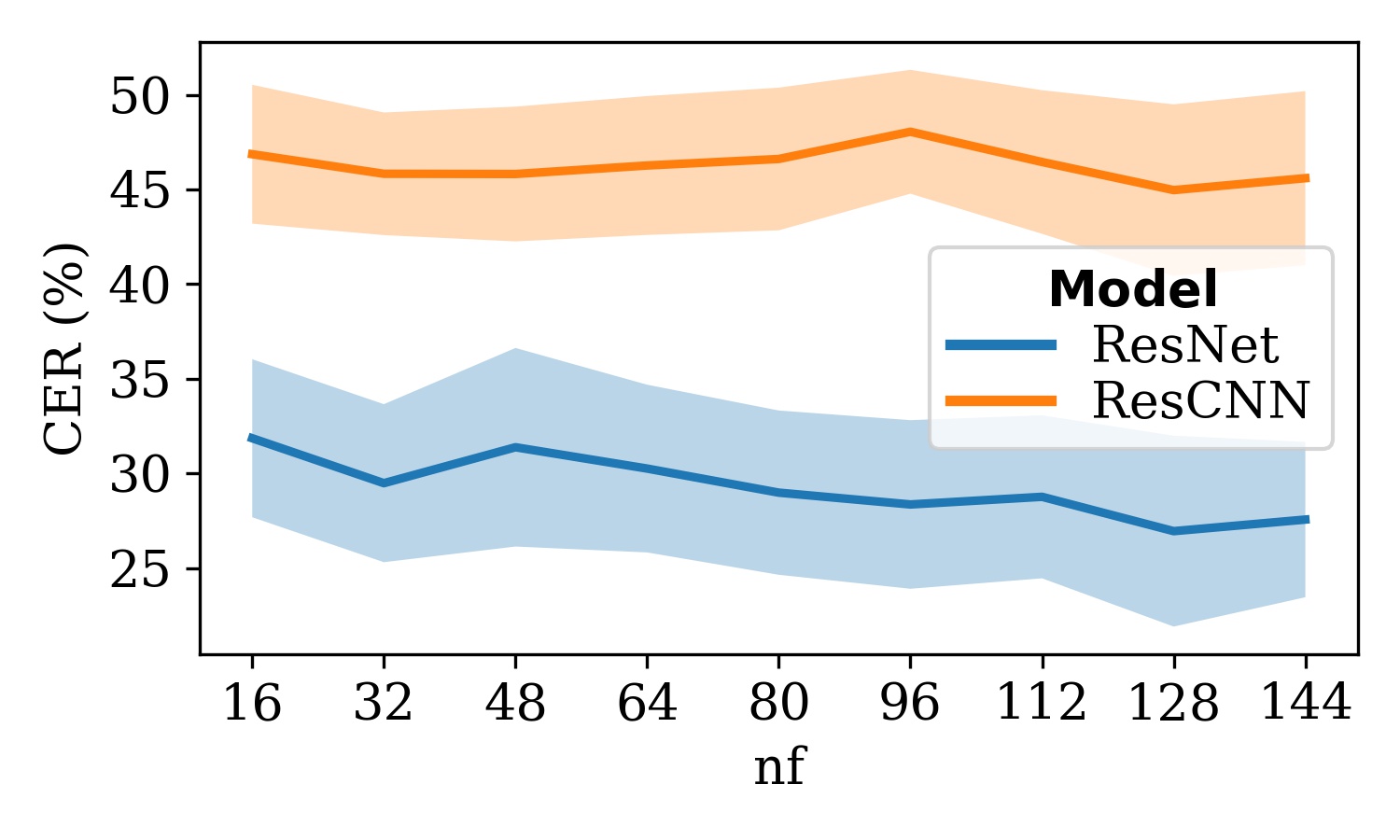}
        \vspace{-0.4cm}
        \subcaption{WI.}
    \end{minipage}
    \hfill
	\begin{minipage}[b]{0.245\linewidth}
        \centering
        \includegraphics[width=1.0\linewidth]{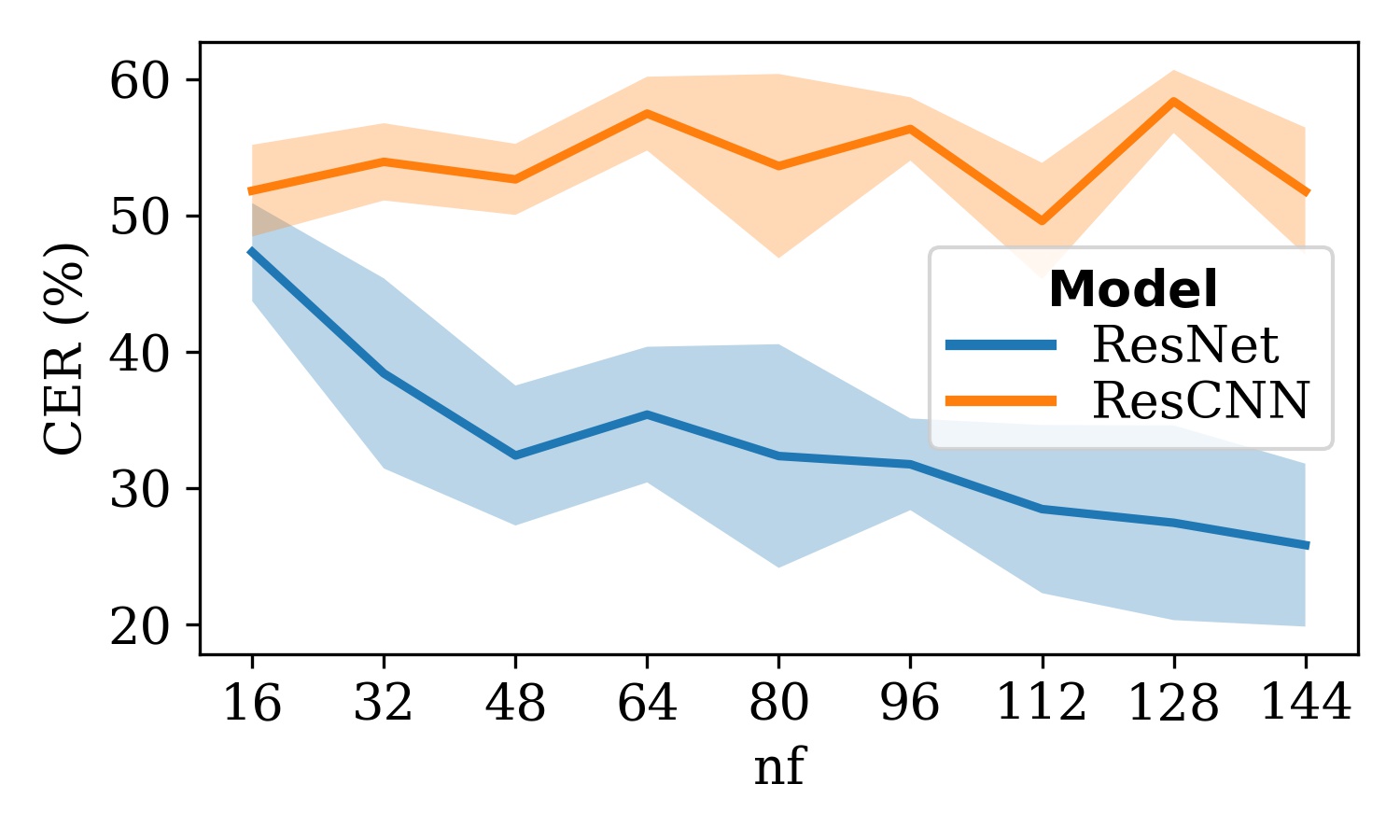}
        \vspace{-0.4cm}
        \subcaption{WI: BiLSTM.}
    \end{minipage}
    \vspace{-0.6cm}
    \caption{Hyperparameter search of $nf$ for ResNet~\cite{wang_2016} and ResCNN~\cite{zou} with and without BiLSTM on the OnHW-equations datasets averaged over 5 splits.}
    \label{image_hps_resnet_rescnn}
\end{figure*}

\paragraph{Comparison to State-of-the-art Techniques.} For comparison, we train nine different well-established time-series classification architectures on our OnHW datasets and InceptionTime~\cite{Fawaz} on the tablet datasets. For these methods we interpolate and zero pad the time-series to 800 timesteps to obtain a fixed sequence length. We use linear spline interpolation. 800 timesteps leads to a low CER (see Figure~\ref{figure_interpolation}), while above 800 timesteps the training time significantly increases. As these methods are introduced for classifying single labels (not sequences of labels), we replace the last linear layer with a max pooling layer (of kernel size 4), a dropout layer (40\%), and an 1D convolutional layer (kernel size 1 and channels are the number of class labels). Similar to our approaches, we further add two BiLSTM layers each of size 60. InceptionTime is an ensemble of CNNs inspired by Inception-v4. As its default parameters (\textit{nf} of 32 and \textit{depth} of 6) lead to inferior performance compared to our methods, we perform a large hyperparameter search for \textit{depth} (between 3 and 12) and \textit{nf} (16, 32, 64, 96 and 128) with and without BiLSTMs for the WD and WI tasks (see Figure~\ref{image_hps}). On the WD dataset, a higher \textit{nf} and greater \textit{depth} leads to a lower CER. For the WI task, the model tends to overfit on specific writers for larger models, and hence, the error rates are constant for \textit{nf} between 64 and 128, while the CER still decreases for a greater \textit{depth}. For \textit{nf} of 32 and \textit{depth} of 11, InceptionTime+BiLSTM can marginally outperform our CNN+BiLSTM model on the OnHW-equations dataset (1.65\% CER WD and 11.28\% CER WI) and is notably better on the ONHW-words500 (WD) dataset (12.96\% CER) without the two additional BiLSTM layers, but is on par with our CNN+BiLSTM model on the WI task (26.08\% CER) and yields marginally higher error rates on the random splits. %For the OnHW-wordsRandom dataset, InceptionTime+BiLSTM is on par with the CNN+BiLSTM model for the WD task, and yields a better CER of 33.31\% but an higher WER of 85.42\% on the WI task. 
Results further suggest that the performance strongly depends on the network size. XceptionTime~\cite{rahimian} consists of depthwise separable convolutions and adaptive average pooling to capture both temporal and spatial contents. We search for the hyperparameter \textit{hf} (see Figure~\ref{image_hps_xceptionTime}) and set $nf = 144$. The small FCN~\cite{wang_2016} model yields high error rates, but ResNet~\cite{wang_2016} (based on FCN) enables the exploitation of class activation maps to find contributing regions in the raw data and improves FCN. ResCNN~\cite{zou} integrates residual networks with CNNs. We set also $nf = 144$ for ResCNN and RestNet (see Figure~\ref{image_hps_resnet_rescnn}), which perform similar, but cannot outperform XceptionTime on our datasets. While additional BiLSTM layers improve the results of InceptionTime, the error rates for XceptionTime, ResNet and ResCNN decrease with additional BiLSTM layers. The univariate models LSTM-FCN~\cite{karim} and GRU-FCN~\cite{elsayed} as well as the multivariate models MLSTM-FCN~\cite{mlstm_fcn} and MGRU-FCN~\cite{mlstm_fcn} that augment the fully convolutional block with a squeeze-and-excitation block improve the FCN results, but are not complex enough to outperform other architectures on our datasets. In general, word beam search \cite{scheidl} did not improve results and even leads to degraded performance. See Appendix~\ref{sec_appendix_detailed_evaluation} for more evaluation details and a comparison to state-of-the-art techniques.

\begin{table*}[t!]
\begin{center}
\setlength{\tabcolsep}{3.5pt}
    \caption{Evaluation results (WER, CER) in \% (mean and standard deviation over 5-fold splits) for different augmentation techniques and sensor choices for the OnHW-equations dataset. \textbf{Bold} are \underline{baseline} improvements.}
    \label{table_results2}
    \small \begin{tabular}{ p{3.7cm} | p{3.5cm} | p{0.5cm} | p{0.5cm} | p{0.5cm} | p{0.5cm} | p{0.5cm} | p{0.5cm} | p{0.5cm} | p{0.5cm}}
    & & \multicolumn{4}{c|}{\textbf{WD}} & \multicolumn{4}{c}{\textbf{WI}} \\
    \multicolumn{1}{c|}{\textbf{Augmentation}} & & \multicolumn{2}{c|}{\textbf{WER}} & \multicolumn{2}{c|}{\textbf{CER}} & \multicolumn{2}{c|}{\textbf{WER}} & \multicolumn{2}{c}{\textbf{CER}} \\
    \multicolumn{1}{c|}{\textbf{Technique}} & \multicolumn{1}{c|}{\textbf{Sensors}} & \multicolumn{1}{c}{\textbf{Mean}} & \multicolumn{1}{c|}{\textbf{STD}} & \multicolumn{1}{c}{\textbf{Mean}} & \multicolumn{1}{c|}{\textbf{STD}} & \multicolumn{1}{c}{\textbf{Mean}} & \multicolumn{1}{c|}{\textbf{STD}} & \multicolumn{1}{c}{\textbf{Mean}} & \multicolumn{1}{c}{\textbf{STD}} \\ \hline
    None & All & \multicolumn{1}{r}{\underline{22.96}} & \multicolumn{1}{r|}{1.83} & \multicolumn{1}{r}{\underline{3.50}} & \multicolumn{1}{r|}{0.38} & \multicolumn{1}{r}{\underline{69.21}} & \multicolumn{1}{r|}{7.91} & \multicolumn{1}{r}{\underline{18.11}} & \multicolumn{1}{r}{5.20} \\
    Scaling (S) & All & \multicolumn{1}{r}{\textbf{22.70}} & \multicolumn{1}{r|}{0.40} & \multicolumn{1}{r}{\textbf{3.43}} & \multicolumn{1}{r|}{0.22} & \multicolumn{1}{r}{69.70} & \multicolumn{1}{r|}{7.90} & \multicolumn{1}{r}{18.80} & \multicolumn{1}{r}{5.84} \\
    Time Warping (TW) & All & \multicolumn{1}{r}{\textbf{20.90}} & \multicolumn{1}{r|}{0.83} & \multicolumn{1}{r}{\textbf{3.18}} & \multicolumn{1}{r|}{0.27} & \multicolumn{1}{r}{\textbf{64.10}} & \multicolumn{1}{r|}{5.51} & \multicolumn{1}{r}{\textbf{15.26}} & \multicolumn{1}{r}{2.27} \\
    Jittering (J) & All & \multicolumn{1}{r}{\textbf{22.87}} & \multicolumn{1}{r|}{0.75} & \multicolumn{1}{r}{\textbf{3.47}} & \multicolumn{1}{r|}{0.33} & \multicolumn{1}{r}{\textbf{68.14}} & \multicolumn{1}{r|}{10.03} & \multicolumn{1}{r}{18.68} & \multicolumn{1}{r}{7.18} \\
    Magnitude Warping (MW) & All & \multicolumn{1}{r}{\textbf{22.88}} & \multicolumn{1}{r|}{1.21} & \multicolumn{1}{r}{3.53} & \multicolumn{1}{r|}{0.29} & \multicolumn{1}{r}{76.80} & \multicolumn{1}{r|}{8.35} & \multicolumn{1}{r}{18.47} & \multicolumn{1}{r}{5.21} \\
    Shifting (SH) & All & \multicolumn{1}{r}{\textbf{22.40}} & \multicolumn{1}{r|}{1.12} & \multicolumn{1}{r}{\textbf{3.43}} & \multicolumn{1}{r|}{0.24} & \multicolumn{1}{r}{69.81} & \multicolumn{1}{r|}{7.59} & \multicolumn{1}{r}{18.80} & \multicolumn{1}{r}{4.88} \\ \hline
    Interpolation & All & \multicolumn{1}{r}{25.04} & \multicolumn{1}{r|}{0.92} & \multicolumn{1}{r}{3.96} & \multicolumn{1}{r|}{0.32} & \multicolumn{1}{r}{70.50} & \multicolumn{1}{r|}{8.30} & \multicolumn{1}{r}{19.42} & \multicolumn{1}{r}{5.96} \\
    Normalization & All & \multicolumn{1}{r}{55.26} & \multicolumn{1}{r|}{2.04} & \multicolumn{1}{r}{7.97} & \multicolumn{1}{r|}{0.51} & \multicolumn{1}{r}{82.48} & \multicolumn{1}{r|}{8.74} & \multicolumn{1}{r}{22.71} & \multicolumn{1}{r}{5.04} \\ \hline
    None & w/o Mag. & \multicolumn{1}{r}{\textbf{22.60}} & \multicolumn{1}{r|}{1.51} & \multicolumn{1}{r}{\textbf{3.44}} & \multicolumn{1}{r|}{0.36} & \multicolumn{1}{r}{\textbf{63.48}} & \multicolumn{1}{r|}{8.32} & \multicolumn{1}{r}{\textbf{16.07}} & \multicolumn{1}{r}{4.73} \\
    None & w/o Front Acc. & \multicolumn{1}{r}{\textbf{21.36}} & \multicolumn{1}{r|}{0.60} & \multicolumn{1}{r}{\textbf{3.28}} & \multicolumn{1}{r|}{0.29} & \multicolumn{1}{r}{70.24} & \multicolumn{1}{r|}{8.25} & \multicolumn{1}{r}{19.55} & \multicolumn{1}{r}{5.52} \\
    None & w/o Rear Acc. & \multicolumn{1}{r}{23.20} & \multicolumn{1}{r|}{0.86} & \multicolumn{1}{r}{3.57} & \multicolumn{1}{r|}{0.26} & \multicolumn{1}{r}{\textbf{68.30}} & \multicolumn{1}{r|}{8.14} & \multicolumn{1}{r}{\textbf{16.64}} & \multicolumn{1}{r}{5.40} \\
    None & w/o Mag., w/o Front Acc. & \multicolumn{1}{r}{\textbf{22.46}} & \multicolumn{1}{r|}{1.55} & \multicolumn{1}{r}{\textbf{3.41}} & \multicolumn{1}{r|}{0.38} & \multicolumn{1}{r}{\textbf{69.12}} & \multicolumn{1}{r|}{8.40} & \multicolumn{1}{r}{\textbf{17.31}} & \multicolumn{1}{r}{4.02} \\
    \end{tabular}
\end{center}
\end{table*}

\begin{figure}[t!]
	\centering
    \includegraphics[width=1.0\linewidth]{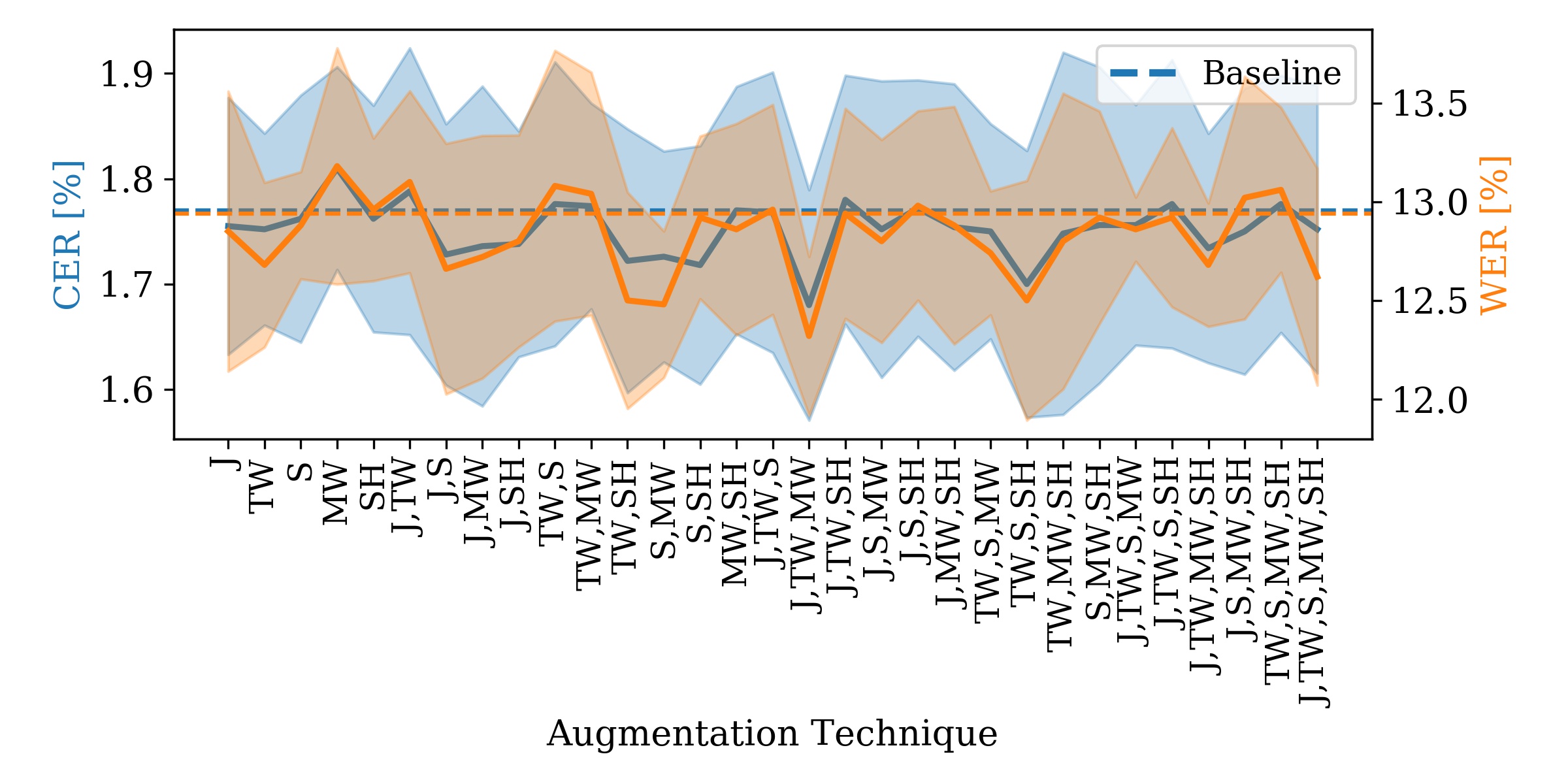}
    \caption{Augmentation results for InceptionTime on the OnHW-equations (WD) dataset over five splits.}
    \label{image_augmentation_inceptiontime_wd}
\end{figure}

\paragraph{Influence of Data Augmentation.} We train the CNN+ LSTM model on the OnHW-equations dataset with the augmentation techniques described in Section~\ref{chap_method}. Results are given in Table~\ref{table_results2}. The baseline WER of 22.96\% (WD) can be improved with all augmentation techniques, while the WI error of 69.21\% is only affected by \textit{time warping} and \textit{jittering}. The most notable improvement is given by \textit{time warping} with 20.90\% for the WD task and 64.10\% for the WI task. Interpolation to 1,000 timesteps did not improve the accuracy, and normalization to $[-1, 1]$ deteriorates training performance. Figure~\ref{image_augmentation_inceptiontime_wd} shows augmentation results and combinations of these for InceptionTime on the OnHW-equations WD dataset. Here, the baseline CER of 1.77\% and WER of 12.94\% can be notably improved by time warping as a single augmentation (comparable to our CNN+LSTM). The combination of jittering, and time and magnitude warping yields the highest error rate reduction.

\paragraph{Influence of Sensor Dropping.} We train the OnHW-equations dataset and drop data from single sensors, e.g., the front or rear accelerometer or the magnetometer data, in order to evaluate the influence of each sensor, see Table~\ref{table_results2}. Only dropping the front accelerometer (WD) and the rear accelerometer (WI) decreases the WER and CER, which could also be attributed to the smaller dataset size while leaving the architecture unchanged. Without magnetometer the WER improves for the WI task as the magnetic field changes with the recording location, but keeps constant for the same writer. Dropping the force sensor leads to a significant higher classification error as the force sensor provides information that allows a segmentation of strokes.

\subsection{Single Character Task Evaluation}
\label{chap_eval_character}

\begin{table*}[t!]
\begin{center}
\setlength{\tabcolsep}{3.5pt}
    \caption{Recognition rates (CRR) in \% for the symbols, split equations and characters WD and WI datasets. \underline{Underlined}: State-of-the-art results. \textbf{Bold}: Best results. SE: squeeze-and-excitation. Att.: attentional LSTM.}
    \label{table_results3}
    \small
\begin{tabular}{ p{1.1cm} | p{0.5cm} | p{0.5cm} | p{0.5cm} | p{0.5cm} | p{0.5cm} | p{0.5cm} | p{0.5cm} | p{0.5cm} | p{0.5cm} | p{0.5cm} | p{0.5cm} | p{0.5cm} | p{0.5cm} }
    \multicolumn{2}{c|}{\textbf{Method}} & \multicolumn{2}{c|}{\textbf{OnHW-}} & \multicolumn{2}{c|}{\textbf{OnHW-}} & \multicolumn{2}{c|}{\textbf{OnHW-sym.$^1$}} & \multicolumn{6}{c}{\textbf{OnHW-chars$^3$~\cite{ott}}} \\
    \multicolumn{2}{c|}{($\mathcal{L}_{CCE}$ loss)} & \multicolumn{2}{c|}{\textbf{symbols$^1$}} & \multicolumn{2}{c|}{\textbf{equations$^{1,2}$}} & \multicolumn{2}{c|}{\textbf{+ equations$^{1,2}$}} & \multicolumn{2}{c|}{\textbf{lower}} & \multicolumn{2}{c|}{\textbf{upper}} & \multicolumn{2}{c}{\textbf{combined}} \\
    \multicolumn{2}{c|}{} & \multicolumn{1}{c}{\textbf{WD}} & \multicolumn{1}{c|}{\textbf{WI}} & \multicolumn{1}{c}{\textbf{WD}} & \multicolumn{1}{c|}{\textbf{WI}} & \multicolumn{1}{c}{\textbf{WD}} & \multicolumn{1}{c|}{\textbf{WI}} & \multicolumn{1}{c}{\textbf{WD}} & \multicolumn{1}{c|}{\textbf{WI}}  & \multicolumn{1}{c}{\textbf{WD}} & \multicolumn{1}{c|}{\textbf{WI}}  & \multicolumn{1}{c}{\textbf{WD}} & \multicolumn{1}{c}{\textbf{WI}} \\ \hline
    \multicolumn{2}{l|}{CNN+LSTM} & \multicolumn{1}{r}{\textbf{96.44}} & \multicolumn{1}{r|}{\textbf{80.00}} & \multicolumn{1}{r}{95.43} & \multicolumn{1}{r|}{84.22} & \multicolumn{1}{r}{\textbf{95.65}} & \multicolumn{1}{r|}{85.11} & \multicolumn{1}{r}{88.85} & \multicolumn{1}{r|}{79.48} & \multicolumn{1}{r}{92.15} & \multicolumn{1}{r|}{85.60} & \multicolumn{1}{r}{78.17} & \multicolumn{1}{r}{68.06} \\
    \multicolumn{2}{l|}{CNN+BiLSTM} & \multicolumn{1}{r}{96.20} & \multicolumn{1}{r|}{79.51} & \multicolumn{1}{r}{95.70} & \multicolumn{1}{r|}{83.88} & \multicolumn{1}{r}{95.50} & \multicolumn{1}{r|}{84.55} & \multicolumn{1}{r}{\textbf{89.66}} & \multicolumn{1}{r|}{\textbf{80.00}} & \multicolumn{1}{r}{\textbf{92.58}} & \multicolumn{1}{r|}{\textbf{85.64}} & \multicolumn{1}{r}{\textbf{78.98}} & \multicolumn{1}{r}{\textbf{68.44}} \\
\end{tabular}
\end{center}
\end{table*}

\begin{table*}[t!]
\begin{center}
\setlength{\tabcolsep}{3.1pt}
    \caption{Recognition rates (CRR) in \% for the symbols, split equations and characters WD and WI datasets for the CNN+BiLSTM architecture trained with different loss functions. \textbf{Bold}: Best results}.
    \label{table_results_losses}
    \small \begin{tabular}{ p{1.1cm} | p{0.5cm} | p{0.5cm} | p{0.5cm} | p{0.5cm} | p{0.5cm} | p{0.5cm} | p{0.5cm} | p{0.5cm} | p{0.5cm} | p{0.5cm} | p{0.5cm} | p{0.5cm} | p{0.5cm} }
    \multicolumn{2}{c|}{\textbf{Loss Function}} & \multicolumn{2}{c|}{\textbf{OnHW-}} & \multicolumn{2}{c|}{\textbf{OnHW-}} & \multicolumn{2}{c|}{\textbf{OnHW-sym.$^1$}} & \multicolumn{6}{c}{\textbf{OnHW-chars$^3$~\cite{ott}}} \\
    \multicolumn{2}{c|}{(CNN+BiLSTM architecture)} & \multicolumn{2}{c|}{\textbf{symbols$^1$}} & \multicolumn{2}{c|}{\textbf{equations$^{1,2}$}} & \multicolumn{2}{c|}{\textbf{+ equations$^{1,2}$}} & \multicolumn{2}{c|}{\textbf{lower}} & \multicolumn{2}{c|}{\textbf{upper}} & \multicolumn{2}{c}{\textbf{combined}} \\
    \multicolumn{1}{c}{} & & \multicolumn{1}{c}{\textbf{WD}} & \multicolumn{1}{c|}{\textbf{WI}} & \multicolumn{1}{c}{\textbf{WD}} & \multicolumn{1}{c|}{\textbf{WI}} & \multicolumn{1}{c}{\textbf{WD}} & \multicolumn{1}{c|}{\textbf{WI}} & \multicolumn{1}{c}{\textbf{WD}} & \multicolumn{1}{c|}{\textbf{WI}}  & \multicolumn{1}{c}{\textbf{WD}} & \multicolumn{1}{c|}{\textbf{WI}}  & \multicolumn{1}{c}{\textbf{WD}} & \multicolumn{1}{c}{\textbf{WI}} \\ \hline
    \multicolumn{2}{l|}{Categorical CE (CCE)} & \multicolumn{1}{r}{96.20} & \multicolumn{1}{r|}{79.51} & \multicolumn{1}{r}{95.57} & \multicolumn{1}{r|}{83.88} & \multicolumn{1}{r}{95.50} & \multicolumn{1}{r|}{84.55} & \multicolumn{1}{r}{89.66} & \multicolumn{1}{r|}{80.00} & \multicolumn{1}{r}{92.58} & \multicolumn{1}{r|}{85.64} & \multicolumn{1}{r}{78.98} & \multicolumn{1}{r}{68.44} \\
    \multicolumn{2}{l|}{Focal loss (FL) \cite{lin_goyal}} & \multicolumn{1}{r}{95.78} & \multicolumn{1}{r|}{79.67} & \multicolumn{1}{r}{95.42} & \multicolumn{1}{r|}{84.53} & \multicolumn{1}{r}{95.25} & \multicolumn{1}{r|}{85.20} & \multicolumn{1}{r}{88.56} & \multicolumn{1}{r|}{78.88} & \multicolumn{1}{r}{91.91} & \multicolumn{1}{r|}{85.62} & \multicolumn{1}{r}{77.48} & \multicolumn{1}{r}{68.15} \\
\end{tabular}
\end{center}
\end{table*}

\paragraph{Method and Architecture Evaluation.} We use our OnHW-symbols and split OnHW-equations datasets, the combination of both (samples randomly shuffled) and the OnHW-chars~\cite{ott} dataset, and interpolate the single characters to the longest single character of the dataset (64 for characters and 79 for number/symbols). We train our network proposed in Figure~\ref{figure_architecture1} with one additional dense layer of 100 units. For all methods we use the categorical CE loss for training and the CRR for evaluation. Network parameter choices are described in Appendix~\ref{sec_appendix_trans_parameters}. The results are summarized in Table~\ref{table_results3}. We also compare to state-of-the-art results provided in \cite{ott}. While GRU~\cite{chung} yields very low accuracies for all datasets, standard LSTM units (2 and 3 stacked layers), BiLSTM units and TCNs can increase the CRR. Further, FCN~\cite{mlstm_fcn} and the spatio-temporal variants RNN-FCN~\cite{mlstm_fcn}, LSTM-FCN~\cite{karim} and GRU-FCN~\cite{elsayed} as well as the multivariate variants MRNN-FCN, MLSTM-FCN~\cite{mlstm_fcn} and MGRU-FCN~\cite{mlstm_fcn} yield better results. MLSTM-FCN~\cite{mlstm_fcn} with a standard or attention-based LSTM and with/out a squeeze-and-excitation (SE) block achieves high accuracies, but cannot improve over state-of-the-art results achieved by \cite{ott}. Due to minor and inconsistent changes in performance, it is not possible to make a statement about the importance of the SE block and the attention-based LSTM. The networks based on CNNs, i.e., ResCNN~\cite{zou}, ResNet~\cite{wang_2016}, XResNet~\cite{he_zhang}, InceptionTime~\cite{Fawaz} and XceptionTime~\cite{rahimian} can partly outperform the FCN variants. For XResNet, a smaller depth of the network is preferable, while for InceptionTime the greater \textit{depth} and larger \textit{nf} generally yields better results. We train TapNet~\cite{tapnet}, an attentional prototypical network for semi-supervised learning, that achieves the lowest accuracies. We propose a benchmark for the Transformer variants \cite{choromanski,jaegle,kitaev,tay_bahri,wang_li} (for details, see Appendix~\ref{sec_appendix_trans_parameters}). The performance improves for all Transformer variants compared to TapNet, but are notably lower than those of the convolutional and spatio-temporal methods. TST~\cite{zerveas} with Gaussian encoding is on par with the convolutional techniques on the WD datasets. While our CNN+BiLSTM outperforms all methods on all OnHW-chars~\cite{ott} datasets, it is not notably different from results achieved by the CNN+LSTM and CNN+TCN architectures, which in turn achieve the best results on the OnHW-symbols and split OnHW-equations datasets as well as on the combined cases.

\begin{table*}[t!]
\begin{center}
\setlength{\tabcolsep}{2.5pt}
    \caption{Evaluation results (WER, CER) in \% (mean and standard deviation) for our left-handed OnHW-equations-L, OnHW-words500-L and OnHW-wordsRandom-L datasets (left), and recognition results (CRR) in \% for our left-handed OnHW-symbols-L, split OnHW-equations-L and OnHW-chars-L datasets (right) for the CNN+BiLSTM architecture. Writer-dependent (WD) and writer-independent (WI) classification tasks.}
    \label{table_results_left}
    \vspace{-0.1cm}
    \small \begin{tabular}{ p{1.1cm} | p{0.5cm} | p{0.5cm} | p{0.5cm} | p{0.5cm} | p{0.5cm} | p{0.5cm} | p{0.5cm} | p{0.5cm} | p{0.5cm} | p{0.5cm} | p{0.5cm} | p{0.5cm} | p{0.5cm} | p{0.5cm} | p{0.5cm} }
    \multicolumn{1}{c|}{\textbf{Dataset}} & \multicolumn{4}{c|}{\textbf{WD}} & \multicolumn{4}{c}{\textbf{WI}} & \multicolumn{1}{c}{} & \multicolumn{2}{c|}{\textbf{Dataset}} & \multicolumn{1}{c|}{\textbf{WD}} & \multicolumn{1}{c}{\textbf{WI}} \\
    \multicolumn{1}{c|}{(CNN+BiLSTM} & \multicolumn{2}{c|}{\textbf{WER}} & \multicolumn{2}{c|}{\textbf{CER}} & \multicolumn{2}{c|}{\textbf{WER}} & \multicolumn{2}{c}{\textbf{CER}} & \multicolumn{1}{c}{} & \multicolumn{2}{c|}{(CNN+BiLSTM} & \multicolumn{1}{c|}{\textbf{CRR}} & \multicolumn{1}{c}{\textbf{CRR}} \\
    \multicolumn{1}{c|}{architecture)} & \multicolumn{1}{r}{\textbf{Mean}} & \multicolumn{1}{r|}{\textbf{STD}} & \multicolumn{1}{r}{\textbf{Mean}} & \multicolumn{1}{r|}{\textbf{STD}} & \multicolumn{1}{r}{\textbf{Mean}} & \multicolumn{1}{r|}{\textbf{STD}} & \multicolumn{1}{r}{\textbf{Mean}} & \multicolumn{1}{r}{\textbf{STD}} & \multicolumn{1}{c}{} & \multicolumn{2}{c|}{architecture)} & \multicolumn{1}{c|}{\textbf{Mean}} & \multicolumn{1}{c}{\textbf{Mean}} \\ \cline{1-9} \cline{11-14}
    \multicolumn{1}{l|}{OnHW-equations-L} & \multicolumn{1}{r}{8.56} & \multicolumn{1}{r|}{1.59} & \multicolumn{1}{r}{1.24} & \multicolumn{1}{r|}{0.25} & \multicolumn{1}{r}{95.73} & \multicolumn{1}{r|}{3.13} & \multicolumn{1}{r}{32.16} & \multicolumn{1}{r}{5.16} & \multicolumn{1}{c}{} & \multicolumn{2}{l|}{OnHW-symbols-L$^1$} & \multicolumn{1}{r|}{92.00} & \multicolumn{1}{r}{54.00} \\
    \multicolumn{1}{l|}{OnHW-words500-L} & \multicolumn{1}{r}{47.90} & \multicolumn{1}{r|}{17.25} & \multicolumn{1}{r}{15.32} & \multicolumn{1}{r|}{6.03} & \multicolumn{1}{r}{97.90} & \multicolumn{1}{r|}{1.10} & \multicolumn{1}{r}{81.43} & \multicolumn{1}{r}{11.66} & \multicolumn{1}{c}{} & \multicolumn{2}{l|}{OnHW-equations-L$^{1,2}$} & \multicolumn{1}{r|}{92.02} & \multicolumn{1}{r}{51.50} \\
    \multicolumn{1}{l|}{OnHW-wordsRandom-L} & \multicolumn{1}{r}{32.73} & \multicolumn{1}{r|}{3.43} & \multicolumn{1}{r}{5.40} & \multicolumn{1}{r|}{1.15} & \multicolumn{1}{r}{99.70} & \multicolumn{1}{r|}{0.30} & \multicolumn{1}{r}{72.27} & \multicolumn{1}{r}{15.55} & \multicolumn{1}{c}{} & \multicolumn{1}{l|}{OnHW-} & \multicolumn{1}{r|}{lower} & \multicolumn{1}{r|}{94.70} & \multicolumn{1}{c}{-} \\
    \multicolumn{9}{l}{$^1$1-fold cross validation split; samples interpolated to 79 timesteps.} & \multicolumn{1}{c}{} & \multicolumn{1}{l|}{chars-L$^3$~\cite{ott}} & \multicolumn{1}{r|}{upper} & \multicolumn{1}{r|}{91.90} & \multicolumn{1}{c}{-} \\
    \multicolumn{11}{l|}{$^2$Split into single symbols and numbers.} & \multicolumn{1}{r|}{combined} & \multicolumn{1}{r|}{82.80} & \multicolumn{1}{c}{-} \\
    \multicolumn{14}{l}{$^3$5-fold cross validation split; samples interpolated to 64 timesteps.} \\
    \end{tabular}
    %\vspace{-0.2cm}
\end{center}
\end{table*}

\paragraph{Loss Functions Evaluation.} We train the CNN+BiLSTM architecture for all single-based datasets with the CCE loss as baseline and the seven variants described in Section~\ref{chap_method}. For FL, we search for the optimal hyperparameters for the OnHW-chars combined dataset and for the other methods for the OnHW-symbols dataset (see Appendix~\ref{sec_appendix_trans_parameters}). We set $\alpha = 0.75$ and $\gamma = 8$. From the hyperparameter searches and literature recommendation, we set $\beta=0.1$ for LSR, $\beta=0.95$ for SBS, $\beta=0.8$ for HBS, and $\alpha=0.95$ for GCE. For the SCE loss, we set $\alpha=0.5$ and $\beta=0.5$ for the weighting of the CCE and RCE losses, respectively. Similar, the regularization terms of the JO loss are weighted by $\alpha=1.2$ and $\beta=0.8$. Table~\ref{table_results_losses} gives an overview of the results for all loss functions for all single character-based datasets. The FL improves the CRR results of the symbols and equations datasets (WI) in comparison to the baseline, but yields worse results for the other datasets. As characters in the OnHW-chars dataset are equally distributed, the FL does not have any benefit on training performance. LSR prevents overconfidence and increases the accuracy for all datasets. LSR also achieves the highest accuracy of all losses for eight of the 12 datasets. As there are many samples that are written similarly, the model is overconfident for such samples by integrating a confidence penalty. Similar to FL, the SBS and HBS losses can only marginally improve results for symbols and split equations datasets, and even decrease performances for the character datasets. HBS is slightly better than SBS. The GCE loss decreases the classification accuracy for the OnHW-chars datasets, while it achieves the second best CRR of all losses for the split OnHW-equations WD (95.81\%) and WI (86.46\%) datasets. Yet, the GCE loss often results in \texttt{NaN} loss (see Figure~\ref{image_training_single_chars}, Appendix~\ref{sec_appendix_detailed_evaluation}), and hence, is non-robust for our datasets. The improvement for the SCE loss is less significant than other losses and even decreases for the OnHW-chars dataset. JO leads to an improvement for all OnHW-chars datasets. Joint optimization further outperforms all losses for the WI upper task, and achieves marginally lower accuracies than the LSR loss for the lower and combined datasets. LSR also achieves the highest accuracies on the OnHW-symbols WD (97.33\%) and WI (82.17\%) datasets. In summary, all loss variants can improve results of the CCE loss for the OnHW-symbols, split OnHW-equations and combined datasets as these are not equally distributed. LSR, SCE and JO can most significantly outperform other techniques. For more details of accuracy plots, see Appendix~\ref{sec_appendix_detailed_evaluation}, Figure~\ref{image_training_single_chars}.

\begin{figure*}[t!]
	\centering
	\begin{minipage}[b]{0.32\linewidth}
        \centering
    	\includegraphics[width=1.0\linewidth]{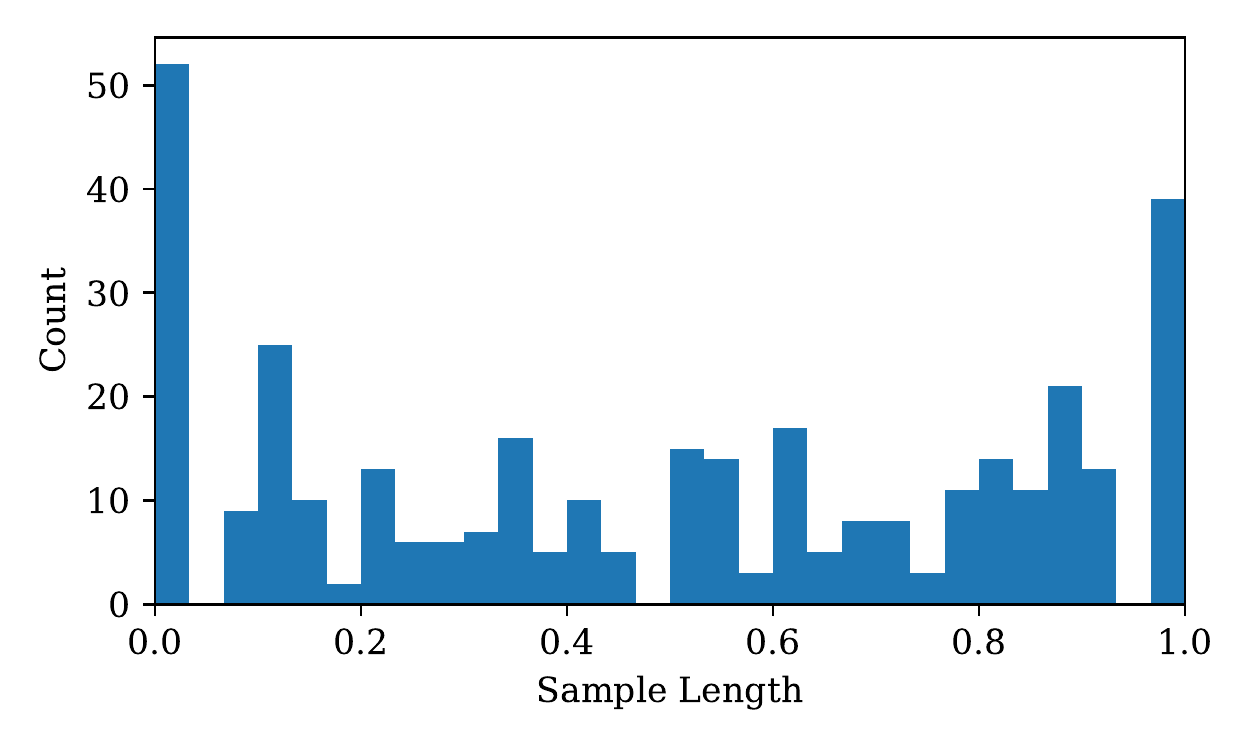}
    	\vspace{-0.4cm}
    	\subcaption{Mismatches.}
    	\label{image_ed_length1_1}
    \end{minipage}
    \hfill
	\begin{minipage}[b]{0.32\linewidth}
        \centering
    	\includegraphics[width=1.0\linewidth]{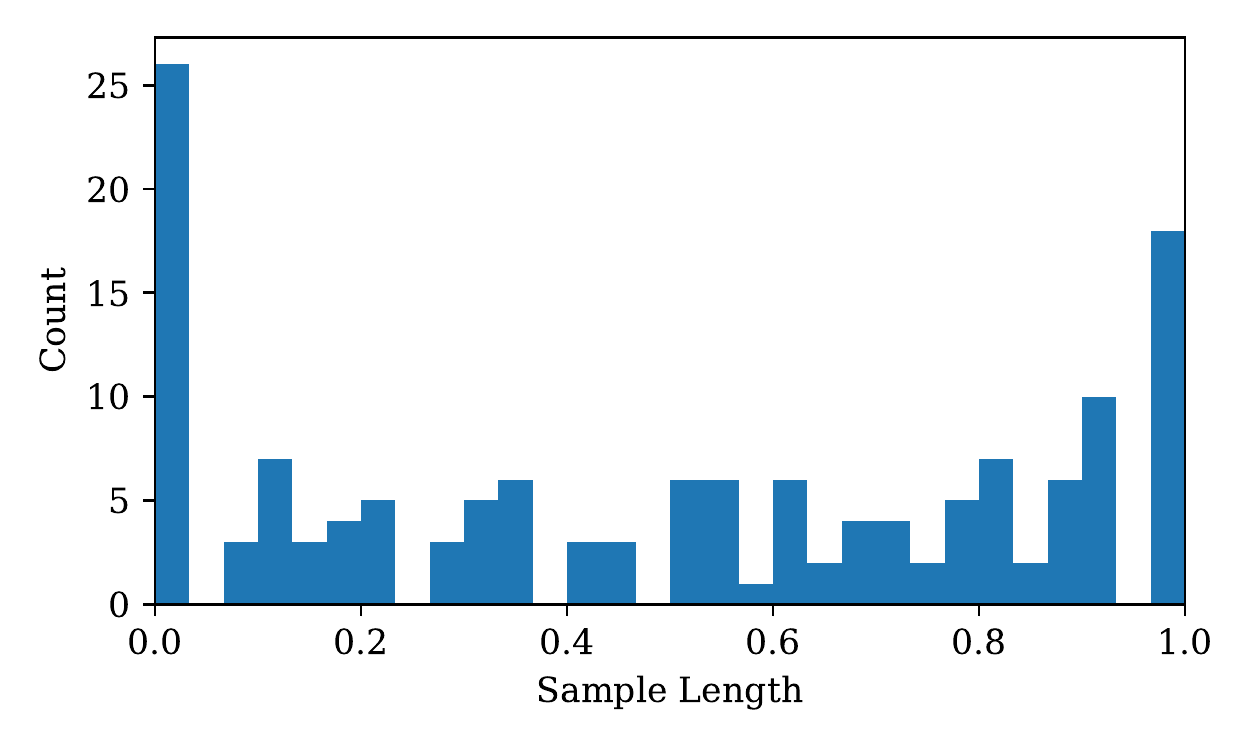}
    	\vspace{-0.4cm}
    	\subcaption{Insertions.}
    	\label{image_ed_length1_2}
    \end{minipage}
    \hfill
	\begin{minipage}[b]{0.32\linewidth}
        \centering
    	\includegraphics[width=1.0\linewidth]{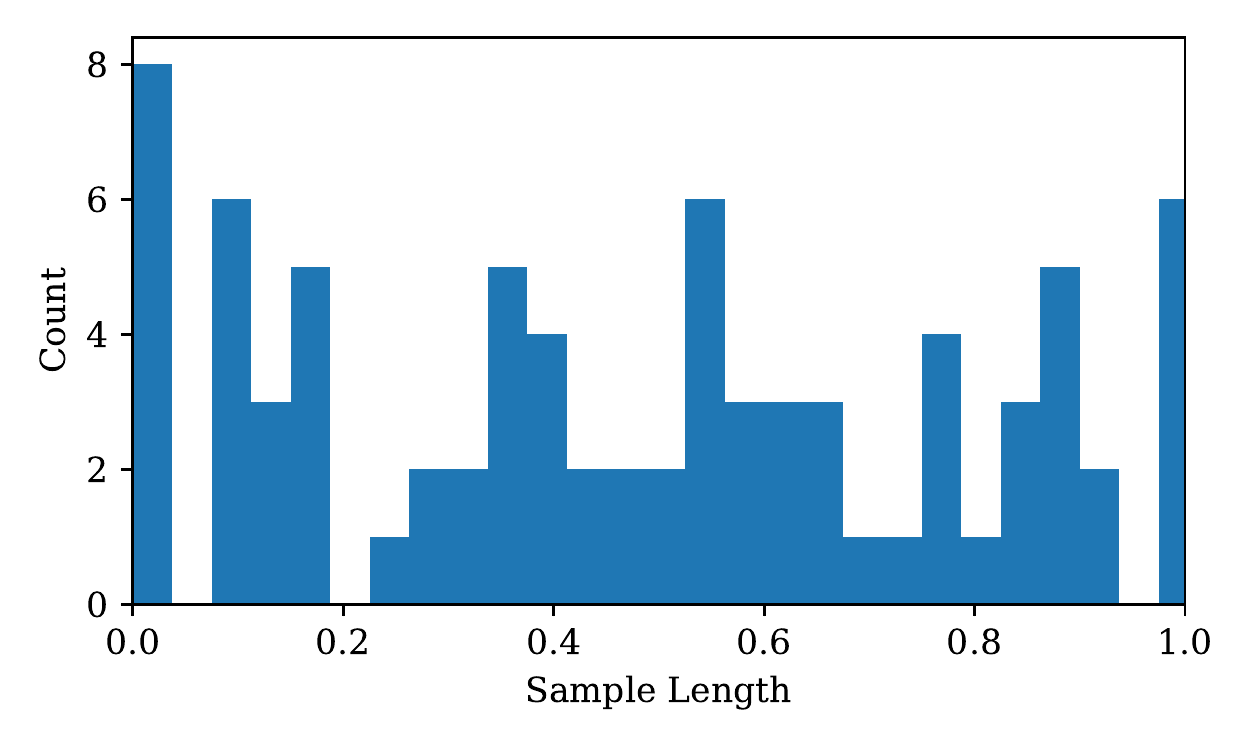}
    	\vspace{-0.4cm}
    	\subcaption{Deletions.}
    	\label{image_ed_length1_3}
    \end{minipage}
    \vspace{-0.1cm}
    \caption{Evaluation of the ED dependent on the normalized sample lengths for the OnHW-equations dataset.}
    \label{image_ed_length1}
\end{figure*}

\begin{figure*}[t!]
	\centering
	\begin{minipage}[b]{0.32\linewidth}
        \centering
    	\includegraphics[width=1.0\linewidth]{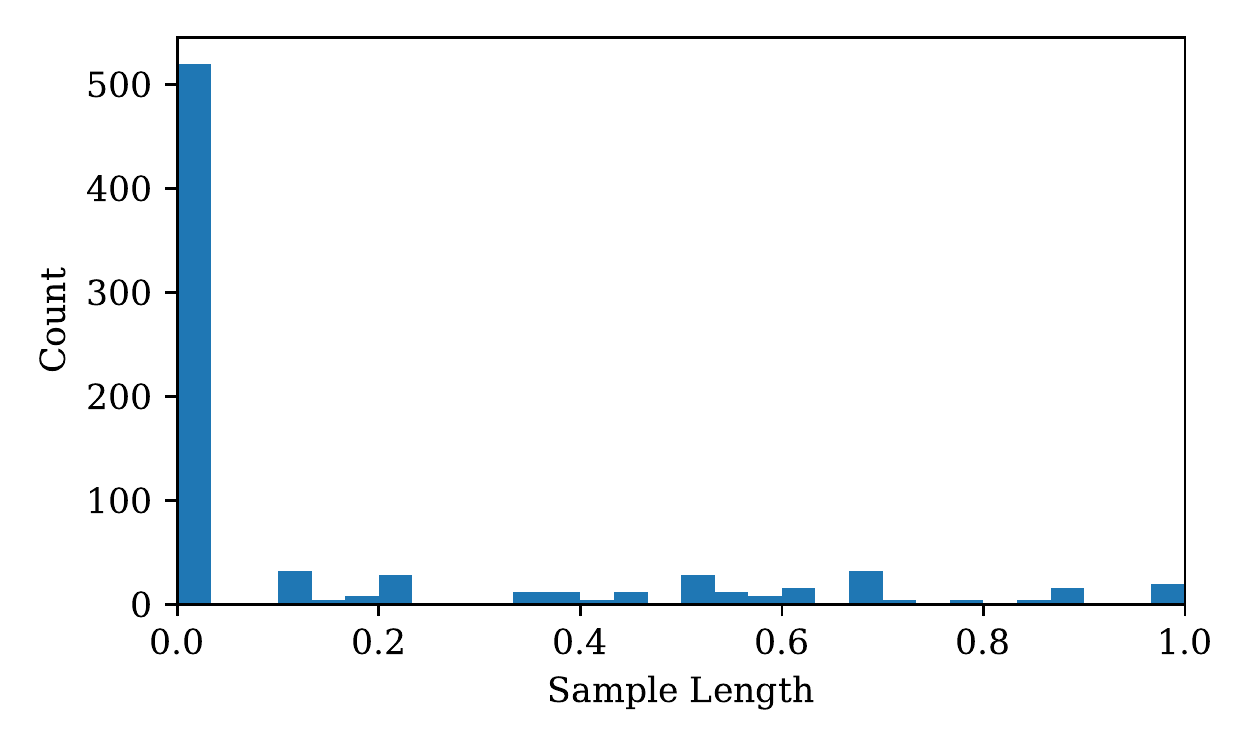}
    	\vspace{-0.4cm}
    	\subcaption{Mismatches.}
    	\label{image_ed_length2_1}
    \end{minipage}
    \hfill
	\begin{minipage}[b]{0.32\linewidth}
        \centering
    	\includegraphics[width=1.0\linewidth]{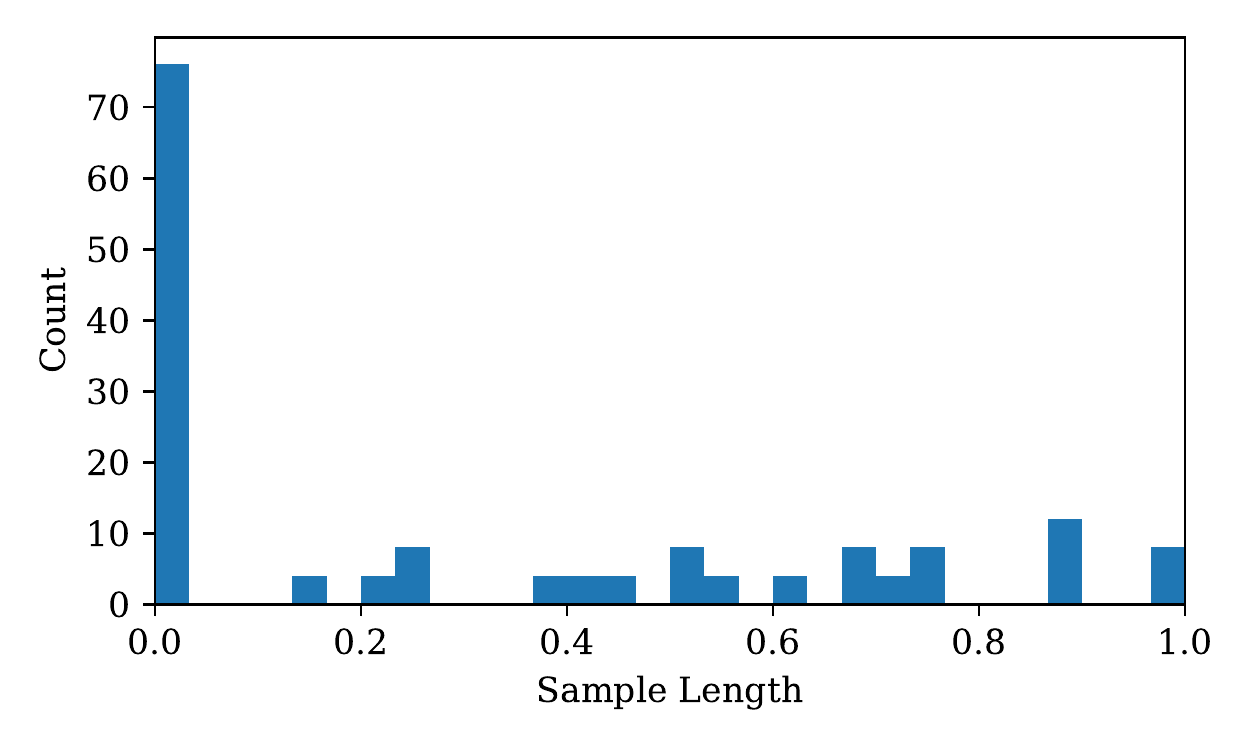}
    	\vspace{-0.4cm}
    	\subcaption{Insertions.}
    	\label{image_ed_length2_2}
    	\end{minipage}
    \hfill
	\begin{minipage}[b]{0.32\linewidth}
        \centering
    	\includegraphics[width=1.0\linewidth]{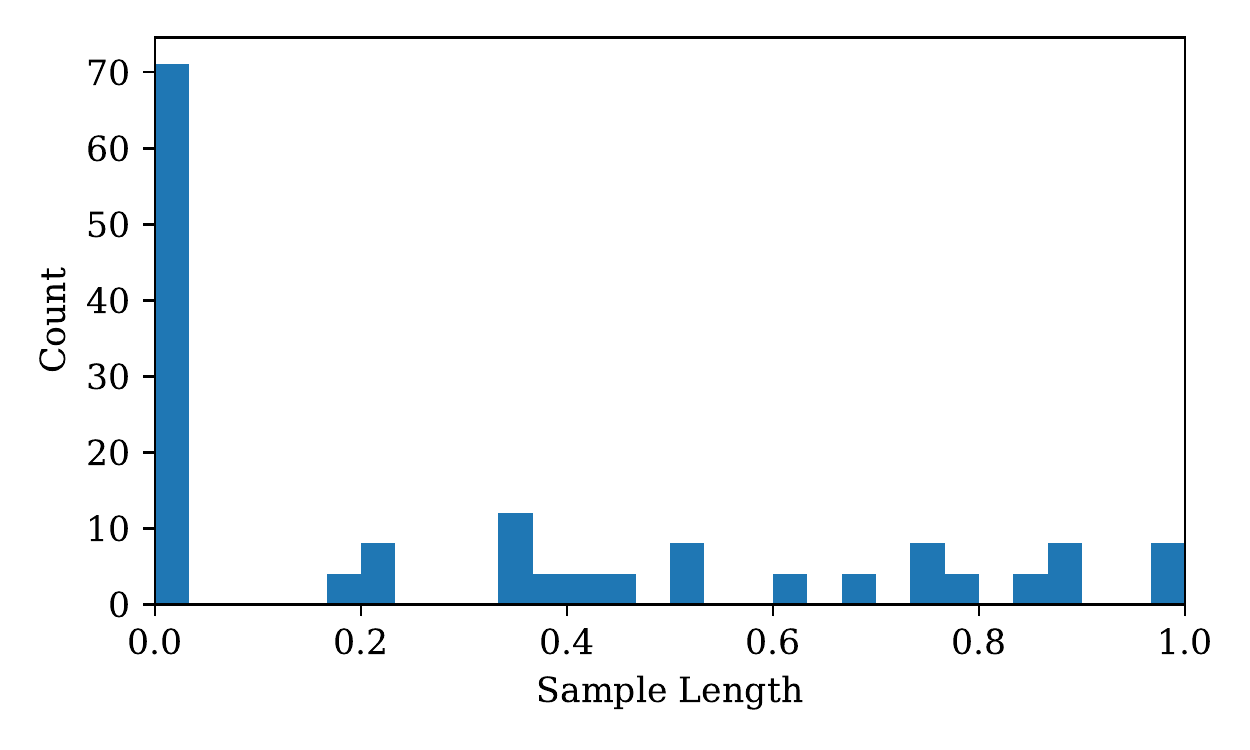}
    	\vspace{-0.4cm}
    	\subcaption{Deletions.}
    	\label{image_ed_length2_3}
    \end{minipage}
    \vspace{-0.1cm}
    \caption{Evaluation of the ED dependent on the normalized sample lengths for the OnHW-wordsTraj dataset.}
    \label{image_ed_length2}
\end{figure*}

\subsection{Left-handed Writers Datasets Evaluation}
\label{chap_eval_left}

For the left-handed writers datasets, we use the pre-trained weights from the right-handed datasets and train the CNN+BiLSTM architecture for 500 epochs. Table~\ref{table_results_left} summarizes all results for the sequence-based classification task (left) and the single character-based classification task (right). The motion dynamics of right- and left-handed writers is very different, especially with different rotations, and hence, also the sensor data are different. The models can still make use of the pre-trained weights and fine tuning leads to 1.24\% CER for the OnHW-equations-L dataset for the WD task, and 15.32\% CER for the OnHW-words500-L dataset, which is better than for the right-handed task. For the OnHW-wordsRandom-L dataset, the CER (5.40\%) increases, while the WER (32.73\%) decreases. Consistently, the results for the WI task decrease as the model overfits to specific writers due to the small amount of different left-handed writers in the training set. For single-based datasets, the fine tuning leads to a high WD classification accuracy of 92\% for the OnHW-symbols-L and split OnHW-equations-L datasets (compared to 96.2\% and 95.57\% for right-handed datasets, respectively), but decreases for WI tasks to 54\% and 51.5\% (compared to 79.51\% and 83.88\% for right-handed datasets, respectively). Due to the smaller size of the left-handed datasets, the models overfit to specific writers \cite{klass_lorenz}.

\begin{figure*}[t!]
	\centering
    \includegraphics[trim=0 0 0 10, clip, width=1.0\linewidth]{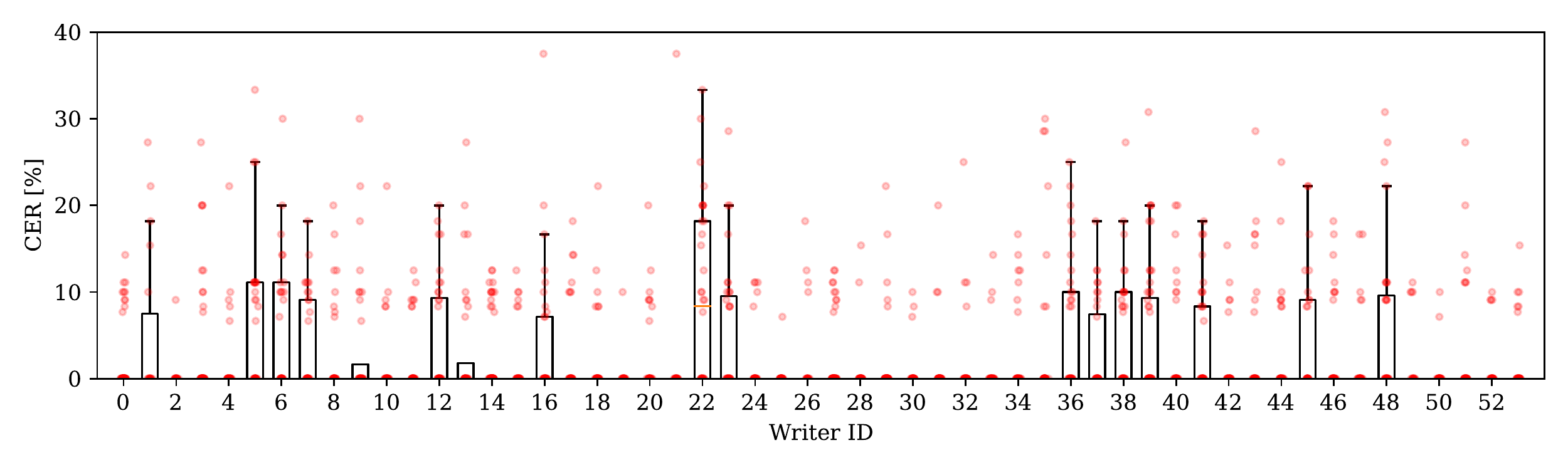}
    \vspace{-0.65cm}
    \caption{Writer-dependent CER (\%) for the OnHW-equations (WD) dataset.}
    \label{image_writer_dependency}
\end{figure*}

\subsection{Edit Distance and Writer Analysis}
\label{chap_eval_ed_writer}

\paragraph{Evaluation of Sample Length Dependent Edit Distance.} We show the sample length dependent counts of wrong predictions, i.e., mismatches, insertions and deletions, for the OnHW-equations (see Figure~\ref{image_ed_length1}) and OnHW-wordsTraj (see Figure~\ref{image_ed_length2}) datasets. For the OnHW-equations dataset, a high appearance of mismatches and insertions appear at the starting and end characters, while deletions emerge more even over the whole equations. The first character of words are significantly often mismatched or has to be inserted or deleted for the OnHW-wordsTraj dataset. This shows the unequal distribution of samples for the words datasets (see Figure~\ref{figure_statistics_dataset3}), while the equations dataset is very equally distributed (see Figure~\ref{figure_statistics_dataset2}).

\paragraph{Writer-Dependent Evaluation.} Figure~\ref{image_writer_dependency} shows the writer-dependent evaluation of the OnHW-equations dataset. The CER of many samples of several writers, e.g., ID 0, 2-4, 24-35, 42-44, and 49-53, is 0\%. The CER increases only for a small number of samples. The range of the CER increases for writer IDs 1, 5-7, 22, 23, and 36-39. Hence, the writing style and with that the sensor data is different and out-of-distribution in the dataset.
\section{Discussion \& Summary}
\label{chap_summary}

\subsection{Social Impact, Applications and Limitations}
\label{chap_summ_social_impact}

Handwriting is important in different fields, in particular \textit{graphomotoric}. The visual feedback provided by the pen, for instance, helps young students and children to learn a new language. Hence, research for HWR is very advanced. However, state-of-the-art methods to recognize handwriting (a) require to write on a special device, which might adversely affect the writing style, (b) require to take images of the handwritten text, or (c) are based on premature technical systems, i.e., the sensor pen is only a prototype \cite{deselaers}. The publicly available sensor pen developed by STABILO International GmbH has previously been used by \cite{koellner,ott} and allows an easier data collection than previous techniques. The research for collecting devices which do not influence the handwriting style is becoming increasingly important and with it also the social impact of resulting datasets. The aim of our dataset is to support the learning of students in schools or self-paced learning from home without additional effort \cite{alonso,wiley}. A well-known bottleneck for many machine learning algorithms is their requirement for large amounts of data samples without under-represented data patterns. For our HWR application, a large variety of different writing styles (cursive or printed characters, left- or right-handed, and beginner or advanced writers), pen rotations, and writing surfaces (especially different vibrations of the paper) are necessary. We provide an evaluation benchmark for right- and left-handed datasets. As motion dynamics between right- and left-handed writers are very different, extracting mutual information is a challenging task. The ratio between both groups approximately fits the real-world distribution, i.e., the under-representation of left-handed writers (10.6\%). Only adults without any selection participated at data recording as the handwriting style of students changes quickly with the age \cite{barrett}.

\subsection{Experimental Results}
\label{chap_summ_exp_results}

We performed several benchmarks and come to the following conclusions: (1) For the seq2seq classification task, we evaluated several methods based on CNNs in combination with RNNs on inertial-based datasets written on paper and on tablet, and evaluated state-of-the-art trajectory-based datasets. Depending on the dataset size, our CNN+BiLSTM model is on par with the InceptionTime+BiLSTM architecture. A search of architecture hyperparameters is important to achieve a generalized model for a real-world application. Our Transformer-based architecture could not outperform simpler convolutional models. (2) Sensor data augmentation leads to a better generalized training. (3) For the single classification task, our simple CNN+[LSTM, BiLSTM, TCN] can outperform state-of-the-art techniques. (4) Cross-entropy variants (i.e., label smoothing) improves results that are dependent on the dataset (i.e., label noise and class balance). (5) Writer-independent classification of (under-represented) left-handed writers is very challenging that is interesting for future research.

\subsection{Collection Consent and Personal Information}
\label{chap_summ_personal_information}

While recording the datasets, we collected the consent of all participants. We only collected the raw data from the sensor-enhanced pen, and for statistics the age and gender of the participant and their handedness. The handedness is necessary because the pen is differently rotated between left- and right-handed writers. The recording localization was Germany. An ID is assigned to every participant such that the dataset is fully pseudonymized. The ID is necessary for the WD and WI evaluation.

\subsection{Conclusion and Future Research}
\label{chap_summ_conclusion}

We proposed several equations and words OnHWR datasets for a seq2seq classification task, as well as one symbol dataset for the single character classification task based on a novel sensor-enhanced pen. By utilizing (Bi)LSTM and TCN models combined with CNNs and different Transformer models, we proposed a broad evaluation benchmark for lexicon-free classification. Various augmentation techniques showed notable improvement in classification accuracy. Our detailed evaluation of the WD and WI tasks sets important challenges for future research and provides a benchmark foundation for novel methodological advancements. For example, semi-supervised learning and few-shot learning such as prototypical networks could improve the classification accuracy of under-represented writers. Exploiting offline datasets for pre-training, or the use of lexicon and language models might further allow the model to better learn the task.

\section*{Acknowledgments}
We sincerely thank all participants taking part in the data recordings, and acknowledge the work of various researchers from the STABILO International GmbH, Kinemic GmbH, Fachdidaktik Deutsch Primarstufe (DID) of the Saarland University, Machine Learning and Data Analytics Lab of the Friedrich-Alexander University (FAU) and Fraunhofer Institute for Integrated Circuits (IIS) for their help with the data collection.

\section*{Funding}
This work was supported by the Federal Ministry of Education and Research (BMBF) of Germany by Grant No. 01IS18036A (David R\"ugamer) and by the research program Human-Computer-Interaction through the project "Schreibtrainer", Grant No. 16SV8228, as well as by the Bavarian Ministry for Economic Affairs, Infrastructure, Transport and Technology through the Center for Analytics-Data-Applications (ADA-Center) within the framework of "BAYERN DIGITAL II".

\section*{Declarations}

\begin{itemize}
\item \textbf{Ethics approval:} see Section~\ref{chap_summ_social_impact}
\item \textbf{Consent to participate:} see Section~\ref{chap_summ_personal_information}
\item \textbf{Availability of data and materials} Data and materials will be publicly available upon publication that includes the OnHW-chars, OnHW-symbols, split OnHW-equations, OnHW-equations, OnHW-wordsTraj and OnHW-words500 dataset. We include left-handed and right-handed as well as writer-dependent (WD) and writer-independent (WI) splits. We will publish the used WD and WI splits of the IAM-OnDB and VNOnDB datasets for reconstruction of results.\\
\href{https://www.iis.fraunhofer.de/de/ff/lv/dataanalytics/anwproj/schreibtrainer/onhw-dataset.html}{www.iis.fraunhofer.de/de/ff/lv/dataanalytics/anwproj/ schreibtrainer/onhw-dataset.html}
\item \textbf{Code availability:} Will be publicly available upon publication.
\end{itemize}

\bibliographystyle{spmpsci.bst}
\bibliography{IJDAR2022}

\section*{Biography}

\setlength{\intextsep}{6pt}
\setlength{\columnsep}{14pt}
\begin{wrapfigure}{l}{23mm} 
    \includegraphics[width=1in,height=1.25in,clip,keepaspectratio]{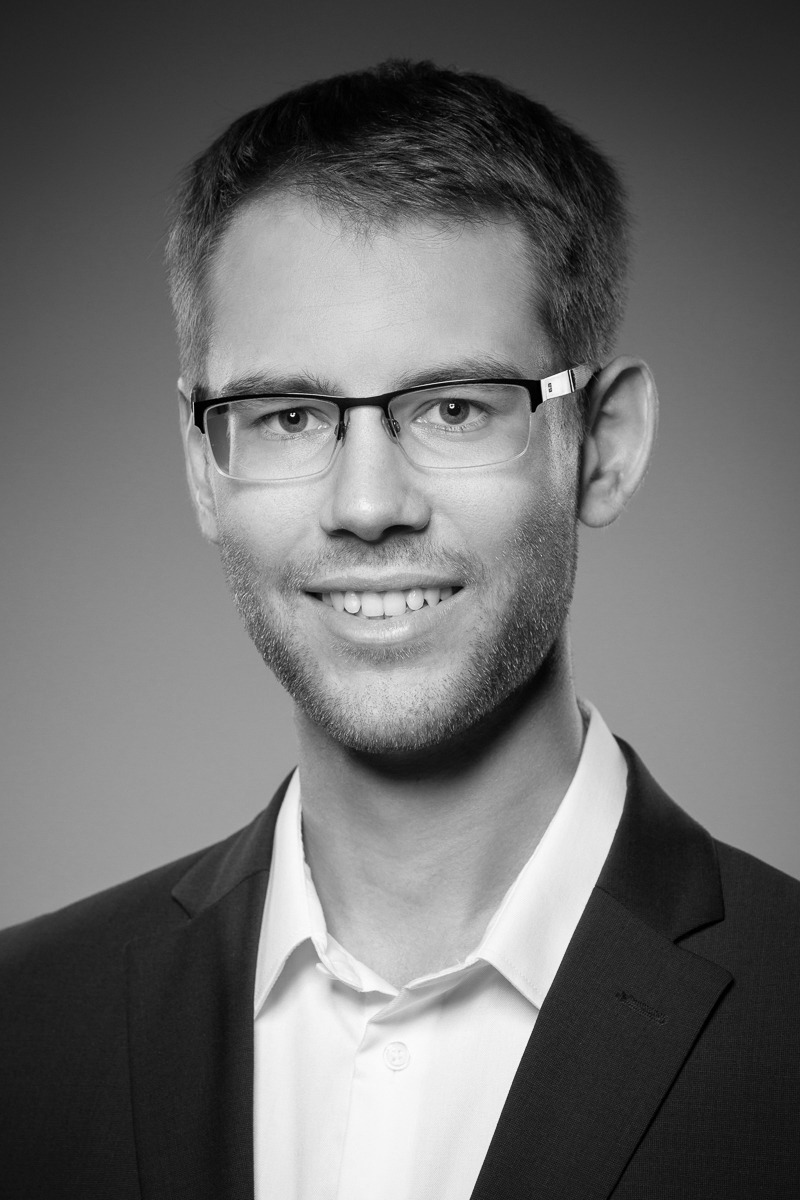}
\end{wrapfigure}\par
\,\,\,\,\,\,\, \textbf{Felix Ott} received his M.Sc. degree in Computational Engineering at the FAU Erlangen-N{\"u}rnberg in 2019. He joined the Hybrid Positioning \& Information Fusion group at Fraunhofer IIS. In 2020, he started his Ph.D. at the LMU Munich in the Probabilistic Machine and Deep Learning group. His research covers online handwriting recognition and multimodal learning.

\setlength{\intextsep}{6pt}
\setlength{\columnsep}{14pt}
\begin{wrapfigure}{l}{23mm} 
    \includegraphics[width=1in,height=1.25in,clip,keepaspectratio]{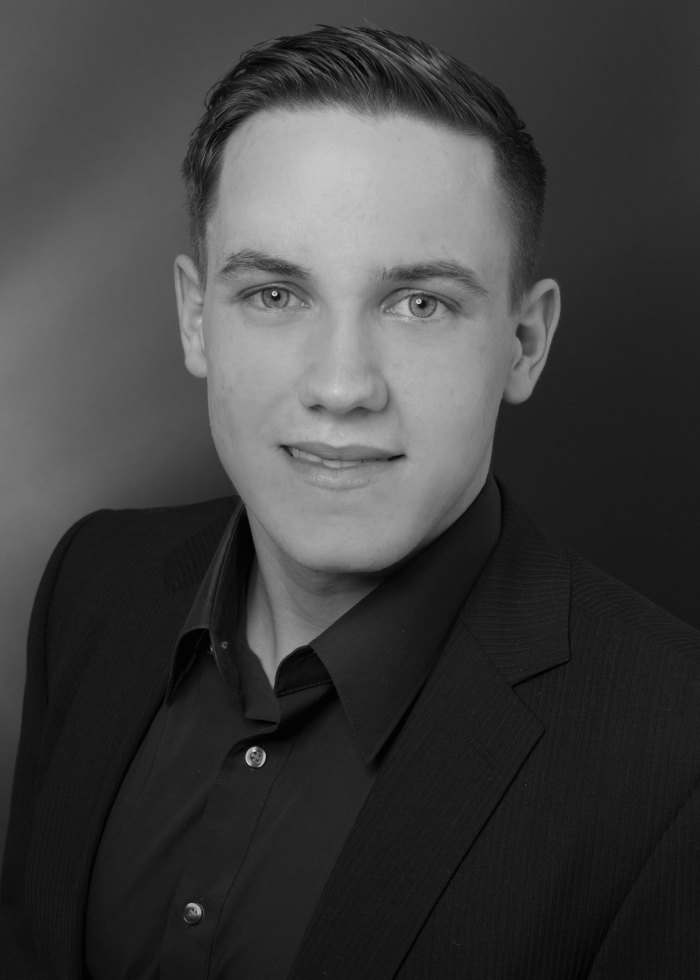}
\end{wrapfigure}\par
\textbf{David Rügamer} is an interim professor for Computational Statistics at the RWTH Aachen. Before he was research associate, lecturer and interim professor for Data Science at the LMU Munich, where he also received his Ph.D. in 2018. His research is concerned with scalability of statistical modeling as well as machine learning for functional and multimodal data.

\setlength{\intextsep}{6pt}
\setlength{\columnsep}{14pt}
\begin{wrapfigure}{l}{23mm} 
    \includegraphics[width=1in,height=1.25in,clip,keepaspectratio]{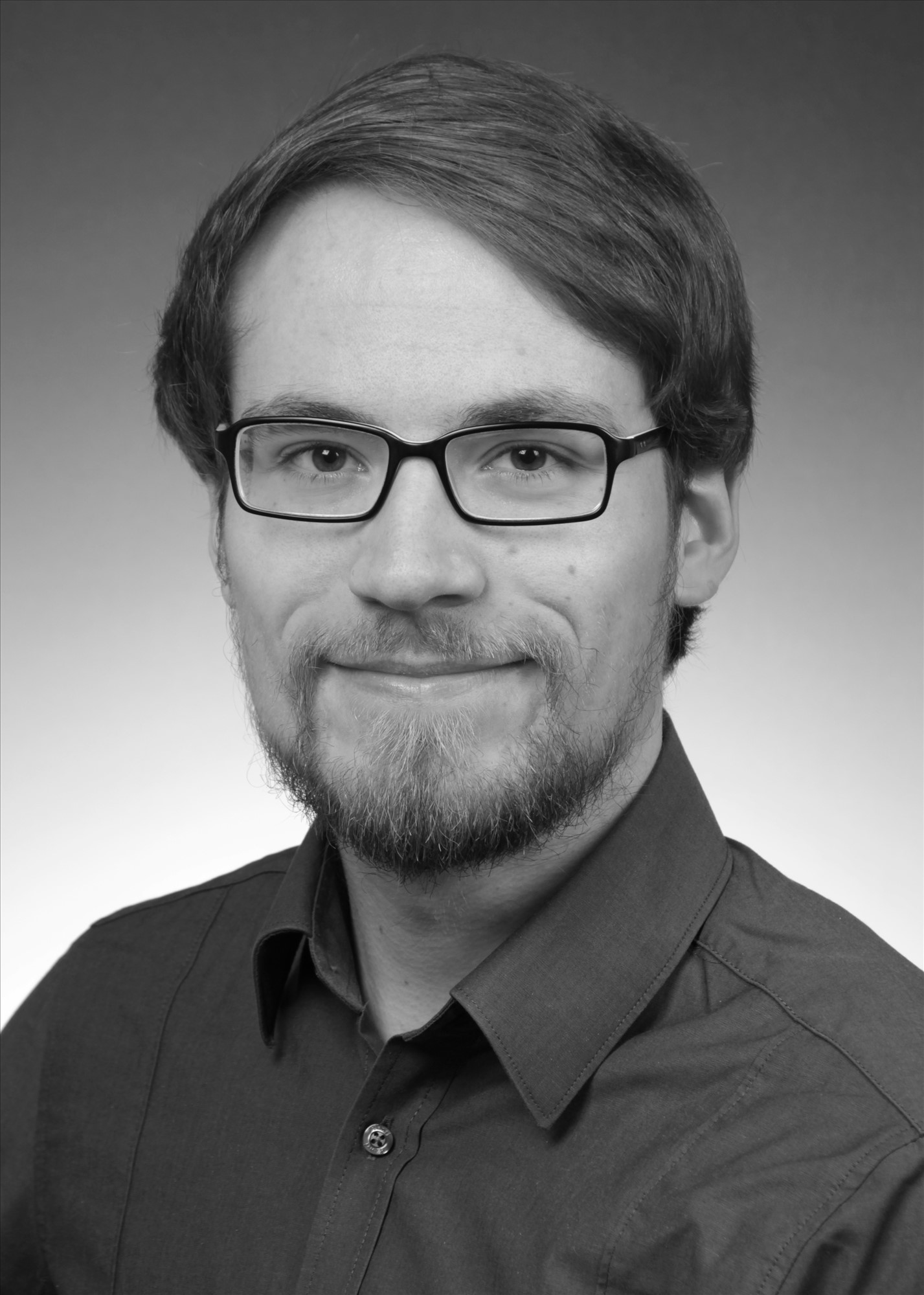}
\end{wrapfigure}\par
\textbf{Lucas Heublein} received his M.Sc. degree in Integrated Life Science at the FAU Erlangen-Nürnberg. In 2020, he started his Computer Science degree at the FAU. He joined the Hybrid Positioning \& Information Fusion group at the Fraunhofer IIS in 2020 as a student assistant. \\

\setlength{\intextsep}{6pt}
\setlength{\columnsep}{14pt}
\begin{wrapfigure}{l}{23mm} 
    \includegraphics[width=1in,height=1.25in,clip,keepaspectratio]{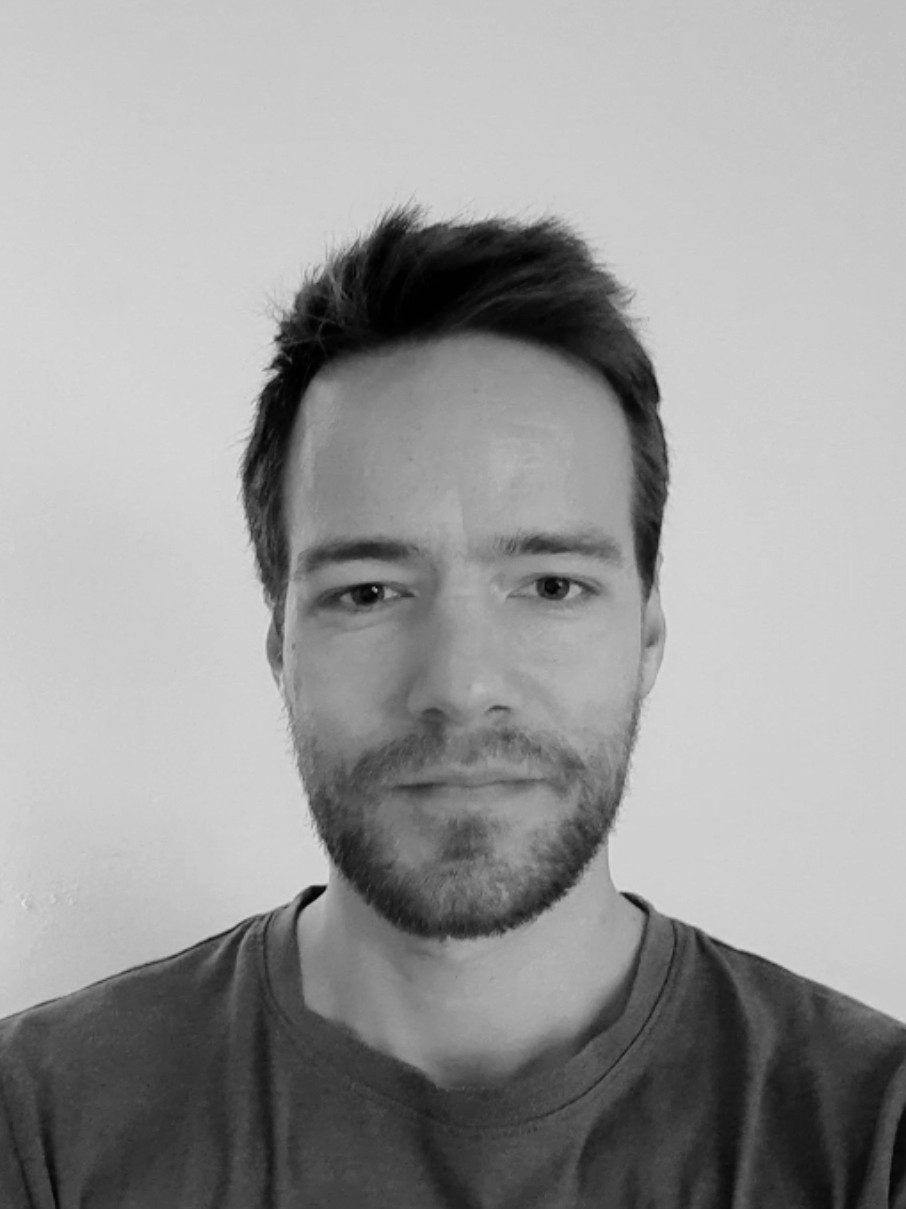}
\end{wrapfigure}\par
\textbf{Tim Hamann} is an AI software developer at STABILO International GmbH. In order to make AI-based applications available to the users, his focus is on model deployment, data management, and research projects. His work revolves around a ballpoint pen equipped with inertial sensors. He has received his M.Sc. in Computer Science at the FAU in 2018.

\setlength{\intextsep}{6pt}
\setlength{\columnsep}{14pt}
\begin{wrapfigure}{l}{23mm} 
    \includegraphics[width=1in,height=1.25in,clip,keepaspectratio]{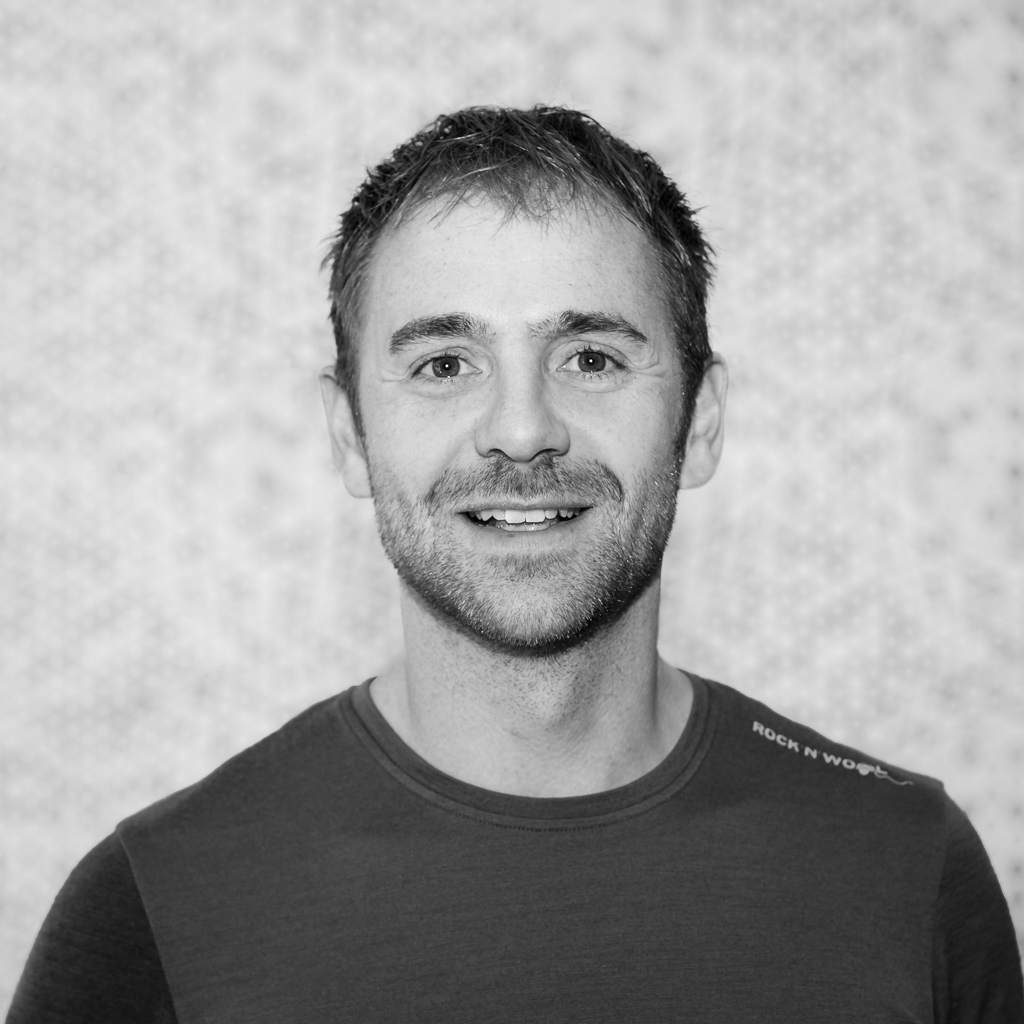}
\end{wrapfigure}\par
\textbf{Jens Barth} is a project manager at STABILO International GmbH and is responsible for digital products in the field of sensor-based writing analysis with a digitized pen. He coordinates ML projects on the topic of handwriting recognition and objective analysis of writing motor skills using inertial sensors in a pen. Jens Barth completed his Ph.D. on sensor-based gait analysis in Parkinson's patients at the Machine Learning and Data Analysis lab at the FAU.

\setlength{\intextsep}{6pt}
\setlength{\columnsep}{14pt}
\begin{wrapfigure}{l}{23mm} 
    \includegraphics[width=1in,height=1.25in,clip,keepaspectratio]{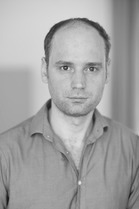}
\end{wrapfigure}\par
\textbf{Bernd Bischl} is a full professor for statistical learning and data science at the LMU Munich and a director of the Munich Center of Machine Learning. His research focuses amongst other things on AutoML, interpretable machine learning and ML benchmarking. \\

\setlength{\intextsep}{6pt}
\setlength{\columnsep}{14pt}
\begin{wrapfigure}{l}{23mm} 
    \includegraphics[width=1in,height=1.25in,clip,keepaspectratio]{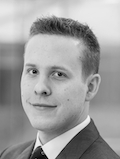}
\end{wrapfigure}\par
\vspace{-0.4cm}
\textbf{Christopher Mutschler} leads the precise positioning and analytics department at Fraunhofer IIS. Prior to that, Christopher headed the Machine Learning \& Information Fusion group. He gives lectures on machine learning at the FAU Erlangen-N{\"u}rnberg, from which he also received both his Diploma and Ph.D. in 2010 and 2014 respectively. Christopher’s research combines machine learning with radio-based localization.

\clearpage
\appendix
\section{Appendices}
\label{sec_appendix}

\normalsize In this appendix, we will give a general overview of related work in Section~\ref{sec_appendix_related_work}. We propose more details about the sensor pen in Section~\ref{sec_appendix_pen}, and present the data acquisition and format in Section~\ref{sec_appendix_data_format}. While Section~\ref{sec_appendix_sensor} shows additional samples, Section~\ref{sec_appendix_statistics} proposes more detailed statistics of the datasets. We state the chosen Transformer parameters in Section~\ref{sec_appendix_trans_parameters}. Section~\ref{sec_appendix_detailed_evaluation} concludes with more evaluation details.

\subsection{General Overview of Related Work}
\label{sec_appendix_related_work}

\paragraph{Temporal Convolutional Networks (TCNs).} TCNs consist of CNNs as encoders to extract spatio-temporal information for low-level feature computation, and a classifier that captures high-level temporal information using a recurrent network. TCNs can take a series of any length and output it with the same length. They perform well in prediction tasks with time-series data. \cite{yan_mu} TCNs have been used for the HWR task in \cite{sharma2,sharma}.

\paragraph{RNNs.} \cite{wigington} proposed a CNN-LSTM model for text detection, segmentation and recognition. The performance of RNNs can be improved using dropout \cite{pham}. \cite{carbune} highly improve classification accuracies by a stack of \textbf{bidirectional LSTMs} \cite{graves}. \cite{tian} combined BiLSTMs in the word encoder with word inter-attention for a multi-task document classification approach. \textbf{Multi-dimensional RNNs} as the MDLSTM-RNNs~\cite{graves3} scan the input in the four possible directions, where LSTM cell inner states and output are computed from previous positions in the vertical and horizontal directions. \cite{voigtlaender} processed the input in a diagonal-wise fashion to enable GPU-based training and explored deeper and wider MDLSTMs architectures for HWR. \cite{bluche} transformed the 2D representation into a sequence of predictions to enable end-to-end processing of paragraphs. However, these architectures are computationally expensive and extract features visually similar to CNNs, hence, 2D long-term dependencies may not be essential \cite{puigcerver}. \cite{dutta} integrated a spatial Transformer network into their RCNN method.

\paragraph{Transformers.} They aim for handling long-range dependencies with ease relying entirely on self-attention to compute representations of its input and output without using sequence-aligned RNNs or convolution. Vaswani et al.~\cite{vaswani} showed their Transformer architecture consisting of a decoder, encoder and multi-head attention to be superior in quality while being more parallelizable and requiring significantly less time to train. \cite{kang} introduced a novel method for offline HWR that  bypasses any recurrence and uses multi-head self-attention layers at visual and textual stages. As Transformer-based models scale quadratically with the sequence length due to their self-attention, the Longformer introduced an attention mechanism that scales linearly, and was applied to process documents of thousands of tokens. The Performer~\cite{choromanski} that estimates (softmax) full-rank attention Transformers also use only linear complexity. The Perceiver~\cite{jaegle} scales to high-dimensional inputs such as audio, videos, images and point-clouds by using cross-attentional principles before using a stack of Transformers in the latent space.

\subsection{Additional Information of the Sensor Pen}
\label{sec_appendix_pen}

The DigiPen by STABILO International GmbH is a sensor-enhanced ballpoint pen with internal data processing capabilities. A Bluetooth module enables live streaming of the integrated sensor at 100\textit{Hz} to a connected device. The DigiPen development kit is also publicly available.\footnote{DigiPen Development Kit: \url{https://stabilodigital.com/devkit-demoapp-introduction/}} The pen has an ergonomic soft-touch grip zone, such that the writing feels comfortable and is as normal writing on paper. The pen's overall length is 167\textit{mm}, its diameter is 15\textit{mm}, and its weighs 25\textit{g}. The pen is equipped with a front accelerometer (STM LSM6DSL), a rear accelerometer (Freescale MMA8451Q), a gyroscope (STM LSM6DSL), a magnetometer (ALPS HSCDTD008A), and a force sensor (ALPS HSFPAR003A). The front and rear accelerometers are differently oriented. The accelerometers were adjusted to a range of $\pm 2g$ with a resolution of 16\,\textit{bit} of the front and 14\,\textit{bit} for the rear accelerometer. The gyroscope has a range of $\pm 1,000 ^\circ/s$ (16\,\textit{bit}), and the magnetometer has a range of $2.4mT$ (14\,\textit{bit}). The measurement range of the force sensor is between 0 and $5.32N$ (12\,\textit{bit}).

\subsection{Data Acquisition and Format}
\label{sec_appendix_data_format}

\begin{figure*}[t!]
	\centering
	\begin{minipage}[b]{0.49\linewidth}
        \centering
    	\includegraphics[width=1.0\linewidth]{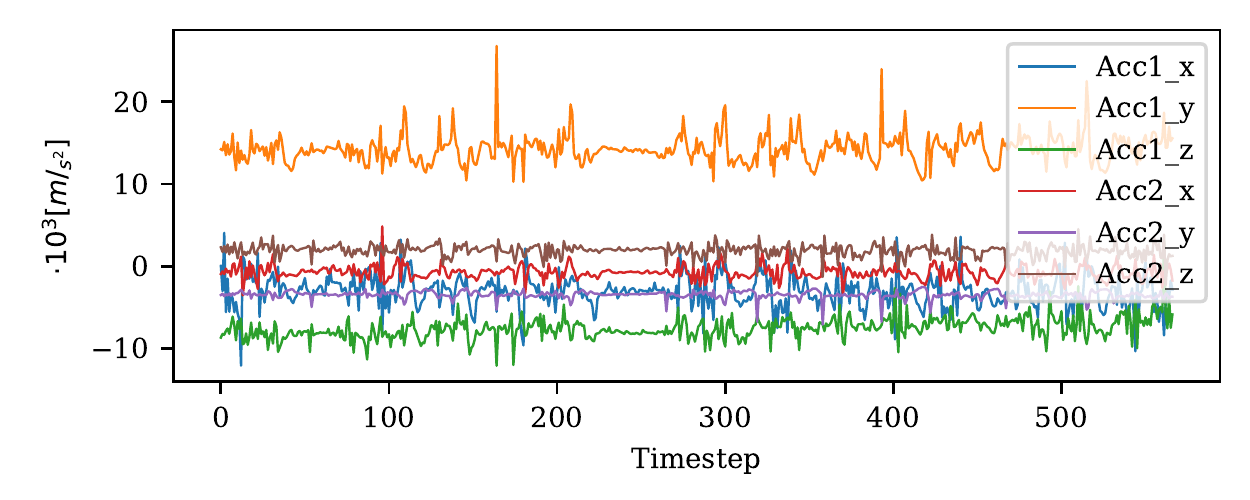}
    	\vspace{-0.6cm}
    	\subcaption{Front and rear accelerometer.}
    	\label{image_app_exe_equ_1}
    \end{minipage}
    \hfill
	\begin{minipage}[b]{0.49\linewidth}
        \centering
    	\includegraphics[width=1.0\linewidth]{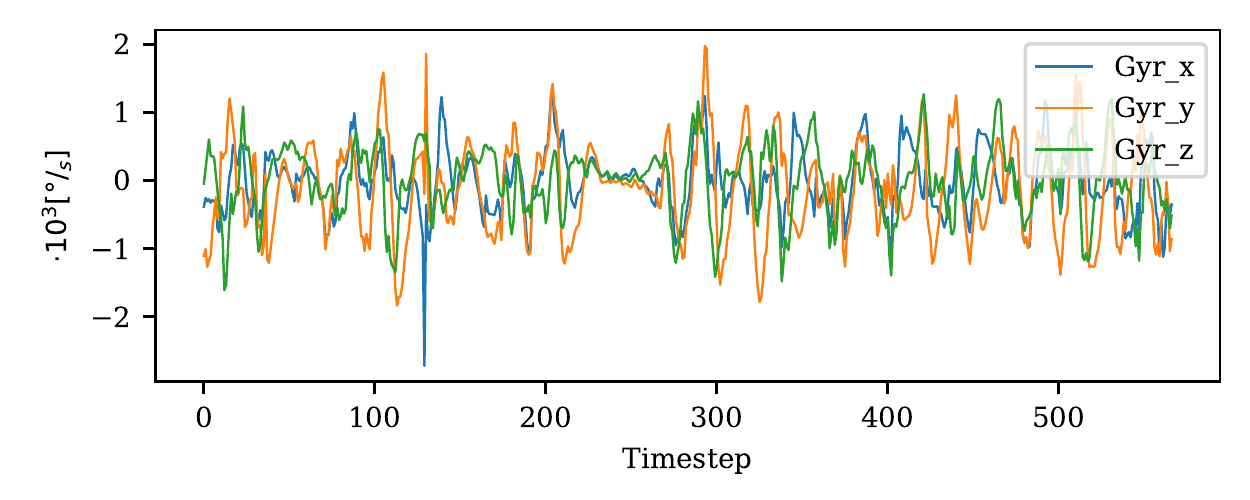}
    	\vspace{-0.6cm}
    	\subcaption{Gyroscope.}
    	\label{image_app_exe_equ_2}
    \end{minipage}
    \hfill
	\begin{minipage}[b]{0.49\linewidth}
        \centering
    	\includegraphics[width=1.0\linewidth]{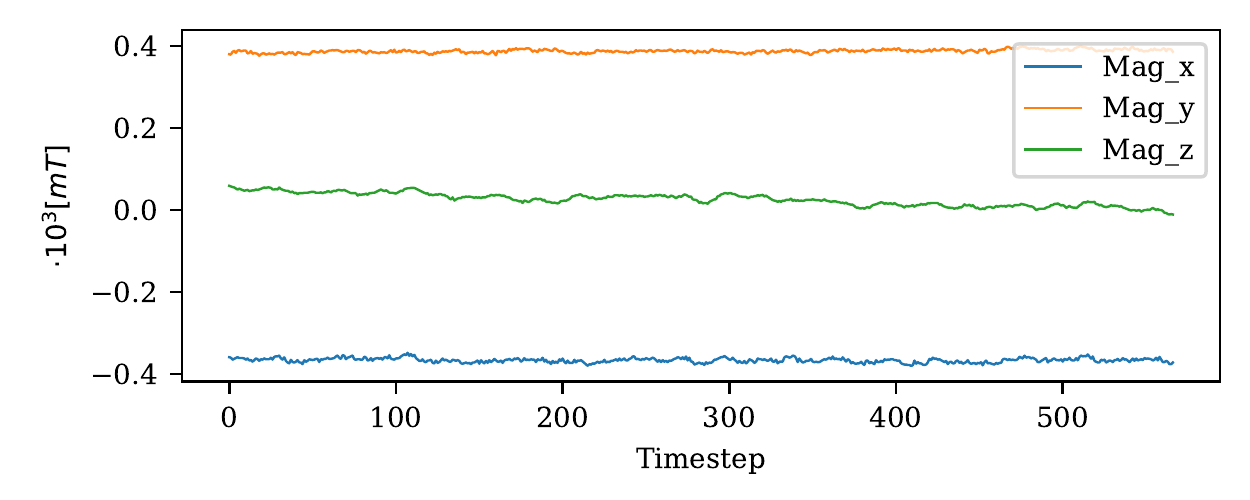}
    	\vspace{-0.6cm}
    	\subcaption{Magnetometer.}
    	\label{image_app_exe_equ_3}
    \end{minipage}
    \hfill
	\begin{minipage}[b]{0.49\linewidth}
        \centering
    	\includegraphics[width=1.0\linewidth]{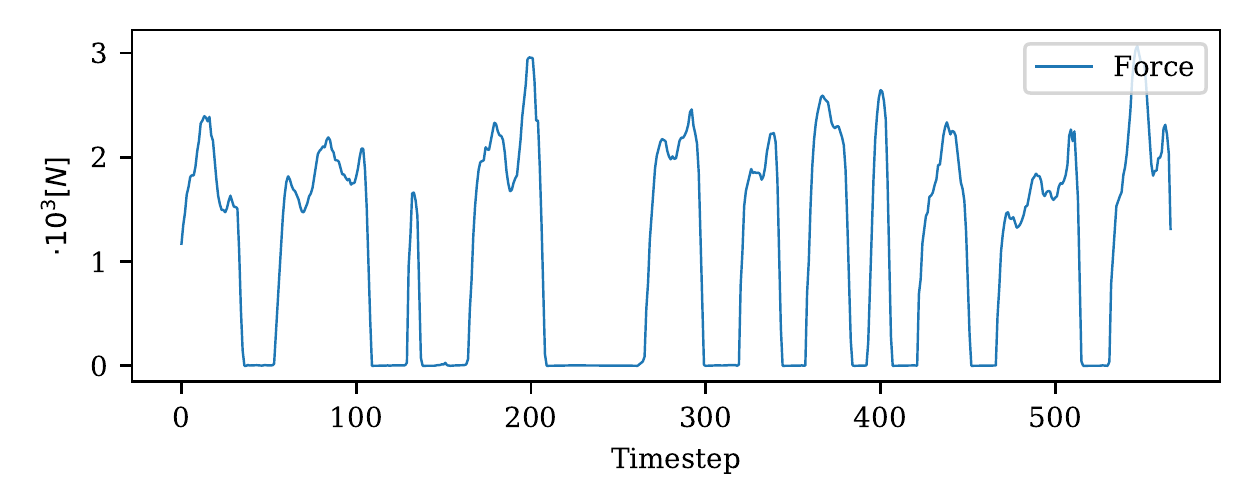}
    	\vspace{-0.6cm}
    	\subcaption{Force sensor.}
    	\label{image_app_exe_equ_4}
    \end{minipage}
    \vspace{-0.1cm}
    \caption{Exemplary sensor data of one sample of the OnHW-equations dataset.}
    \label{image_app_exe_equ}
\end{figure*}

\begin{figure*}[t!]
	\centering
	\begin{minipage}[b]{0.49\linewidth}
        \centering
    	\includegraphics[width=1.0\linewidth]{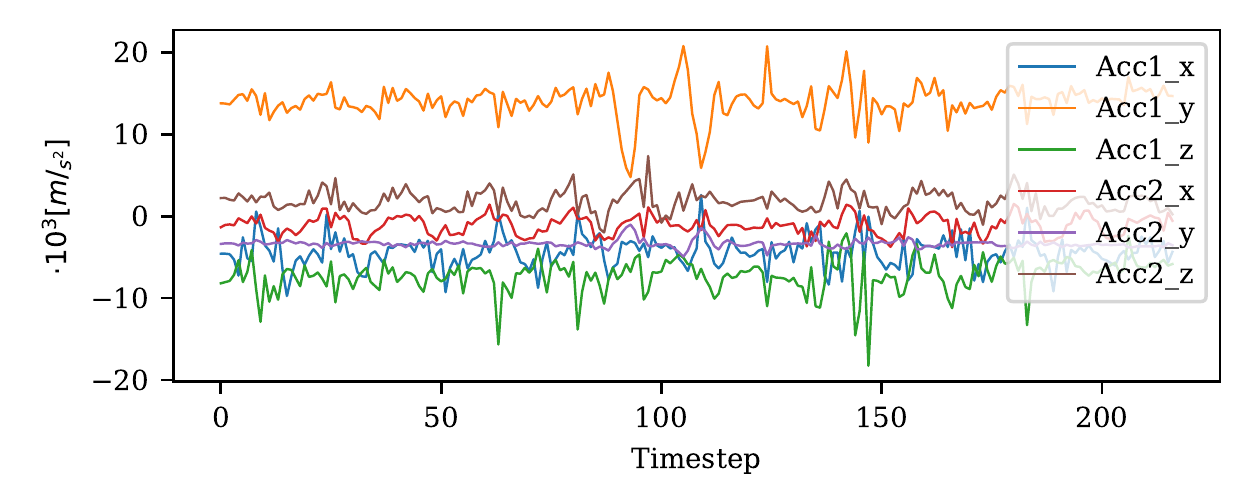}
    	\vspace{-0.6cm}
    	\subcaption{Front and rear accelerometer.}
    	\label{image_app_exe_word_1}
    \end{minipage}
    \hfill
	\begin{minipage}[b]{0.49\linewidth}
        \centering
    	\includegraphics[width=1.0\linewidth]{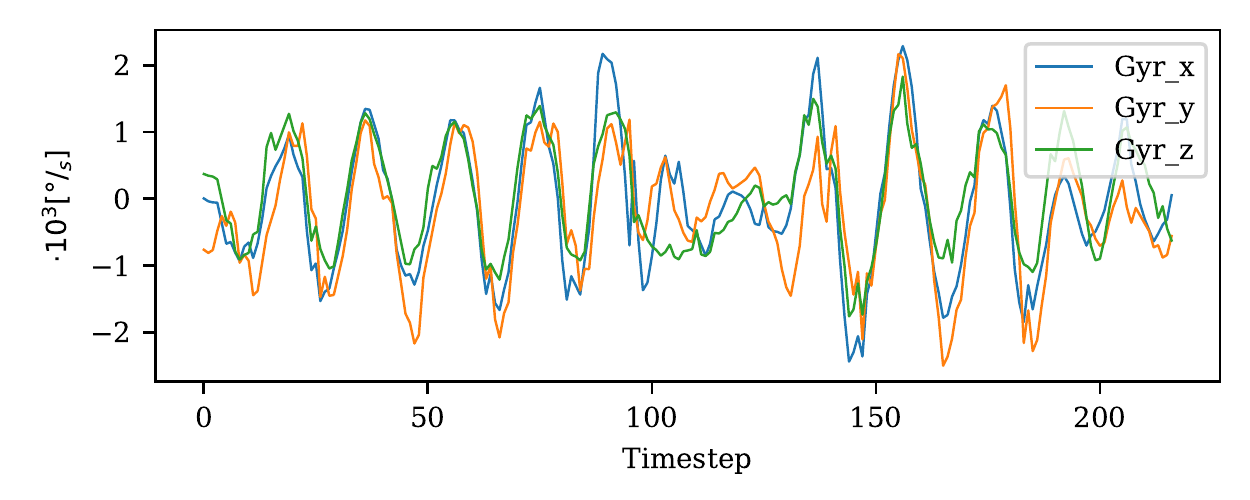}
    	\vspace{-0.6cm}
    	\subcaption{Gyroscope.}
    	\label{image_app_exe_word_2}
    \end{minipage}
    \hfill
	\begin{minipage}[b]{0.49\linewidth}
        \centering
    	\includegraphics[width=1.0\linewidth]{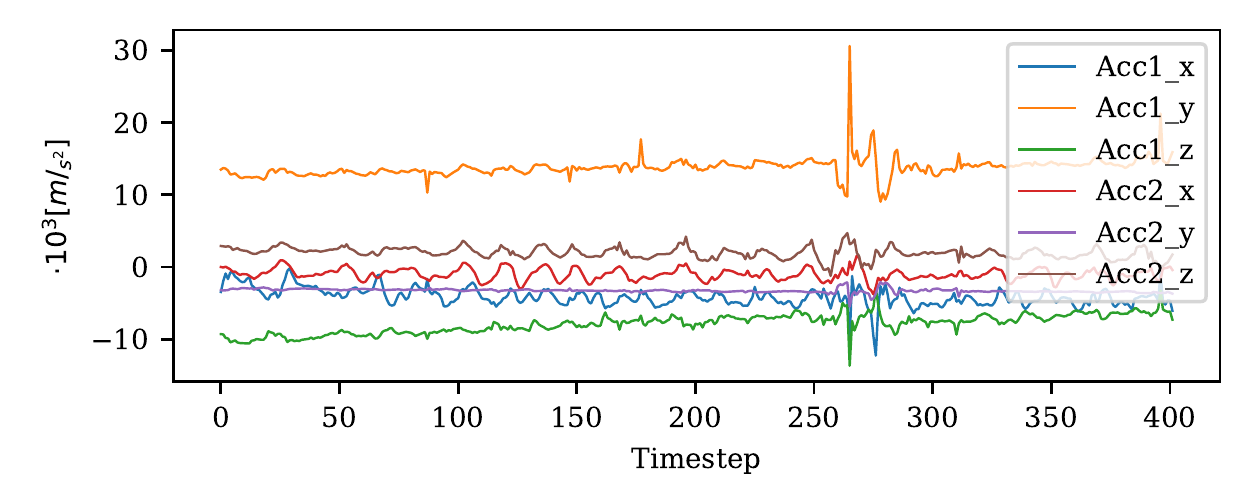}
    	\vspace{-0.6cm}
    	\subcaption{Front and rear accelerometer.}
    	\label{image_app_exe_word_tablet_1}
    \end{minipage}
    \hfill
	\begin{minipage}[b]{0.49\linewidth}
        \centering
    	\includegraphics[width=1.0\linewidth]{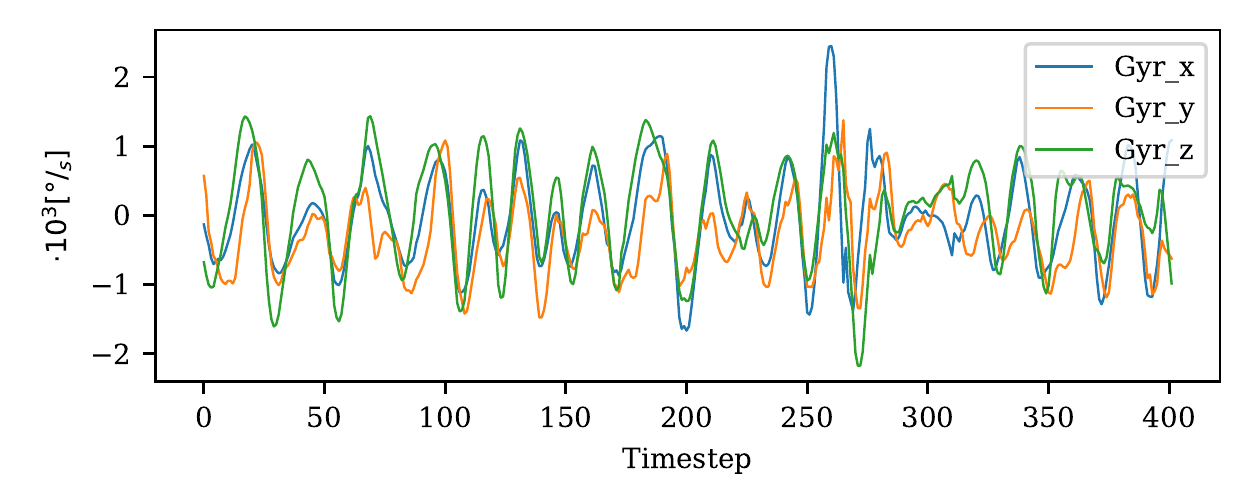}
    	\vspace{-0.6cm}
    	\subcaption{Gyroscope.}
    	\label{image_app_exe_word_tablet_2}
    \end{minipage}
	\vspace{-0.05cm}
    \caption{Exemplary accelerometer and gyroscope data of one sample of the OnHW-words500 (top) and OnHW-wordsTraj (bottom) datasets.}
    \label{image_app_exe_word}
\end{figure*}

STABILO International GmbH provides a recording app to obtain the sensor data that is publicly available. Through this setup we also recorded the ground truth labels. The data was recorded over a period of 1.5 years. To achieve equally distributed datasets, we apply the following constraints. The writer has to write on a normal, white paper padded by five additional sheets, and has to sit on a chair in front of a table. The logo of the pen needs to face upwards. Users are allowed to write in a cursive or printed style. The way of holding the pen and the size of handwriting was not constrained. Prior to recording the gyroscope and magnetometer biases and the magnetometer scaling has to be determined by calibrating the pen. We do not use the calibration data, but publish the calibration files along the datasets for possible future research. For more information, see \cite{ott}.

The data format is given as following. For each dataset we will publish the raw data that consists of the calibration file, a labels file with start and end timestep, and a data file with the corresponding 13 channels for each timestep. Additionally, we already preprocess the data and upload pickle (\textit{.pkl}) files. For each dataset and each of the five cross validation splits, we generated a train and validation file with the sensor data, the corresponding label, and the writer IDs. 

\begin{table*}
\begin{center}
\setlength{\tabcolsep}{3.0pt}
    \caption{Overview of sample data length average $L_A$ and standard deviation $L_D$ and number of strokes average $S_A$ and standard deviation $S_D$ per number of labels.}
    \label{table_length_datasets_overview}
    \small \begin{tabular}{ p{1.1cm} | p{0.5cm} | p{0.5cm} | p{0.5cm} | p{0.5cm} | p{0.5cm} | p{0.5cm} | p{0.5cm} | p{0.5cm} | p{0.5cm} | p{0.5cm} | p{0.5cm} | p{0.5cm} | p{0.5cm} | p{0.5cm} | p{0.5cm} }
    \multicolumn{1}{c|}{\textbf{Dataset}} & \multicolumn{1}{c|}{} & \multicolumn{1}{c|}{\textbf{2}} & \multicolumn{1}{c|}{\textbf{3}} & \multicolumn{1}{c|}{\textbf{4}} & \multicolumn{1}{c|}{\textbf{5}} & \multicolumn{1}{c|}{\textbf{6}} & \multicolumn{1}{c|}{\textbf{7}} & \multicolumn{1}{c|}{\textbf{8}} & \multicolumn{1}{c|}{\textbf{9}} & \multicolumn{1}{c|}{\textbf{10}} & \multicolumn{1}{c|}{\textbf{11}} & \multicolumn{1}{c|}{\textbf{12}} & \multicolumn{1}{c|}{\textbf{13}} & \multicolumn{1}{c}{\textbf{14}} \\ \hline
    \multicolumn{1}{l|}{OnHW-equations} & \multicolumn{1}{l|}{$L_A$} & \multicolumn{1}{c|}{-} & \multicolumn{1}{c|}{-} & \multicolumn{1}{c|}{-} & \multicolumn{1}{r|}{228} & \multicolumn{1}{r|}{332} & \multicolumn{1}{r|}{420} & \multicolumn{1}{r|}{482} & \multicolumn{1}{r|}{559} & \multicolumn{1}{r|}{642} & \multicolumn{1}{r|}{703} & \multicolumn{1}{r|}{777} & \multicolumn{1}{r|}{855} & \multicolumn{1}{r}{906} \\
    \multicolumn{1}{l|}{} & \multicolumn{1}{l|}{$L_D$} & \multicolumn{1}{c|}{-} & \multicolumn{1}{c|}{-} & \multicolumn{1}{c|}{-} & \multicolumn{1}{r|}{37} & \multicolumn{1}{r|}{105} & \multicolumn{1}{r|}{158} & \multicolumn{1}{r|}{167} & \multicolumn{1}{r|}{192} & \multicolumn{1}{r|}{207} & \multicolumn{1}{r|}{211} & \multicolumn{1}{r|}{230} & \multicolumn{1}{r|}{235} & \multicolumn{1}{r}{260} \\
    \multicolumn{1}{l|}{} & \multicolumn{1}{l|}{$S_A$} & \multicolumn{1}{c|}{-} & \multicolumn{1}{c|}{-} & \multicolumn{1}{c|}{-} & \multicolumn{1}{r|}{6.21} & \multicolumn{1}{r|}{7.90} & \multicolumn{1}{r|}{9.28} & \multicolumn{1}{r|}{10.37} & \multicolumn{1}{r|}{11.76} & \multicolumn{1}{r|}{13.05} & \multicolumn{1}{r|}{14.21} & \multicolumn{1}{r|}{15.57} & \multicolumn{1}{r|}{17.03} & \multicolumn{1}{r}{18.28} \\
    \multicolumn{1}{l|}{} & \multicolumn{1}{l|}{$S_D$} & \multicolumn{1}{c|}{-} & \multicolumn{1}{c|}{-} & \multicolumn{1}{c|}{-} & \multicolumn{1}{r|}{2.21} & \multicolumn{1}{r|}{1.67} & \multicolumn{1}{r|}{1.88} & \multicolumn{1}{r|}{2.32} & \multicolumn{1}{r|}{2.61} & \multicolumn{1}{r|}{2.69} & \multicolumn{1}{r|}{2.86} & \multicolumn{1}{r|}{2.93} & \multicolumn{1}{r|}{3.76} & \multicolumn{1}{r}{3.00} \\ \hline
    \multicolumn{1}{l|}{OnHW-words500} & \multicolumn{1}{l|}{$L_A$} & \multicolumn{1}{r|}{87} & \multicolumn{1}{r|}{120} & \multicolumn{1}{r|}{150} & \multicolumn{1}{r|}{187} & \multicolumn{1}{r|}{223} & \multicolumn{1}{r|}{272} & \multicolumn{1}{r|}{320} & \multicolumn{1}{r|}{349} & \multicolumn{1}{r|}{393} & \multicolumn{1}{r|}{425} & \multicolumn{1}{r|}{486} & \multicolumn{1}{r|}{502} & \multicolumn{1}{r}{603} \\
    \multicolumn{1}{l|}{} & \multicolumn{1}{l|}{$L_D$} & \multicolumn{1}{r|}{258} & \multicolumn{1}{r|}{36} & \multicolumn{1}{r|}{35} & \multicolumn{1}{r|}{51} & \multicolumn{1}{r|}{51} & \multicolumn{1}{r|}{63} & \multicolumn{1}{r|}{72} & \multicolumn{1}{r|}{95} & \multicolumn{1}{r|}{86} & \multicolumn{1}{r|}{90} & \multicolumn{1}{r|}{106} & \multicolumn{1}{r|}{118} & \multicolumn{1}{r}{140} \\
    \multicolumn{1}{l|}{} & \multicolumn{1}{l|}{$S_A$} & \multicolumn{1}{r|}{2.46} & \multicolumn{1}{r|}{3.25} & \multicolumn{1}{r|}{3.84} & \multicolumn{1}{r|}{4.57} & \multicolumn{1}{r|}{5.32} & \multicolumn{1}{r|}{6.64} & \multicolumn{1}{r|}{7.40} & \multicolumn{1}{r|}{8.32} & \multicolumn{1}{r|}{8.88} & \multicolumn{1}{r|}{9.55} & \multicolumn{1}{r|}{10.36} & \multicolumn{1}{r|}{11.81} & \multicolumn{1}{r}{14.75} \\
    \multicolumn{1}{l|}{} & \multicolumn{1}{l|}{$S_D$} & \multicolumn{1}{r|}{1.00} & \multicolumn{1}{r|}{1.30} & \multicolumn{1}{r|}{1.41} & \multicolumn{1}{r|}{1.76} & \multicolumn{1}{r|}{1.94} & \multicolumn{1}{r|}{2.33} & \multicolumn{1}{r|}{2.64} & \multicolumn{1}{r|}{2.95} & \multicolumn{1}{r|}{2.94} & \multicolumn{1}{r|}{3.30} & \multicolumn{1}{r|}{3.44} & \multicolumn{1}{r|}{4.26} & \multicolumn{1}{r}{4.68} \\ \hline
    \multicolumn{1}{l|}{OnHW-wordsRandom} & \multicolumn{1}{l|}{$L_A$} & \multicolumn{1}{r|}{111} & \multicolumn{1}{r|}{167} & \multicolumn{1}{r|}{201} & \multicolumn{1}{r|}{240} & \multicolumn{1}{r|}{290} & \multicolumn{1}{r|}{340} & \multicolumn{1}{r|}{397} & \multicolumn{1}{r|}{438} & \multicolumn{1}{r|}{493} & \multicolumn{1}{r|}{538} & \multicolumn{1}{r|}{598} & \multicolumn{1}{r|}{648} & \multicolumn{1}{r}{703} \\
    \multicolumn{1}{l|}{} & \multicolumn{1}{l|}{$L_D$} & \multicolumn{1}{r|}{41} & \multicolumn{1}{r|}{80} & \multicolumn{1}{r|}{86} & \multicolumn{1}{r|}{94} & \multicolumn{1}{r|}{112} & \multicolumn{1}{r|}{128} & \multicolumn{1}{r|}{158} & \multicolumn{1}{r|}{165} & \multicolumn{1}{r|}{179} & \multicolumn{1}{r|}{194} & \multicolumn{1}{r|}{209} & \multicolumn{1}{r|}{222} & \multicolumn{1}{r}{234} \\
    \multicolumn{1}{l|}{} & \multicolumn{1}{l|}{$S_A$} & \multicolumn{1}{r|}{2.83} & \multicolumn{1}{r|}{3.89} & \multicolumn{1}{r|}{4.64} & \multicolumn{1}{r|}{5.25} & \multicolumn{1}{r|}{5.95} & \multicolumn{1}{r|}{6.82} & \multicolumn{1}{r|}{7.69} & \multicolumn{1}{r|}{8.50} & \multicolumn{1}{r|}{9.23} & \multicolumn{1}{r|}{10.08} & \multicolumn{1}{r|}{11.11} & \multicolumn{1}{r|}{11.74} & \multicolumn{1}{r}{13.01} \\
    \multicolumn{1}{l|}{} & \multicolumn{1}{l|}{$S_D$} & \multicolumn{1}{r|}{1.07} & \multicolumn{1}{r|}{1.41} & \multicolumn{1}{r|}{1.72} & \multicolumn{1}{r|}{1.80} & \multicolumn{1}{r|}{2.11} & \multicolumn{1}{r|}{2.34} & \multicolumn{1}{r|}{2.65} & \multicolumn{1}{r|}{2.96} & \multicolumn{1}{r|}{3.12} & \multicolumn{1}{r|}{3.42} & \multicolumn{1}{r|}{3.56} & \multicolumn{1}{r|}{3.82} & \multicolumn{1}{r}{4.04} \\
    \multicolumn{1}{c|}{} & \multicolumn{1}{c|}{} & \multicolumn{1}{c|}{\textbf{15}} & \multicolumn{1}{c|}{\textbf{16}} & \multicolumn{1}{c|}{\textbf{17}} & \multicolumn{1}{c|}{\textbf{18}} & \multicolumn{1}{c|}{\textbf{19}} & \multicolumn{1}{c|}{\textbf{20}} & \multicolumn{1}{c|}{\textbf{21}} & \multicolumn{1}{c|}{\textbf{22}} & \multicolumn{1}{c|}{\textbf{23}} & \multicolumn{1}{c|}{\textbf{24}} & \multicolumn{1}{c|}{\textbf{25}} & \multicolumn{1}{c|}{\textbf{26}} & \multicolumn{1}{c}{\textbf{27}} \\ \cline{1-15}
    \multicolumn{1}{l|}{OnHW-equations} & \multicolumn{1}{l|}{$L_A$} & \multicolumn{1}{r|}{1,080} & \multicolumn{1}{c|}{-} & \multicolumn{1}{c|}{-} & \multicolumn{1}{c|}{-} & \multicolumn{1}{c|}{-} & \multicolumn{1}{c|}{-} & \multicolumn{1}{c|}{-} & \multicolumn{1}{c|}{-} & \multicolumn{1}{c|}{-} & \multicolumn{1}{c|}{-} & \multicolumn{1}{c|}{-} & \multicolumn{1}{c|}{-} & \multicolumn{1}{c}{-} \\
    \multicolumn{1}{l|}{} & \multicolumn{1}{l|}{$L_D$} & \multicolumn{1}{r|}{221} & \multicolumn{1}{c|}{-} & \multicolumn{1}{c|}{-} & \multicolumn{1}{c|}{-} & \multicolumn{1}{c|}{-} & \multicolumn{1}{c|}{-} & \multicolumn{1}{c|}{-} & \multicolumn{1}{c|}{-} & \multicolumn{1}{c|}{-} & \multicolumn{1}{c|}{-} & \multicolumn{1}{c|}{-} & \multicolumn{1}{c|}{-} & \multicolumn{1}{c}{-} \\
    \multicolumn{1}{l|}{} & \multicolumn{1}{l|}{$S_A$} & \multicolumn{1}{r|}{18.90} & \multicolumn{1}{c|}{-} & \multicolumn{1}{c|}{-} & \multicolumn{1}{c|}{-} & \multicolumn{1}{c|}{-} & \multicolumn{1}{c|}{-} & \multicolumn{1}{c|}{-} & \multicolumn{1}{c|}{-} & \multicolumn{1}{c|}{-} & \multicolumn{1}{c|}{-} & \multicolumn{1}{c|}{-} & \multicolumn{1}{c|}{-} & \multicolumn{1}{c}{-} \\
    \multicolumn{1}{l|}{} & \multicolumn{1}{l|}{$S_D$} & \multicolumn{1}{r|}{3.15} & \multicolumn{1}{c|}{-} & \multicolumn{1}{c|}{-} & \multicolumn{1}{c|}{-} & \multicolumn{1}{c|}{-} & \multicolumn{1}{c|}{-} & \multicolumn{1}{c|}{-} & \multicolumn{1}{c|}{-} & \multicolumn{1}{c|}{-} & \multicolumn{1}{c|}{-} & \multicolumn{1}{c|}{-} & \multicolumn{1}{c|}{-} & \multicolumn{1}{c}{-} \\ \cline{1-15}
    \multicolumn{1}{l|}{OnHW-words500} & \multicolumn{1}{l|}{$L_A$} & \multicolumn{1}{r|}{609} & \multicolumn{1}{r|}{608} & \multicolumn{1}{c|}{-} & \multicolumn{1}{r|}{707} & \multicolumn{1}{r|}{712} & \multicolumn{1}{c|}{-} & \multicolumn{1}{c|}{-} & \multicolumn{1}{c|}{-} & \multicolumn{1}{c|}{-} & \multicolumn{1}{c|}{-} & \multicolumn{1}{c|}{-} & \multicolumn{1}{c|}{-} & \multicolumn{1}{c}{-} \\
    \multicolumn{1}{l|}{} & \multicolumn{1}{l|}{$L_D$} & \multicolumn{1}{r|}{115} & \multicolumn{1}{r|}{119} & \multicolumn{1}{c|}{-} & \multicolumn{1}{r|}{124} & \multicolumn{1}{r|}{144} & \multicolumn{1}{c|}{-} & \multicolumn{1}{c|}{-} & \multicolumn{1}{c|}{-} & \multicolumn{1}{c|}{-} & \multicolumn{1}{c|}{-} & \multicolumn{1}{c|}{-} & \multicolumn{1}{c|}{-} & \multicolumn{1}{c}{-} \\
    \multicolumn{1}{l|}{} & \multicolumn{1}{l|}{$S_A$} & \multicolumn{1}{r|}{14.41} & \multicolumn{1}{r|}{15.27} & \multicolumn{1}{c|}{-} & \multicolumn{1}{r|}{16.96} & \multicolumn{1}{r|}{15.58} & \multicolumn{1}{c|}{-} & \multicolumn{1}{c|}{-} & \multicolumn{1}{c|}{-} & \multicolumn{1}{c|}{-} & \multicolumn{1}{c|}{-} & \multicolumn{1}{c|}{-} & \multicolumn{1}{c|}{-} & \multicolumn{1}{c}{-} \\
    \multicolumn{1}{l|}{} & \multicolumn{1}{l|}{$S_D$} & \multicolumn{1}{r|}{4.52} & \multicolumn{1}{r|}{3.94} & \multicolumn{1}{c|}{-} & \multicolumn{1}{r|}{5.16} & \multicolumn{1}{r|}{5.01} & \multicolumn{1}{c|}{-} & \multicolumn{1}{c|}{-} & \multicolumn{1}{c|}{-} & \multicolumn{1}{c|}{-} & \multicolumn{1}{c|}{-} & \multicolumn{1}{c|}{-} & \multicolumn{1}{c|}{-} & \multicolumn{1}{c}{-} \\ \cline{1-15}
    \multicolumn{1}{l|}{OnHW-wordsRandom} & \multicolumn{1}{l|}{$L_A$} & \multicolumn{1}{r|}{748} & \multicolumn{1}{r|}{762} & \multicolumn{1}{r|}{854} & \multicolumn{1}{r|}{891} & \multicolumn{1}{r|}{922} & \multicolumn{1}{r|}{1,018} & \multicolumn{1}{r|}{967} & \multicolumn{1}{r|}{982} & \multicolumn{1}{r|}{1,019} & \multicolumn{1}{r|}{1,199} & \multicolumn{1}{r|}{1,078} & \multicolumn{1}{r|}{1,087} & \multicolumn{1}{r}{737} \\
    \multicolumn{1}{l|}{} & \multicolumn{1}{l|}{$L_D$} & \multicolumn{1}{r|}{237} & \multicolumn{1}{r|}{233} & \multicolumn{1}{r|}{266} & \multicolumn{1}{r|}{260} & \multicolumn{1}{r|}{249} & \multicolumn{1}{r|}{250} & \multicolumn{1}{r|}{221} & \multicolumn{1}{r|}{209} & \multicolumn{1}{r|}{193} & \multicolumn{1}{r|}{206} & \multicolumn{1}{r|}{355} & \multicolumn{1}{r|}{143} & \multicolumn{1}{r}{0} \\
    \multicolumn{1}{l|}{} & \multicolumn{1}{l|}{$S_A$} & \multicolumn{1}{r|}{13.93} & \multicolumn{1}{r|}{14.22} & \multicolumn{1}{r|}{15.64} & \multicolumn{1}{r|}{16.19} & \multicolumn{1}{r|}{17.32} & \multicolumn{1}{r|}{18.61} & \multicolumn{1}{r|}{17.91} & \multicolumn{1}{r|}{19.88} & \multicolumn{1}{r|}{20.00} & \multicolumn{1}{r|}{25.25} & \multicolumn{1}{r|}{23.33} & \multicolumn{1}{r|}{25.00} & \multicolumn{1}{r}{1.00} \\
    \multicolumn{1}{l|}{} & \multicolumn{1}{l|}{$S_D$} & \multicolumn{1}{r|}{4.45} & \multicolumn{1}{r|}{4.97} & \multicolumn{1}{r|}{5.03} & \multicolumn{1}{r|}{5.04} & \multicolumn{1}{r|}{5.48} & \multicolumn{1}{r|}{5.51} & \multicolumn{1}{r|}{6.14} & \multicolumn{1}{r|}{7.17} & \multicolumn{1}{r|}{5.41} & \multicolumn{1}{r|}{6.67} & \multicolumn{1}{r|}{5.25} & \multicolumn{1}{r|}{2.33} & \multicolumn{1}{r}{0.00} \\
    \end{tabular}
\end{center}
\end{table*}

\subsection{Exemplary Sensor Data}
\label{sec_appendix_sensor}

Figure~\ref{image_app_exe_equ} and Figure~\ref{image_app_exe_word} show the sensor data of the 13 channels for an exemplary equation and words written on normal paper and on tablet. The accelerometer data is given in $m/s^2$, the gyroscope data in $^\circ/s$, the magnetometer data in $mT$, and the force sensor in $N$. The equation sample consists of 567 timesteps, while the word sample on paper consists of 217 timesteps and on tablet of 402 timesteps. It can be shown that for all three samples the single strokes can be clearly separated through the force sensor (see Figure~\ref{image_app_exe_equ_4}). By comparing the accelerometer and gyroscope data of a selected word written on normal paper (see Figure~\ref{image_app_exe_word_1} and \ref{image_app_exe_word_2}) with the word written on tablet (see Figure~\ref{image_app_exe_word_tablet_1} and \ref{image_app_exe_word_tablet_2}), we can see that the surface of the paper introduces higher sensor noise than the surface of the tablet.

\subsection{Statistics of the Datasets}
\label{sec_appendix_statistics}

The characteristics of a dataset influences the behavior of a deep learning model. If the deployed context does not match the evaluation datasets, a model is unlikely to perform well. Hence, we will propose more detailed statistics of our proposed datasets, in this section, while we already compared our datasets with state-of-the-art datasets in Section~\ref{chap_eval_data}. Table~\ref{table_length_datasets_overview} proposes average and deviation timesteps and number of strokes for each sample length of the OnHW-equations, OnHW-words500 and OnHW-wordsRandom datasets. The number of timesteps per sample is significantly larger for the OnHW-equations dataset than for the words datasets. We can conclude that writing numbers and symbols requires more time as words are mostly written in cursive font, while equations are written in printed font. Hence, the deviation of timesteps is also larger for equations. The deviation in timestep lengths is important as the data has to be split by the CTC loss, and a larger deviation leads to more split varieties. Additionally, the number of strokes per sample is an significant feature for the classification task, which can be learned by the model from the force sensor data. The average number of strokes is clearly larger (about one to three strokes) for the OnHW-equations dataset than the words datasets, while the deviation of strokes is less. We can state from that, that number and symbols requires many strokes for printed writing, while cursive writing of words leads to less number of strokes with a use-specific writing style. Hence, training a model for a writer-independent classification task is more difficult. To split the OnHW-equations dataset into single symols and numbers, we use the following split constraints, where the possible number of strokes per character is \texttt{0} [1], \texttt{1} [1], \texttt{2} [1], \texttt{3} [1], \texttt{4} [1,2], \texttt{5} [2], \texttt{6} [1], \texttt{7} [1,2], \texttt{8} [1], \texttt{9} [1], \texttt{+} [2], \texttt{-} [1], $\cdot$ [1], \texttt{:} [2] and \texttt{=} [2].

\begin{figure*}[t!]
	\centering
    \includegraphics[width=1.0\linewidth]{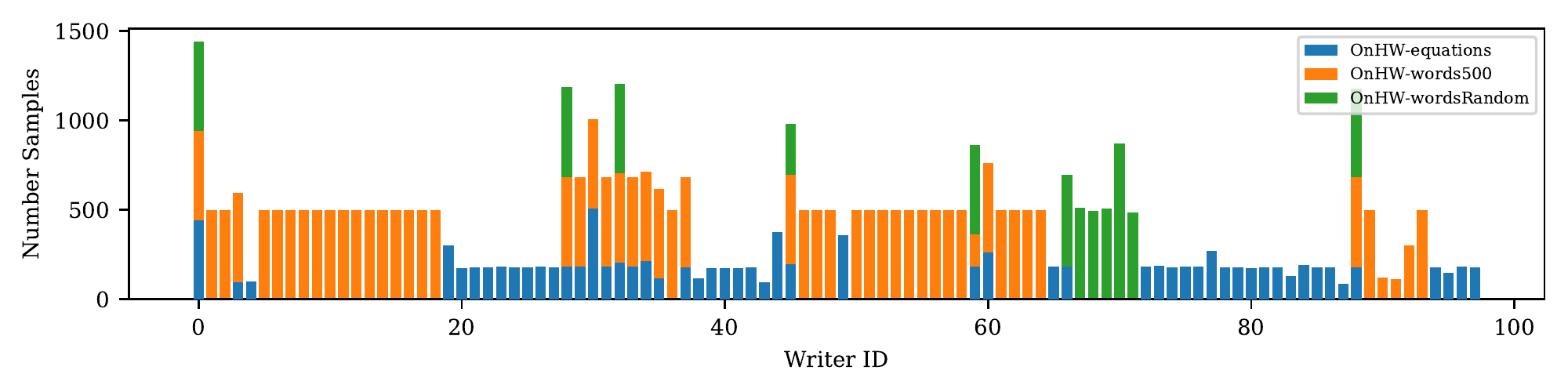}
    \vspace{-0.4cm}
    \caption{Overview of writer IDs contributed to the OnHW-equations, OnHW-words500, and OnHW-wordsRandom datasets.}
    \label{image_writer_overview}
\end{figure*}

\begin{figure*}[t!]
	\centering
    \includegraphics[width=1.0\linewidth]{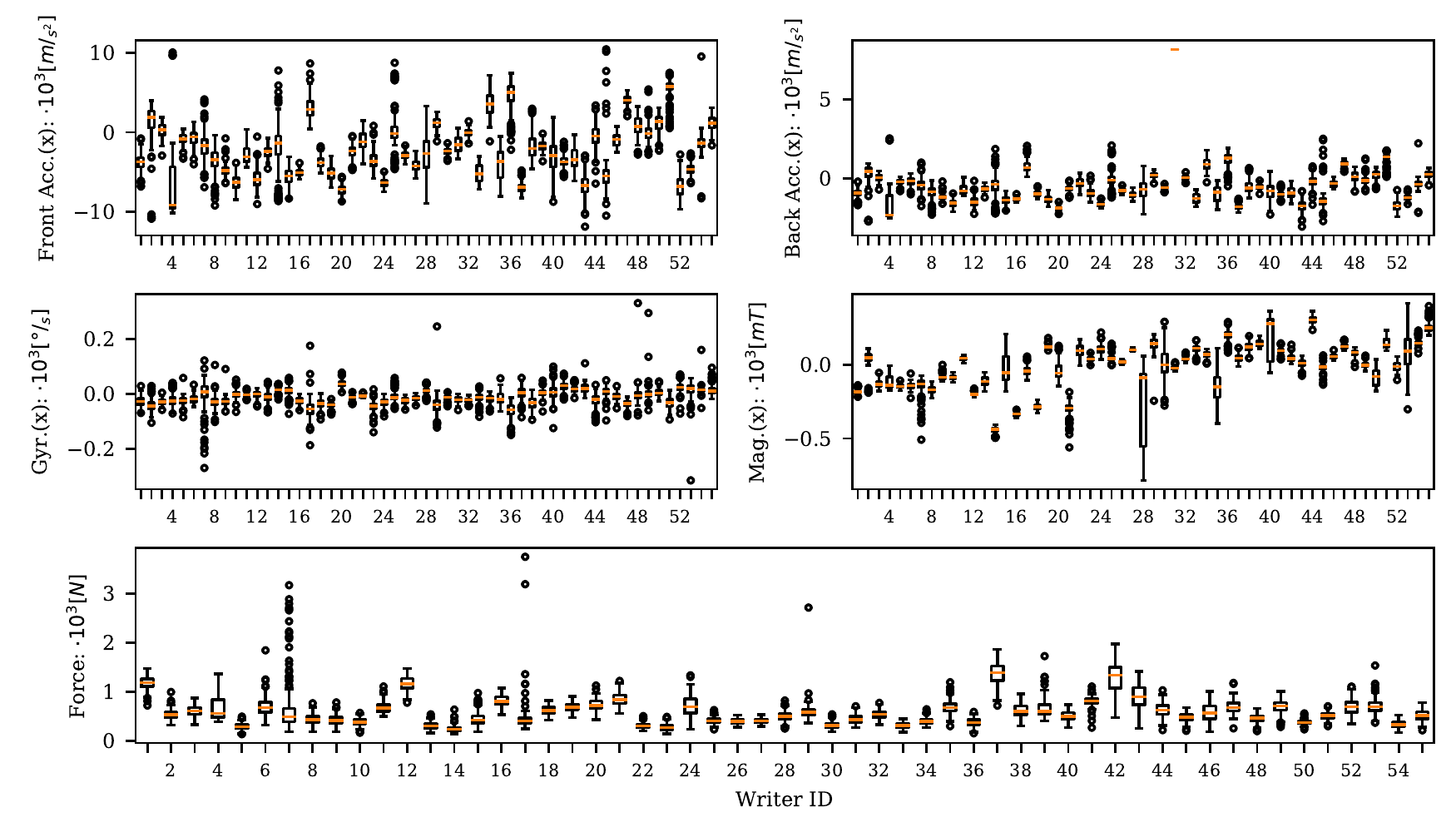}
    \vspace{-0.3cm}
    \caption{Overview of mean sensor distribution per writer for the OnHW-equations dataset.}
    \label{image_writer_overview_sensor}
\end{figure*}

Figure~\ref{image_writer_overview} gives an overview of writers contributed to the sequence-based datasets. For the OnHW-equations dataset most participants wrote about 180 equations, while for thw OnHW-words500 and OnHW-wordsRandom each writer contributed about 500 words. This leads to a equally balanced dataset and a proper writer-independent evaluation.

Not only the diversity of the samples per participant is important, but also the diversity of the sensor data. Especially, out-of-distribution sensor data from one writer can decrease classification accuracy. Figure~\ref{image_writer_overview_sensor} gives an overview of the mean distribution per writer for the x-axis of the sensors. For the force data, the writers with IDs 7 and 17 have many outliers, while the writer with IDs 12, 37 and 42 press the pen tip strongly to the paper. While the front accelerometer data is very diverse between $-10^3$ to $10^3$ (e.g., writer 14, 25 and 45 with many outliers, against writer 16, 19 and 37 with consistent sensor data), the movement of the rear accelerometer is slower between $-4 \cdot 10^3$ to $3 \cdot 10^3$, as the pen tip typically moves faster than the rear accelerometer. The gyroscope distribution per writer draws conclusion of the writing style. These findings lead to the conclusion that the writer-dependent problem is an easier classification task than the writer-independent problem.

\subsection{Transformer Parameters and Hyperparameter Searches}
\label{sec_appendix_trans_parameters}

\paragraph{Transformer Parameters.} This section describes the Transformer parameters. For our attention-based model, we search for the optimal parameters $d_\text{model} = [150, 300]$, $d_k = [32,64]$, $d_v = [32, 64]$, the number of multi-head attentions $n_\text{head} = [3,4,5]$, and a convolutional factor $c_\text{fac} = [4, 6, 8, 10, 16]$, while the network consists of the 1D convolutions $(c_\text{fac}, 2 \cdot c_\text{fac}, 4 \cdot c_\text{fac}, 8 \cdot c_\text{fac})$ and the (Bi)LSTM layers $(4 \cdot c_\text{fac}, 2 \cdot c_\text{fac})$. We train only 500 epochs, as each training takes 4\textit{h}. We choose $d_\text{model} = 150$, $d_k = 32$, $d_v = 64$, $c_\text{fac} = 6$, with BiLSTM and time distribution for follow-up trainings. The number of heads $\text{n}_{\text{heads}}$ is 3. We apply the Transformer variants Perceiver~\cite{jaegle}, Sinkhorn Transformer~\cite{tay_bahri}, Performer~\cite{choromanski}, Reformer~\cite{kitaev} and Linformer~\cite{wang_li} to the single character classification task with the following parameters. We choose non reversible Transformers without a language model or a lexicon. The input is the inertial MTS. We evaluated different combinations of last layers for all variants, i.e., with and without 1D convolution or 1D max pooling. The best results yielded a permutation with a 1D max pooling of kernel size 5 and stride 5, in combination with a linear layer of size $(\text{in}_{\text{dim}}, \text{n}_{\text{classes}})$. For the Perceiver~\cite{jaegle}, we set $\text{cross}_{\text{heads}} = 1$, $\text{num}_{\text{freq}} = 4$, $\text{depth} = 2$, $\text{num}_{\text{latents}} = 64$, $\text{latent}_{\text{dim,heads}} = 128$, $\text{max}_{\text{frequ}} = 10$, and $\text{latent}_{\text{heads}} = 4$. We set $\text{attn}_{\text{drop}}$ and $\text{ff}_{\text{drop}}$ to 0.2. $\text{n}_{\text{classes}}$ depends on the dataset, and is for the OnHW-symbols and OnHW-equations dataset 15, and for the OnHW-chars dataset 26 for the lower and upper datasets and 52 for the combined dataset. We choose the parameters of the Sinkhorn Transformer~\cite{tay_bahri} $\text{dim} = 1,024$, $\text{heads} = 8$, $\text{depth} = 12$, $\text{dim}_{\text{head}} = 6$, and $\text{bucket}_{\text{size}} = 20$. For the Performer~\cite{choromanski}, we choose $\text{dim} = 512$, $\text{depth} = 1$, $\text{heads} = 5$, $\text{dim}_{\text{head}} = 4$, and $\text{causal} = \text{True}$. The parameters of the Reformer~\cite{kitaev} are $\text{dim} = 128$, $\text{heads} = 8$, $\text{bucket}_{\text{size}} = 20$, $\text{dim}_{\text{head}} = 6$, $\text{depth} = 12$, $\text{lsh}_{\text{drop}} = 0.1$, and $\text{causal} = \text{True}$. We set for the Linformer Transformer~\cite{wang_li} the parameters $\text{dim} = 512$, $\text{seq}_{\text{len}} = 79$ for split OnHW-equations and OnHW-symbols and $\text{seq}_{\text{len}} = 64$ for OnHW-chars~\cite{ott}, $\text{depth} = 12$, $\text{heads} = 5$, $\text{share}_{\text{kv}} = \text{True}$, and $\text{k} = 256$. For the Sinkhorn and Reformer Transformers, the sequence length has to be divisible by the bucket size. For Performer and Linformer Transformers, the input dimension has to be divisible by the number of heads, and hence, we exclude the magnetometer data.

\begin{figure}[t!]
    \includegraphics[trim=25 35 25 70, clip,width=1\linewidth]{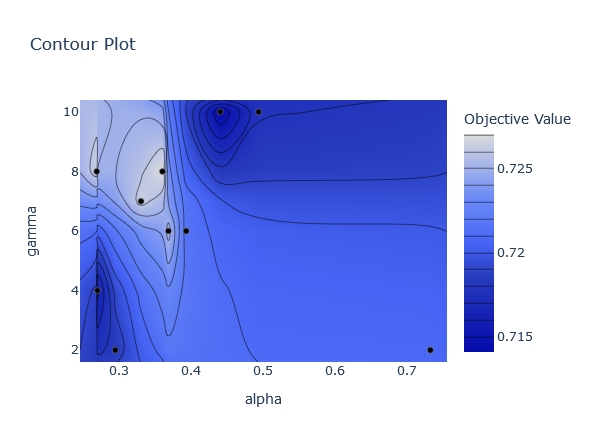}
    \caption{Parameter search for $\alpha$ and $\gamma$ in the Focal loss \cite{lin_goyal} with Optuna.}
    \label{figure_introduction}
\end{figure}

\paragraph{Hyperparameter Search.} We search for the Focal loss \cite{lin_goyal} for the class balance factor $\alpha \in [0,1]$ and $\gamma \geq 0$ in the modulating factor $(1 - p_i)^{\gamma}$. We use the combined OnHW-chars (WI) dataset. Figure~\ref{figure_introduction} shows the hyperparameter search for $\alpha$ and $\gamma$ with Optuna\footnote{Optuna: \url{https://optuna.org/}}. The objective value is the character recognition rate. The optimal parameters are $\alpha = 0.75$ and a large $\gamma = 8$. Note that the search space is in a small range between 71\% and 73\%. We use these parameters, for follow-up trainings.

\subsection{Detailed Evaluation}
\label{sec_appendix_detailed_evaluation}

\begin{figure*}[t!]
	\centering
	\begin{minipage}[b]{0.49\linewidth}
        \centering
        \includegraphics[trim=18 0 24 0, clip, width=0.8\linewidth]{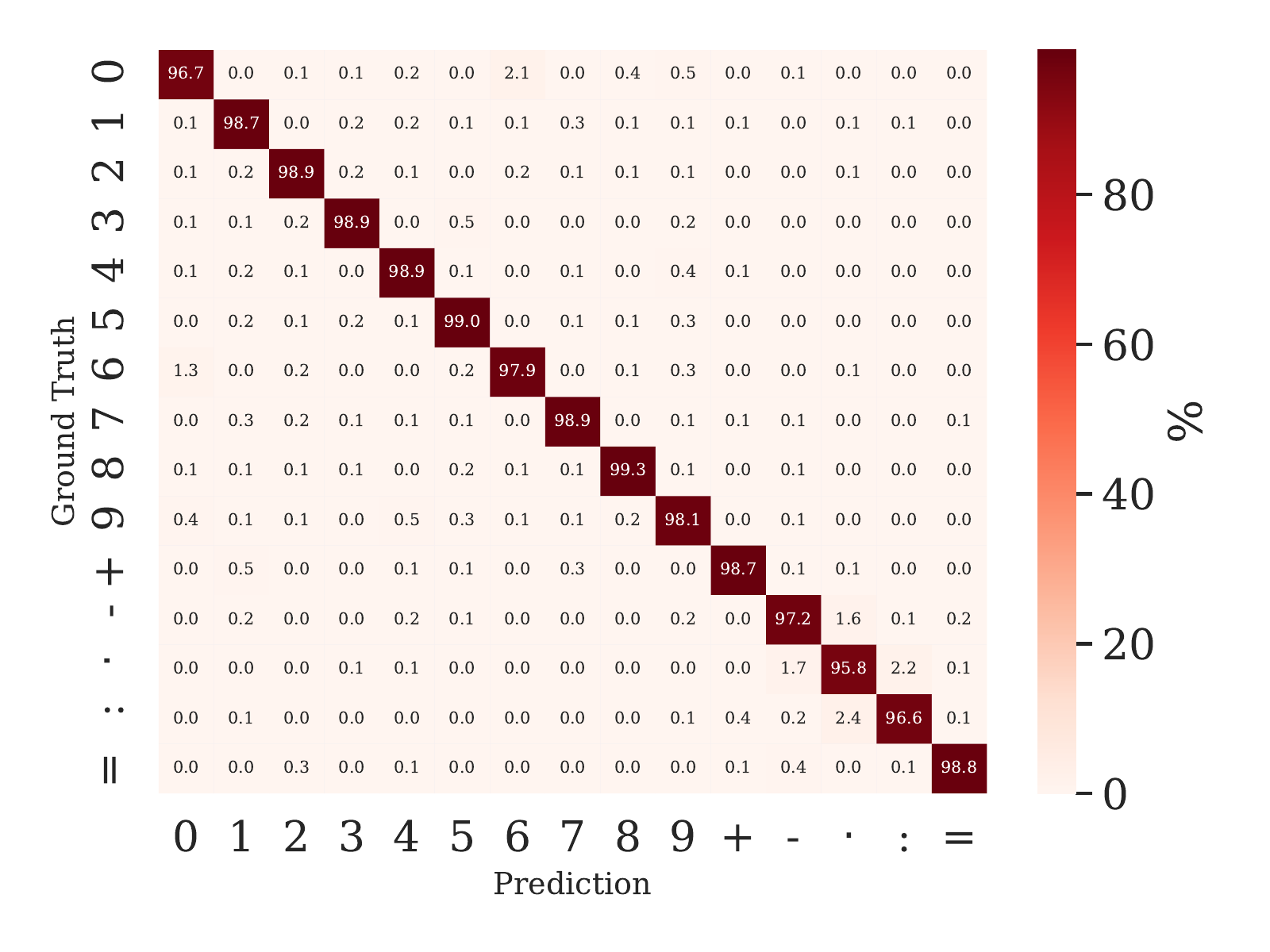}
        \subcaption{OnHW-equations dataset.}
        \label{image_confusion_matrix1}
    \end{minipage}
    \hfill
	\begin{minipage}[b]{0.49\linewidth}
        \centering
        \includegraphics[trim=4 0 24 0, clip, width=0.8\linewidth]{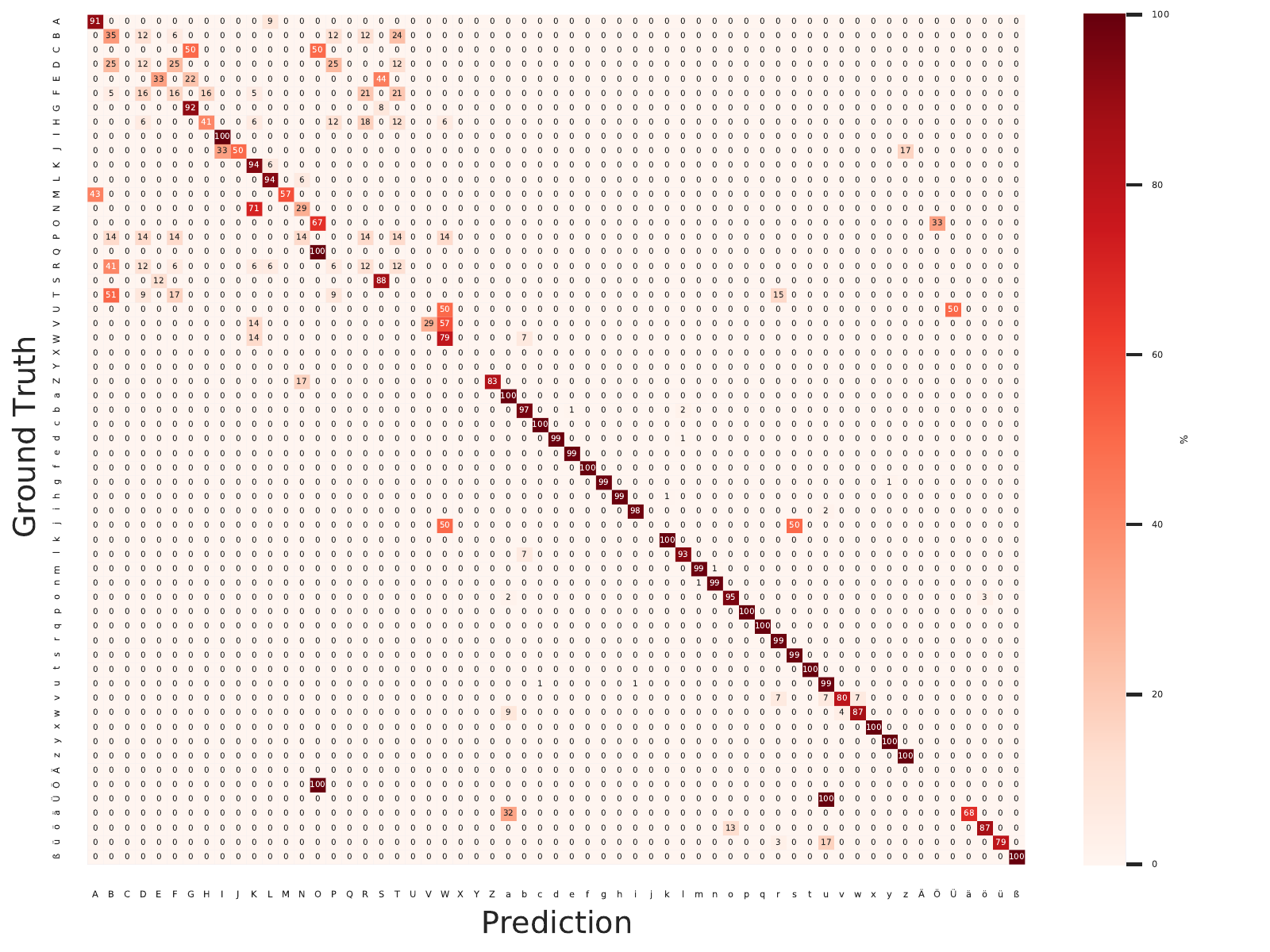}
        \subcaption{OnHW-wordsTraj dataset.}
        \label{image_confusion_matrix2}
    \end{minipage}
    \caption{Confusion matrices for mismatches.}
    \label{image_confusion_matrix}
    %\vspace{-0.15cm}
\end{figure*}

\begin{figure*}[t!]
	\centering
	\begin{minipage}[b]{0.49\linewidth}
        \centering
    	\includegraphics[width=1.0\linewidth]{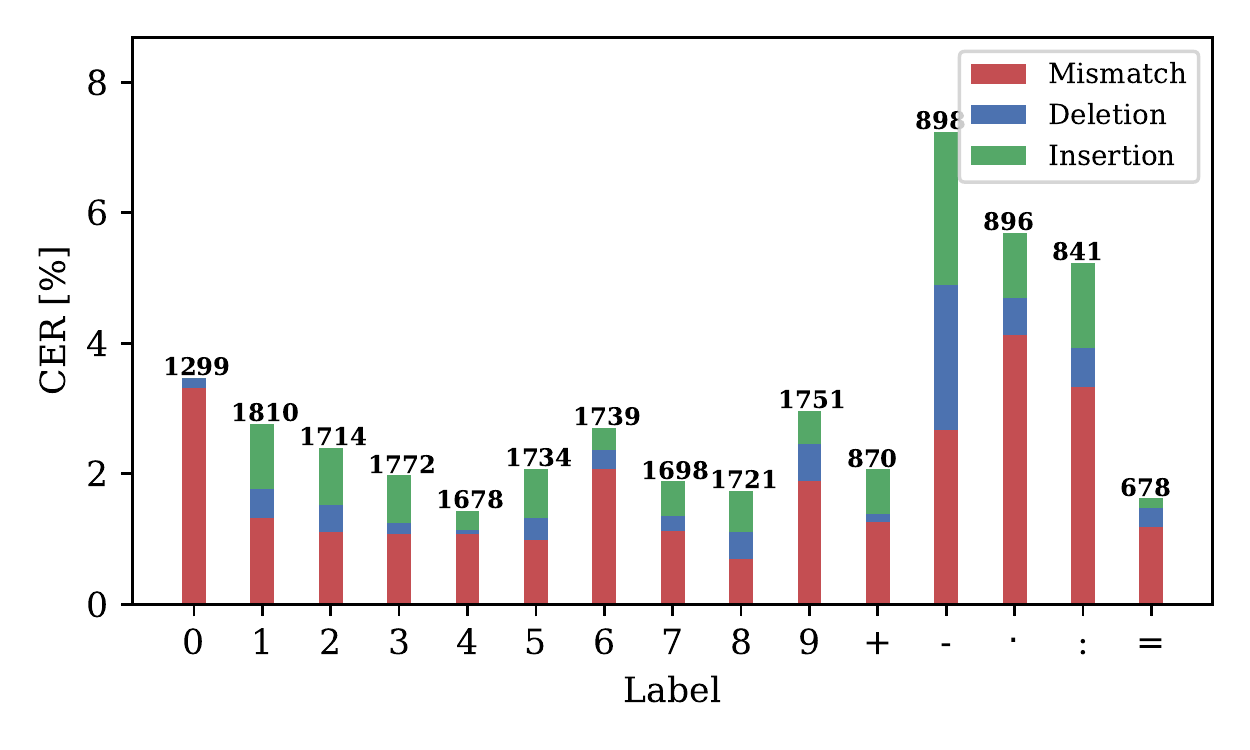}
        \vspace{-0.1cm}
    	\subcaption{OnHW-equations dataset.}
    	\label{image_eval_ed1}
    \end{minipage}
    \hfill
	\begin{minipage}[b]{0.49\linewidth}
        \centering
    	\includegraphics[width=1.0\linewidth]{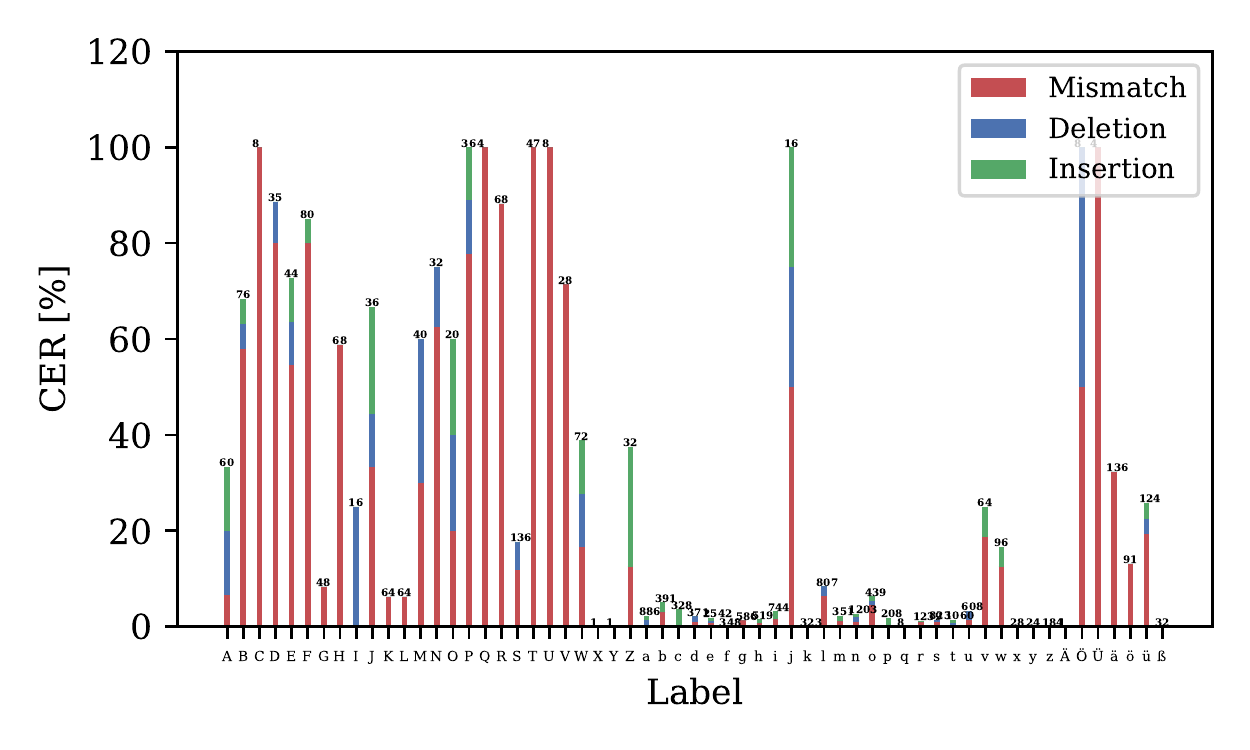}
        \vspace{-0.1cm}
    	\subcaption{OnHW-wordsTraj dataset.}
    	\label{image_eval_ed2}
    \end{minipage}
    \caption{Evaluation of the Edit distance with mismatch, deletion and insertion for every character prediction.}
    \label{image_eval_ed}
\end{figure*}

\paragraph{Evaluation of the Accuracy per Label.} Figure~\ref{image_confusion_matrix} shows confusion matrices for sequence-based classification tasks for the accuracy of predicted single class labels regarding the ground truth class labels in \%. For the OnHW-equations dataset (see Figure~\ref{image_confusion_matrix1}), the accuracies per labels are between 96.6\% and 99.3\%. While the ground truth \texttt{'0'} is confused with \texttt{'6'} and \texttt{'9'} because of the similar round shape of these numbers, the \texttt{'-'} is misclassified with the '$\cdot$' as both symbols are short samples, and the \texttt{':'} is misclassified with a single dot '$\cdot$'. From analyzing the confusion matrix of the OnHW-wordsTraj (see Figure~\ref{image_confusion_matrix2}) dataset, we see two significant patterns. First, small letters are highly accurate starting from 80\% (see the second part of the diagonal), while only \texttt{'j'} is misclassified with \texttt{'W'} and \texttt{'s'}. Second, capital letters are highly incorrect (see the first part of the diagonal). Letters as \texttt{'C'}, \texttt{'P'} and \texttt{'T'} are indistinguishable from other letters, while \texttt{'Q'} and \texttt{'O'} are interchanged. The reason is the under-representation of capital letters in the dataset (see Figure~\ref{figure_statistics_dataset3}), as capital letters only appear at the starting letter of a word in German. By plotting the confusion matrix for the OnHW-wordsRandom dataset, the mismatches for capital letters improve compared to the OnHW-wordsTraj dataset, but is still significantly higher compared with small letters. The vowel mutations (\texttt{'ä'}, \texttt{'ö'}, \texttt{'ü'}, \texttt{'Ä'}, \texttt{'Ö'}, \texttt{'Ü'}) are also highly under-represented, and the classification accuracy highly decreases. Figure~\ref{image_eval_ed} separates the mismatches, deletions and insertions per labels (see Equ.~\eqref{equ_edit_distance}). The number on top of the box plot indicates the amount of occurrences in the validation test. For the OnHW-equations dataset (see Figure~\ref{image_eval_ed1}), the number \texttt{'0'} does not have to be inserted, and number \texttt{'4'} does not have to be deleted. It is significant that that the symbols with less timesteps \texttt{'-'}, '$\cdot$' and \texttt{':'} are often mismatched and missed, while only \texttt{'-'} has to be deleted. Numbers are distinguishable more easily. For the OnHW-wordsTraj (see Figure~\ref{image_eval_ed2}), again, the CER of capital letters is considerably higher than small letters. While some letters, i.e., \texttt{'C'}, \texttt{'Q'}, \texttt{'T'}, \texttt{'U'}, and \texttt{'V'}, are only mismatched, other edit errors appear for \texttt{'A'}, \texttt{'I'}, \texttt{'J'}, and \texttt{'Z'}. A dataset with more capital letters could mitigate these errors.

\paragraph{Evaluation of Sample Length Dependent Edit Distance.} We show the sample length dependent counts of wrong predictions, i.e., mismatches, insertions and deletions, for the OnHW-equations (see Figure~\ref{image_ed_length1}) and OnHW-wordsTraj (see Figure~\ref{image_ed_length2}) datasets. For the OnHW-equations dataset, a high appearance of mismatches and insertions appear at the starting and end character, while deletions emerge more even over the whole equations. As previously shown, the first character of words are significantly often mismatched or has to be inserted or deleted for the OnHW-wordsTraj dataset. This shows the unequal distribution of samples for the words datasets (see Figure~\ref{figure_statistics_dataset3}), while the equations dataset is very equally distributed (see Figure~\ref{figure_statistics_dataset2}).

\begin{table*}
\begin{center}
\setlength{\tabcolsep}{3.3pt}
    \caption{State-of-the-art evaluation results in \% for the online IAM-OnDB~\cite{liwicki}, VNOnDB-words~\cite{nguyen}, and IBM-UB-1~\cite{shivram} datasets.}
    \label{table_results4}
    \small \begin{tabular}{ p{0.5cm} | p{0.5cm} | p{0.5cm} | p{0.5cm} | p{0.5cm} | p{0.5cm} | p{0.5cm} }
    \multicolumn{1}{c|}{\textbf{Method}} & \multicolumn{2}{c|}{\textbf{IAM-OnDB}~\cite{liwicki}} & \multicolumn{2}{c|}{\textbf{VNOnDB}~\cite{nguyen}} & \multicolumn{2}{c}{\textbf{IBM\_UB\_1}~\cite{shivram}} \\
    \multicolumn{1}{c|}{} & \multicolumn{1}{c}{\textbf{WER}} & \multicolumn{1}{c|}{\textbf{CER}} & \multicolumn{1}{c}{\textbf{WER}} & \multicolumn{1}{c|}{\textbf{CER}} & \multicolumn{1}{c}{\textbf{WER}} & \multicolumn{1}{c}{\textbf{CER}} \\ \hline
    \multicolumn{1}{l|}{BiLSTM~\cite{carbune}$^2$} & \multicolumn{1}{r}{6.50} & \multicolumn{1}{r|}{2.50} & \multicolumn{1}{r}{12.20} & \multicolumn{1}{r|}{6.10} & \multicolumn{1}{r}{15.10} & \multicolumn{1}{r}{4.10} \\
    \multicolumn{1}{r|}{curve, w/o FF$^1$} & \multicolumn{1}{r}{18.60} & \multicolumn{1}{r|}{5.90} & \multicolumn{1}{r}{-} & \multicolumn{1}{r|}{-} & \multicolumn{1}{r}{25.10} & \multicolumn{1}{r}{6.00} \\
    \multicolumn{1}{r|}{curve, w/ FF$^1$} & \multicolumn{1}{r}{10.60} & \multicolumn{1}{r|}{4.00} & \multicolumn{1}{r}{-} & \multicolumn{1}{r|}{-} & \multicolumn{1}{r}{15.10} & \multicolumn{1}{r}{4.10} \\
    \multicolumn{1}{l|}{BiLSTM~\cite{frinken}} & \multicolumn{1}{r}{24.99} & \multicolumn{1}{r|}{12.26} & \multicolumn{1}{r}{-} & \multicolumn{1}{r|}{-} & \multicolumn{1}{r}{-} & \multicolumn{1}{r}{-} \\
    \multicolumn{1}{l|}{BiLSTM~\cite{graves}} & \multicolumn{1}{r}{20.30} & \multicolumn{1}{r|}{11.50} & \multicolumn{1}{r}{-} & \multicolumn{1}{r|}{-} & \multicolumn{1}{r}{-} & \multicolumn{1}{r}{-} \\
    \multicolumn{1}{l|}{LSTM~\cite{liwicki_bunke}} & \multicolumn{1}{r}{18.93} & \multicolumn{1}{r|}{-} & \multicolumn{1}{r}{-} & \multicolumn{1}{r|}{-} & \multicolumn{1}{r}{-} & \multicolumn{1}{r}{-} \\
    \multicolumn{1}{r|}{combination$^2$} & \multicolumn{1}{r}{13.84} & \multicolumn{1}{r|}{-} & \multicolumn{1}{r}{-} & \multicolumn{1}{r|}{-} & \multicolumn{1}{r}{-} & \multicolumn{1}{r}{-} \\
    \multicolumn{1}{l|}{BiLSTM~\cite{keysers}$^2$} & \multicolumn{1}{r}{26.70} & \multicolumn{1}{r|}{8.80} & \multicolumn{1}{r}{-} & \multicolumn{1}{r|}{-} & \multicolumn{1}{r}{22.20} & \multicolumn{1}{r}{6.70} \\
    \multicolumn{1}{l|}{Seg-and-Dec~\cite{keysers}$^2$} & \multicolumn{1}{r}{10.40} & \multicolumn{1}{r|}{4.30} & \multicolumn{1}{r}{-} & \multicolumn{1}{r|}{-} & \multicolumn{1}{r}{-} & \multicolumn{1}{r}{-} \\
    \multicolumn{1}{l|}{GoogleTask2$^4$} & \multicolumn{1}{r}{-} & \multicolumn{1}{r|}{-} & \multicolumn{1}{r}{19.00} & \multicolumn{1}{r|}{6.86} & \multicolumn{1}{r}{-} & \multicolumn{1}{r}{-} \\
    \multicolumn{1}{l|}{IVTOVTask2$^{3,4}$} & \multicolumn{1}{r}{-} & \multicolumn{1}{r|}{-} & \multicolumn{1}{r}{14.11} & \multicolumn{1}{r|}{3.24} & \multicolumn{1}{r}{-} & \multicolumn{1}{r}{-} \\
    \multicolumn{1}{l|}{MyScriptTask2\_1$^{3,4}$} & \multicolumn{1}{r}{-} & \multicolumn{1}{r|}{-} & \multicolumn{1}{r}{2.02} & \multicolumn{1}{r|}{1.02} & \multicolumn{1}{r}{-} & \multicolumn{1}{r}{-} \\
    \multicolumn{1}{l|}{MyScriptTask2\_2$^{3,4}$} & \multicolumn{1}{r}{-} & \multicolumn{1}{r|}{-} & \multicolumn{1}{r}{1.57} & \multicolumn{1}{r|}{4.02} & \multicolumn{1}{r}{-} & \multicolumn{1}{r}{-} \\
    \multicolumn{7}{l}{$^1$ Feature functions (FF)  $^2$ Open training set  $^3$ VieTreeBank (VTB) corpus} \\
    \multicolumn{7}{l}{$^4$ Results available: \href{https://github.com/undertheseanlp/NLP-Vietnamese-progress/blob/master/tasks/optical_text_recognition.md}{here}}
    \end{tabular}
\end{center}
\end{table*}

\paragraph{Evaluation Results of State-of-the-Art Techniques for Online HWR.} This section summarizes state-of-the-art results for the IAM-OnDB~\cite{liwicki}, VNOnDB~\cite{nguyen} and IBM-UB-1~\cite{shivram} datasets (see Table~\ref{table_results4}). Graves et al.~\cite{graves} (2008) started to improve the classification accuracy by proposing an alternative approach based on a RNN specifically designed for sequence labelling tasks where data contains long range interdependencies and that is hard to segment. Liwicki et al.~\cite{liwicki_bunke} (2011) introduced recognizers based on Hidden Markov Models and BiLSTMs, and on different set of features from online and offline data. Frinken et al.~\cite{frinken} (2015) showed that a deep BiLSTM neural network outperforms the standard BiLSTM model by combining \texttt{ReLU} activation with BiLSTM layers, but get a high WER of 24.99\% and a CER of 12.26\% on the IAM-OnDB dataset. Keysers et al.~\cite{keysers} (2017) used a training for feature combination, a trainable segmentation technique, unified time- and position-based input interpretation, and a cascade of pruning strategies. The method achieves a WER of 26.7\%  and a CER of 8.80\% with a BiLSTM, and up to 10.4\% WER and 4.30\% CER with a segmentation approach. The system is used in several Google products such as for translation. Carbune et al.~\cite{carbune} (2020) used bi-directional recurrent layers in combination with a softmax layer and the CTC loss. Their approach supports 102 languages. Hence, the architecture is based on a language model. The system combines methods from sequence recognition with a new input encoding using Bézier curves. This technique achieves the currently best results for the IAM-OnDB and IBM\_UB\_1 datasets. Feature functions (FF) introduces prior knowledge about the underlying language into the system. This method was used for the ICFHR2018 competition on Vietnamese online handwritten text recognition using VNOnDB. Along with this challenge, results from GoogleTask2, IVTOVTask2 and MyScriptTask2 are available, where MyScriptTask2\_2 achieves the lowest WER of 1.57\% on the VNOnDB dataset. This method uses a segmentation component with a feed-forward network along with BiLSTMs and the CTC loss. The IVTOVTask2 system also uses BiLSTM layers with the CTC loss similar to our approach. Unfortunately, public code is not available for these approaches. A direct comparison of these results with our results is not possible, as we used a 5-fold cross validation of the IAM-OnDB~\cite{liwicki} and VNOnDB-words~\cite{nguyen} datasets, different to the train/test splits used for public results. But for our task, we can differentiate between WD and WI classification tasks. With our best model CNN+BiLSTM we achieve a CER of 6.94\% (WD) and 9.11\% (WI) for the IAM-OnDB dataset that is better than the BiLSTM approaches by \cite{frinken,graves,keysers}, but worse than the BiLSTM by \cite{carbune} and the method by \cite{keysers}. On the VNOnDB-words dataset, our CER of 6.71\% (WD) and the WER of 15.54\% (WD) is lower than GoogleTask2, but higher than \cite{carbune} and MyScriptTask2.

\begin{figure*}[t!]
	\centering
	\begin{minipage}[b]{0.245\linewidth}
        \centering
    	\includegraphics[width=1.0\linewidth]{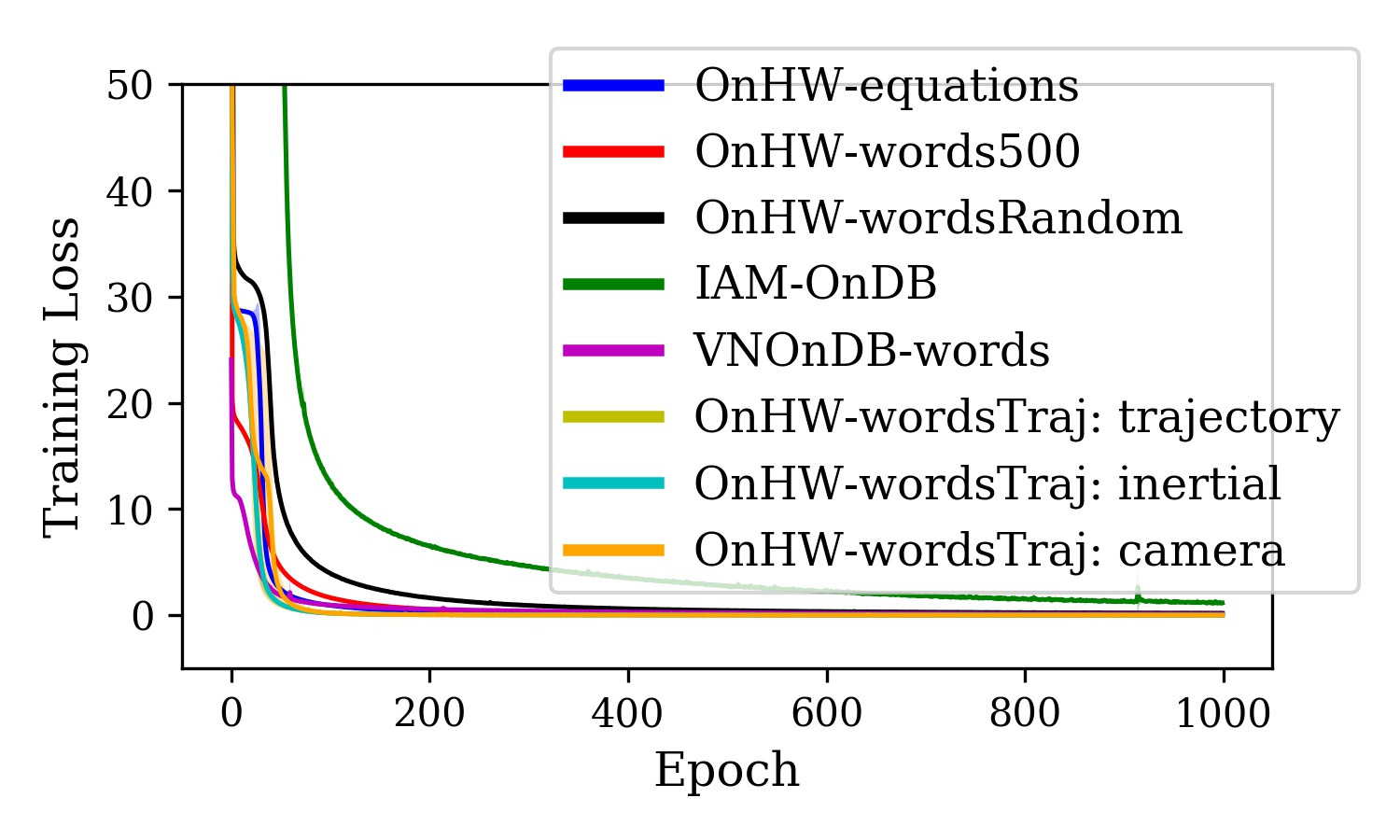}
        \vspace{-0.1cm}
    	\subcaption{Training loss (WD).}
    	\label{image_training_1}
    \end{minipage}
    \hfill
	\begin{minipage}[b]{0.245\linewidth}
        \centering
    	\includegraphics[width=1.0\linewidth]{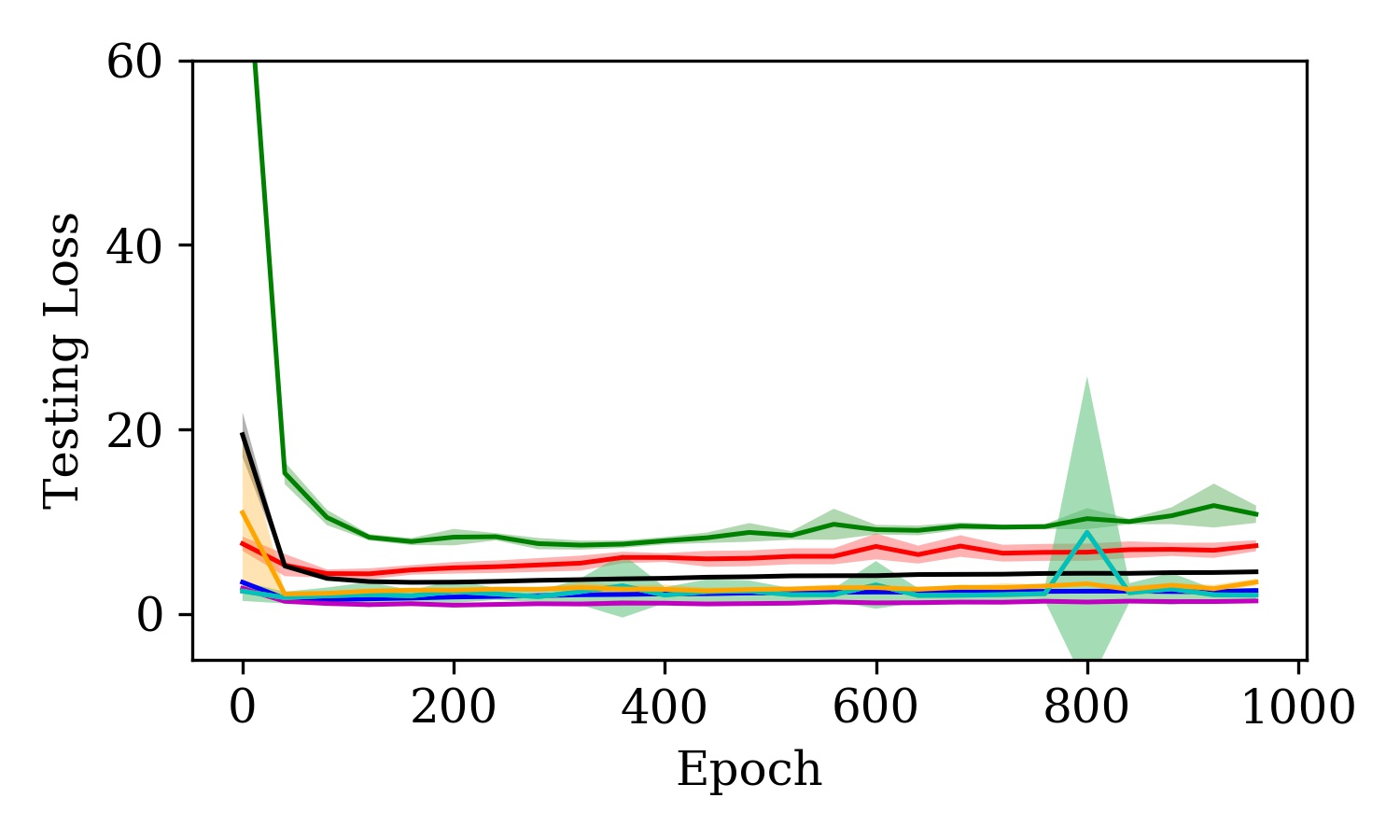}
        \vspace{-0.1cm}
    	\subcaption{Testing loss (WD).}
    	\label{image_training_2}
    \end{minipage}
    \hfill
	\begin{minipage}[b]{0.245\linewidth}
        \centering
    	\includegraphics[width=1.0\linewidth]{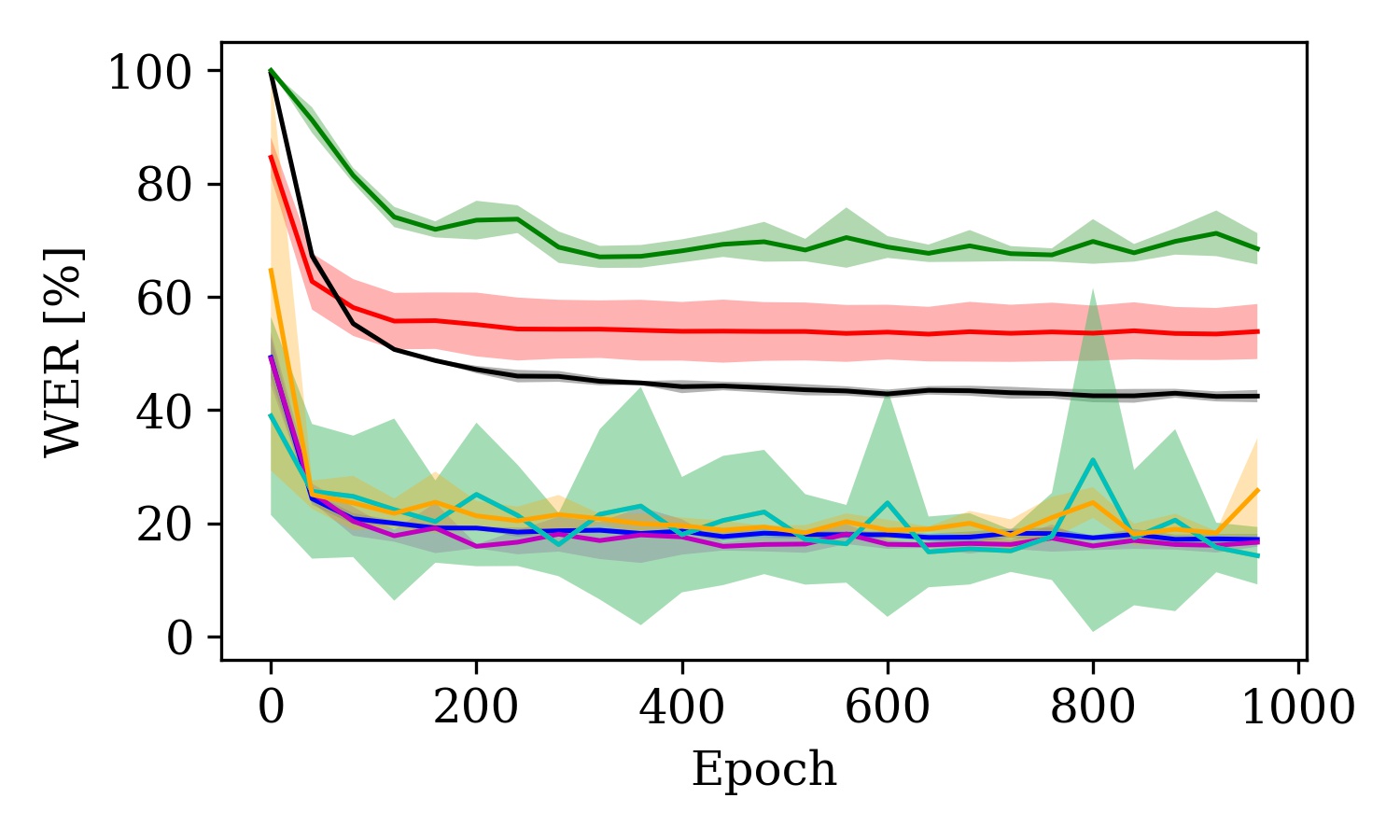}
        \vspace{-0.1cm}
    	\subcaption{WER (WD).}
    	\label{image_training_3}
    \end{minipage}
    \hfill
	\begin{minipage}[b]{0.245\linewidth}
        \centering
    	\includegraphics[width=1.0\linewidth]{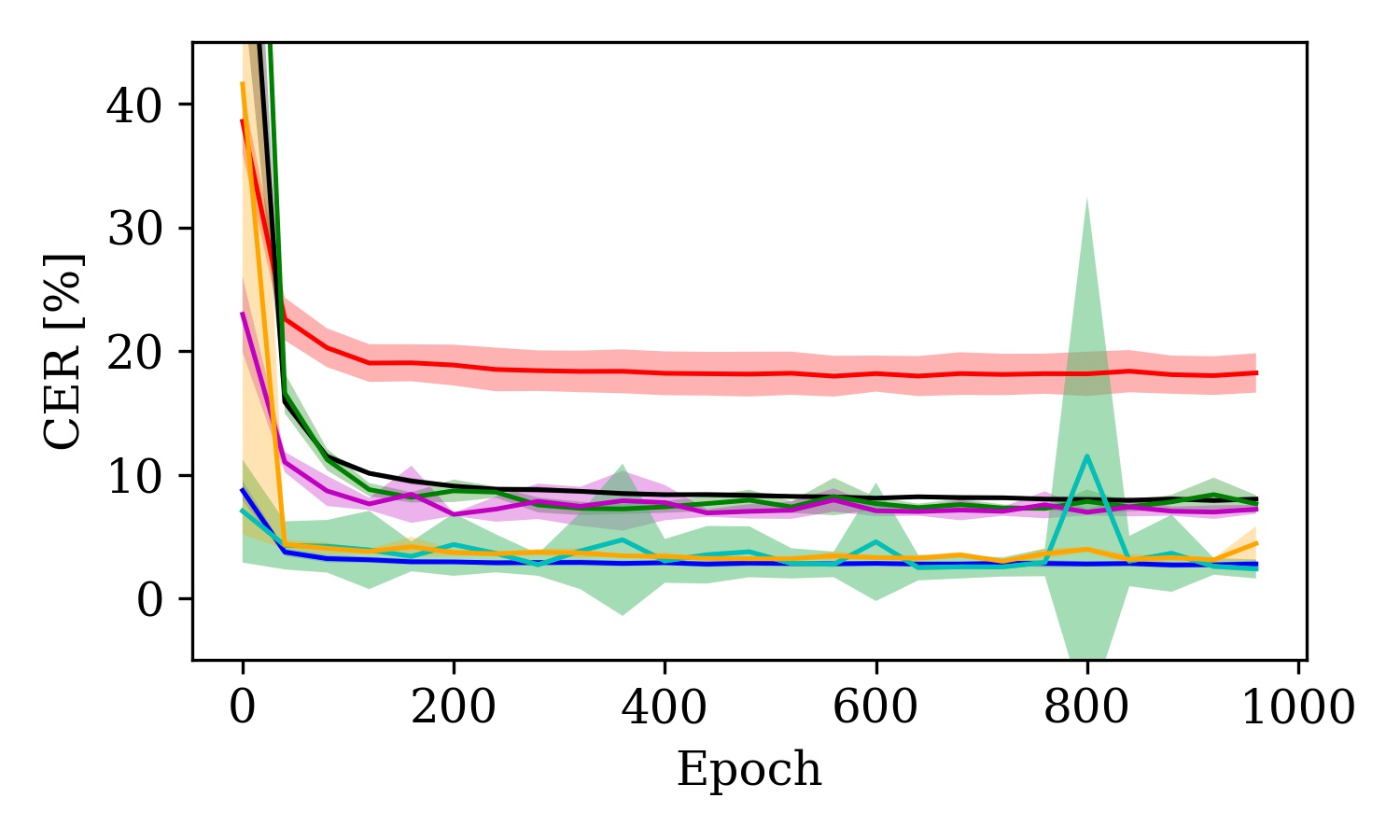}
        \vspace{-0.1cm}
    	\subcaption{CER (WD).}
    	\label{image_training_4}
    \end{minipage}
	\begin{minipage}[b]{0.245\linewidth}
        \centering
    	\includegraphics[width=1.0\linewidth]{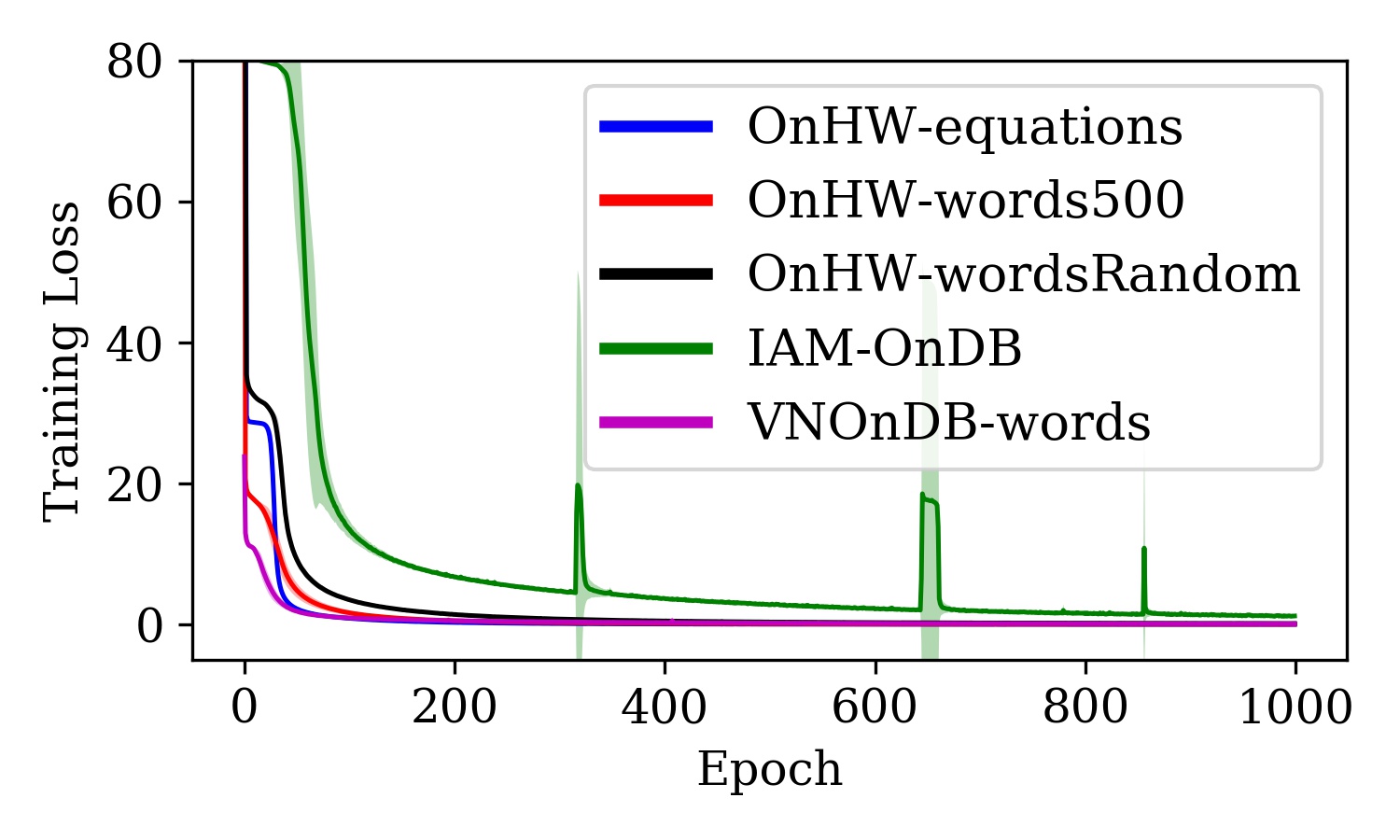}
        \vspace{-0.1cm}
    	\subcaption{Training loss (WI).}
    	\label{image_training_5}
    \end{minipage}
    \hfill
	\begin{minipage}[b]{0.245\linewidth}
        \centering
    	\includegraphics[width=1.0\linewidth]{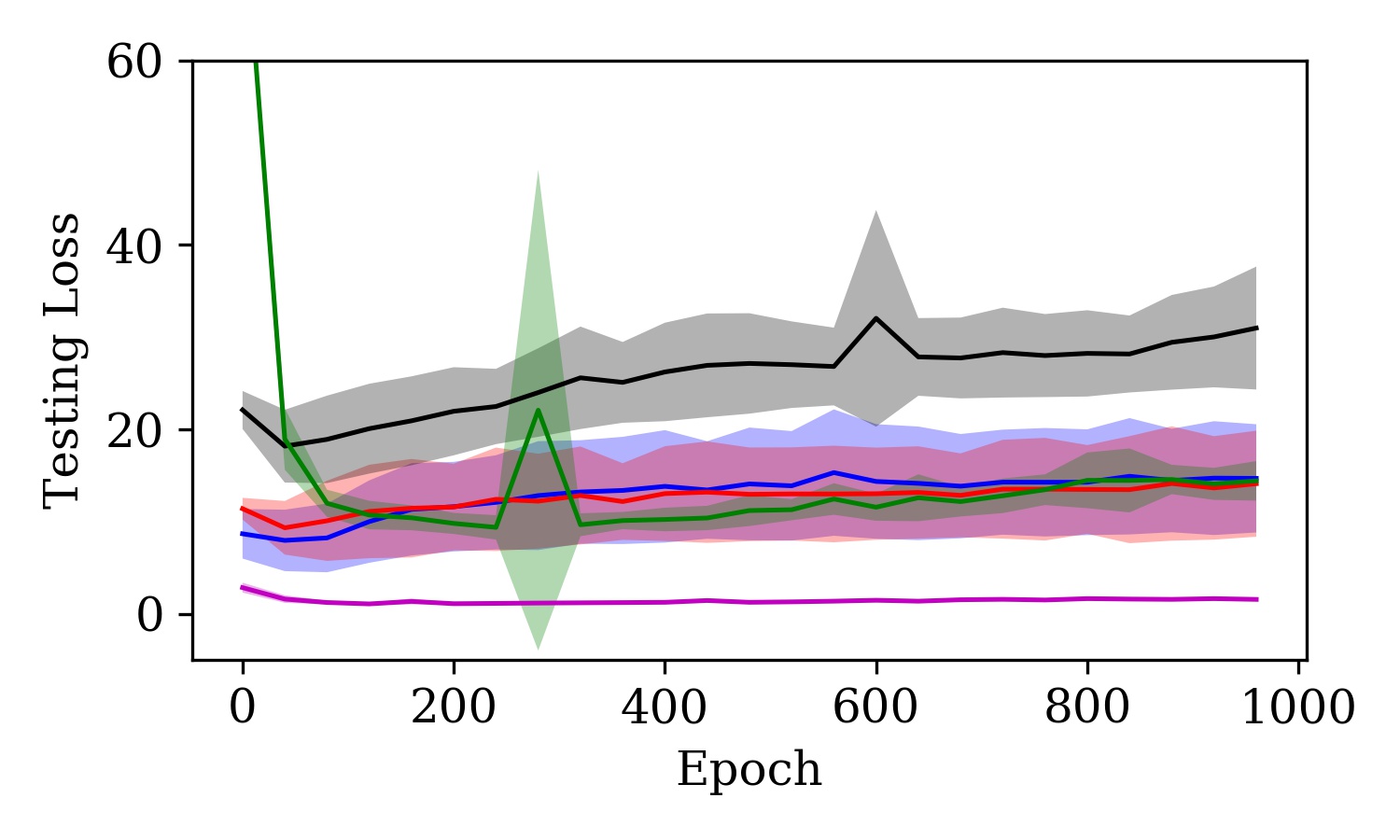}
        \vspace{-0.1cm}
    	\subcaption{Testing loss (WI).}
    	\label{image_training_6}
    \end{minipage}
    \hfill
	\begin{minipage}[b]{0.245\linewidth}
        \centering
    	\includegraphics[width=1.0\linewidth]{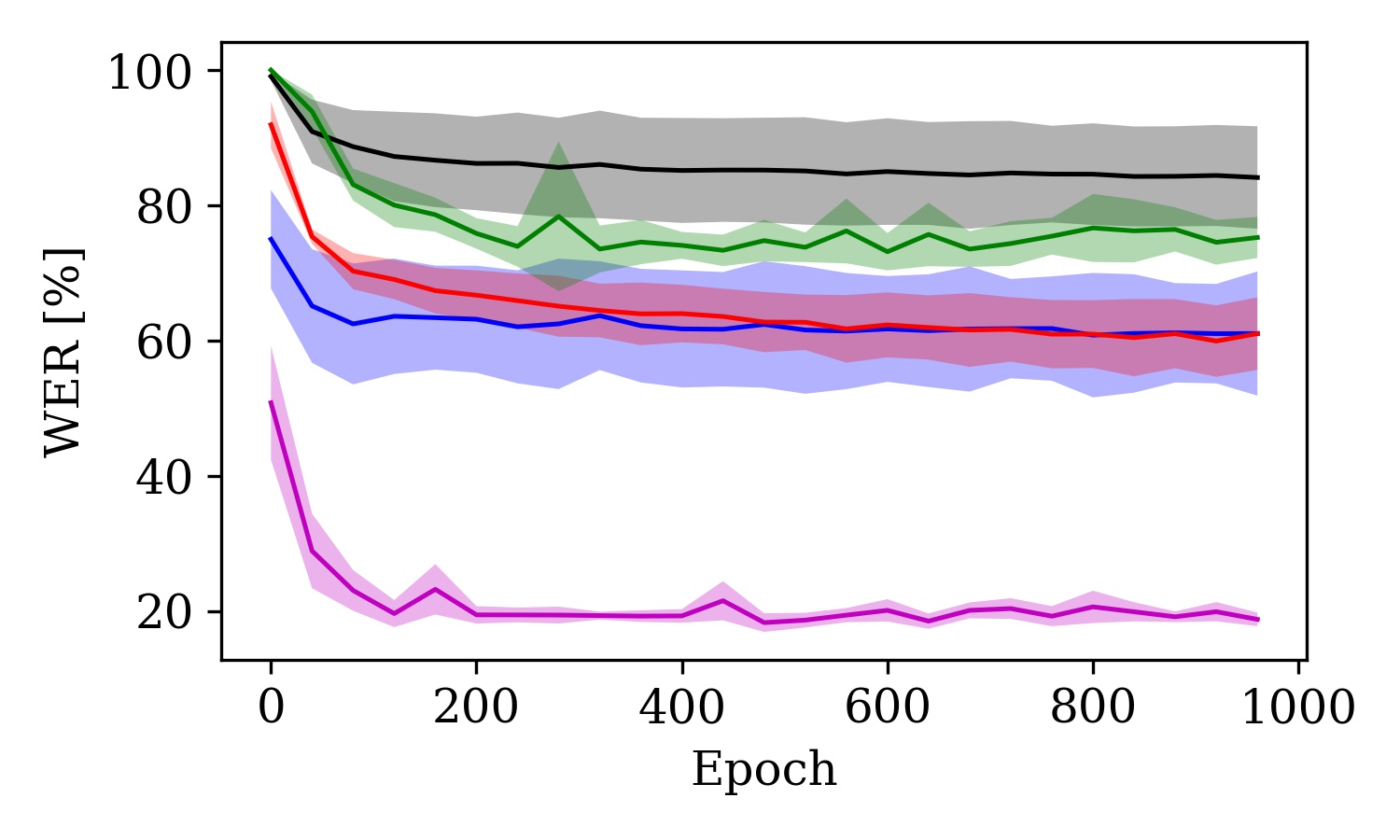}
        \vspace{-0.1cm}
    	\subcaption{WER (WI).}
    	\label{image_training_7}
    \end{minipage}
    \hfill
	\begin{minipage}[b]{0.245\linewidth}
        \centering
    	\includegraphics[width=1.0\linewidth]{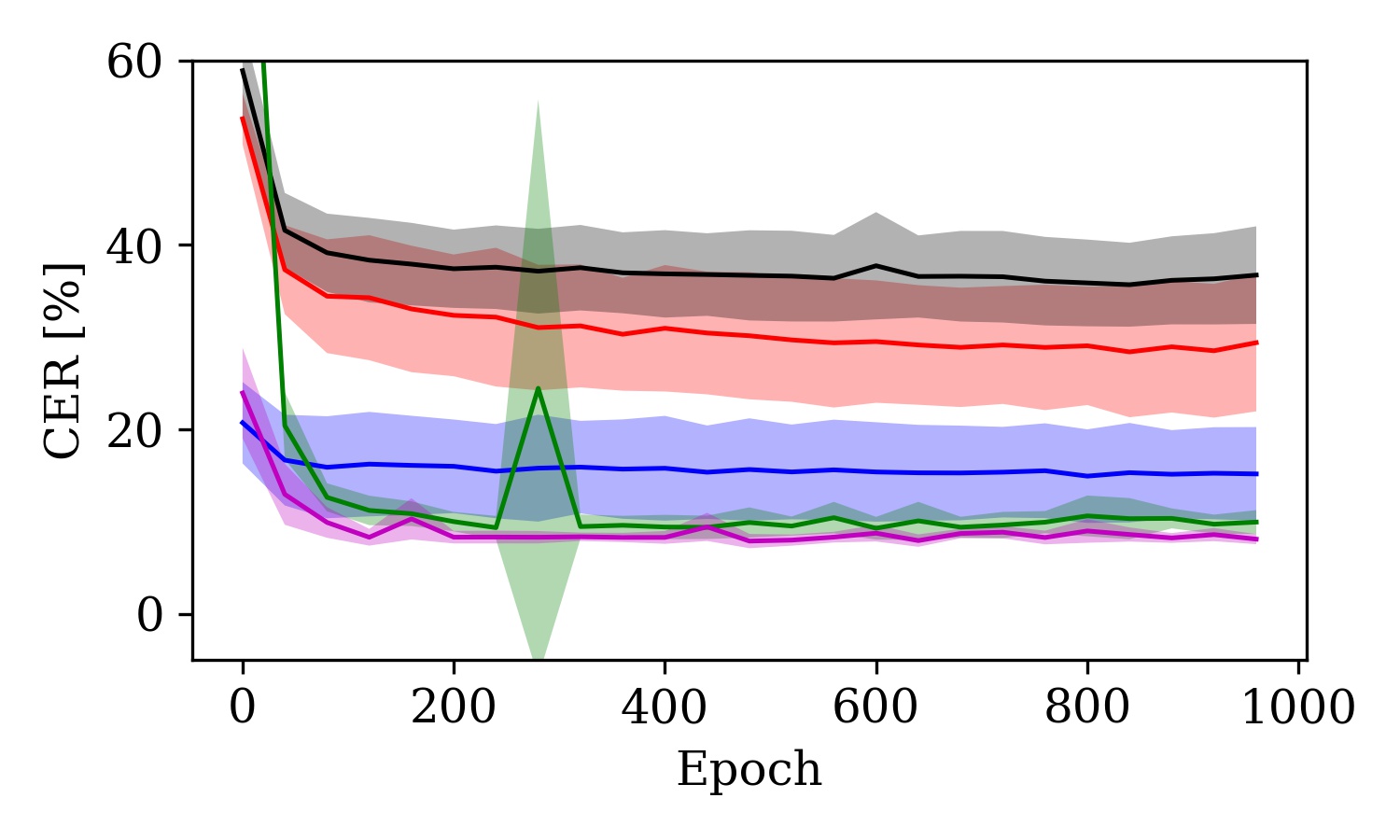}
        \vspace{-0.1cm}
    	\subcaption{CER (WI).}
    	\label{image_training_8}
    \end{minipage}
    \vspace{-0.1cm}
    \caption{Overview of training and testing losses and the evaluation metrics WER and CER in \% (mean and standard deviation over 5-fold cross validation) for all WD and WI datasets.}
    \label{image_training_overview}
\end{figure*}

\begin{figure*}[t!]
	\centering
	\begin{minipage}[b]{0.245\linewidth}
        \centering
    	\includegraphics[width=1.0\linewidth]{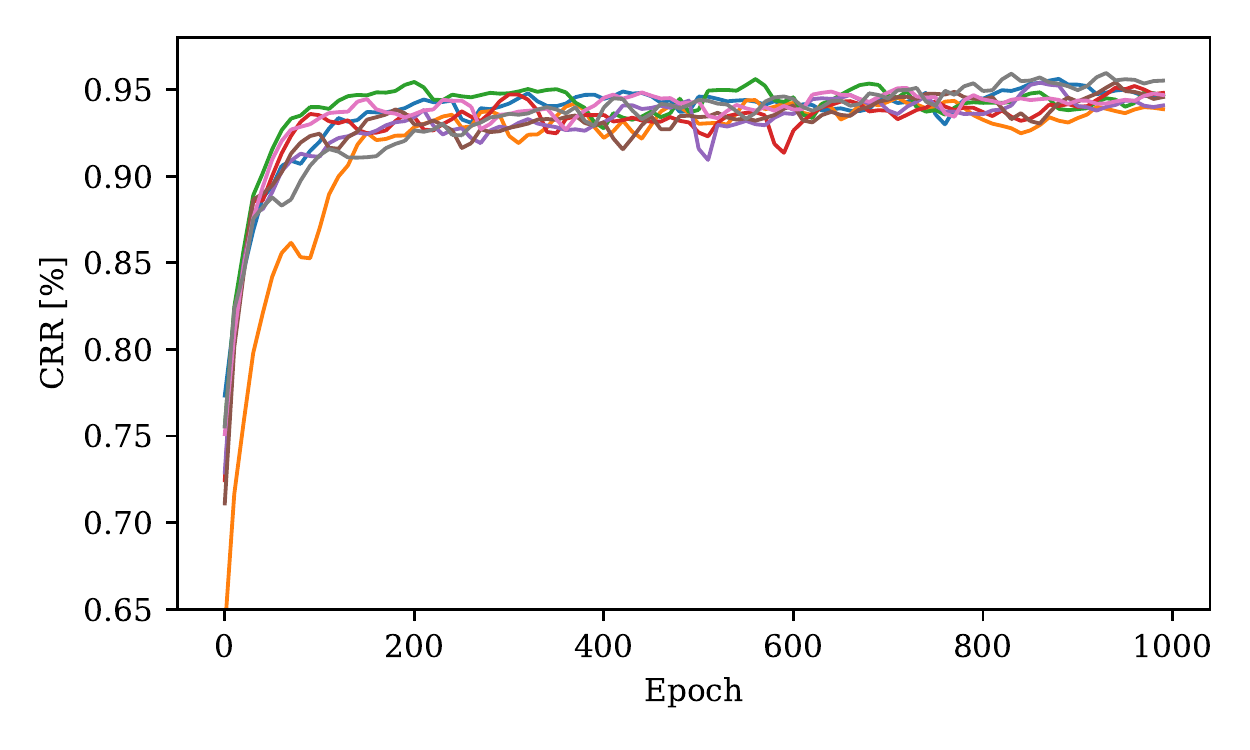}
        \vspace{-0.1cm}
    	\subcaption{OnHW-symbols: WD.}
    	\label{image_chars_train1}
    \end{minipage}
    \hfill
	\begin{minipage}[b]{0.245\linewidth}
        \centering
    	\includegraphics[width=1.0\linewidth]{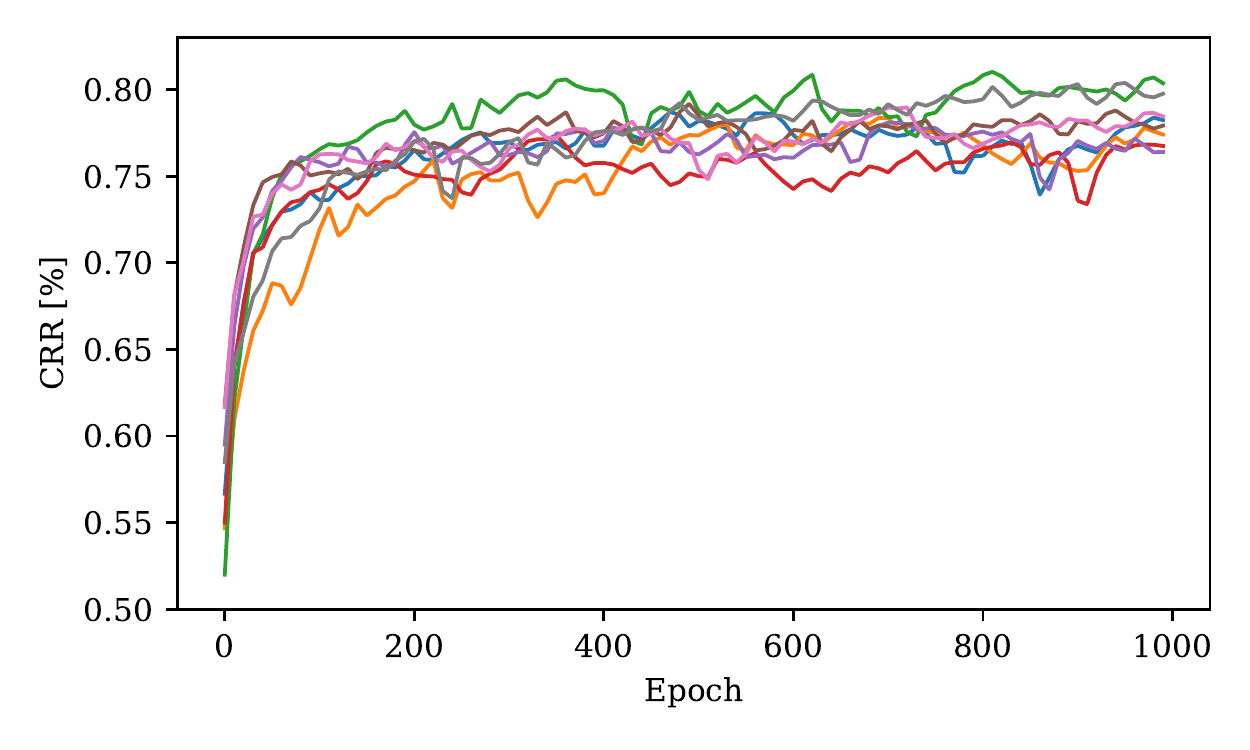}
        \vspace{-0.1cm}
    	\subcaption{OnHW-symbols: WI.}
    	\label{image_chars_train2}
    \end{minipage}
    \hfill
	\begin{minipage}[b]{0.245\linewidth}
        \centering
    	\includegraphics[width=1.0\linewidth]{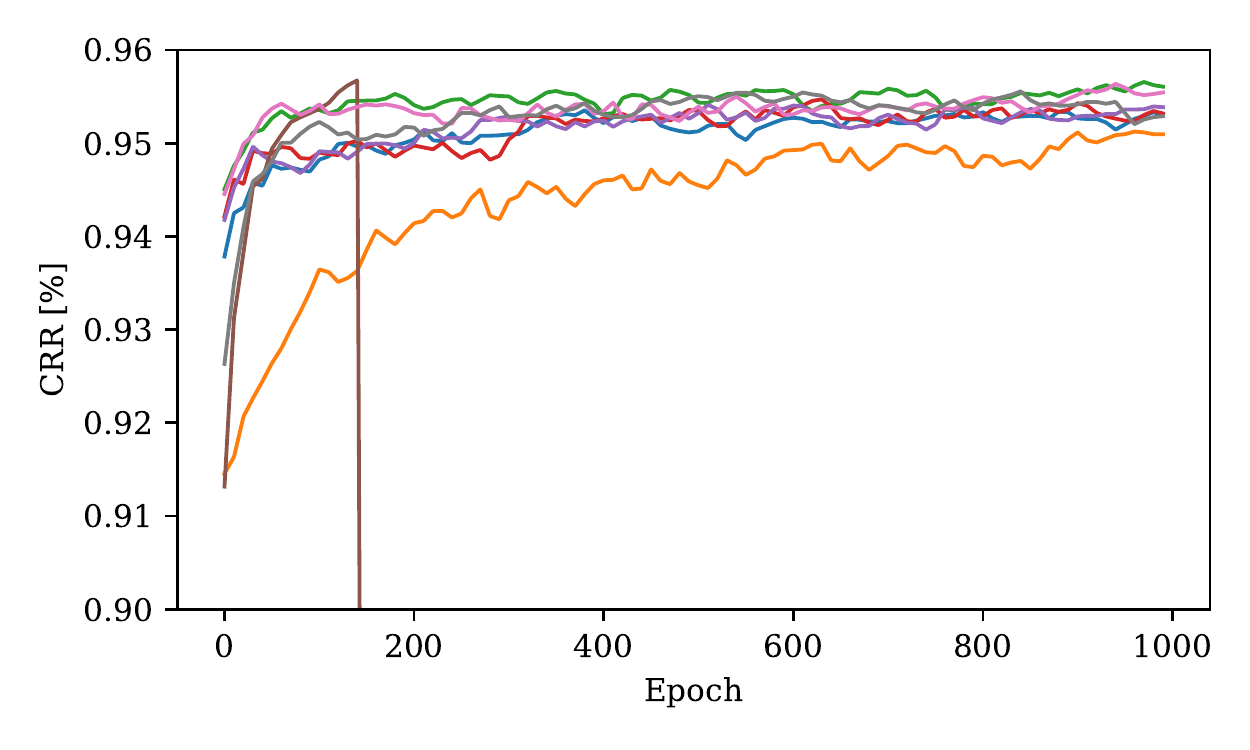}
        \vspace{-0.1cm}
    	\subcaption{OnHW-equations: WD.}
    	\label{image_chars_train3}
    \end{minipage}
    \hfill
	\begin{minipage}[b]{0.245\linewidth}
        \centering
    	\includegraphics[width=1.0\linewidth]{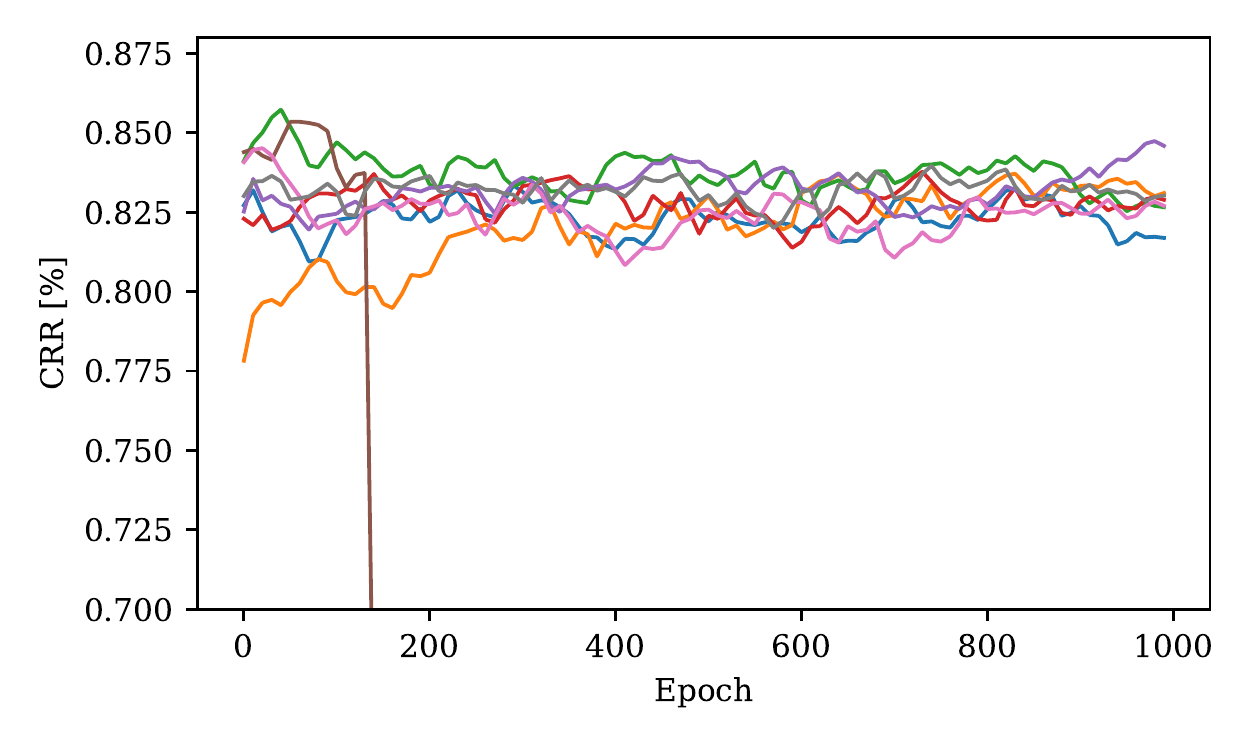}
        \vspace{-0.1cm}
    	\subcaption{OnHW-equations: WI.}
    	\label{image_chars_train4}
    \end{minipage}
	\begin{minipage}[b]{0.329\linewidth}
        \centering
    	\includegraphics[width=1.0\linewidth]{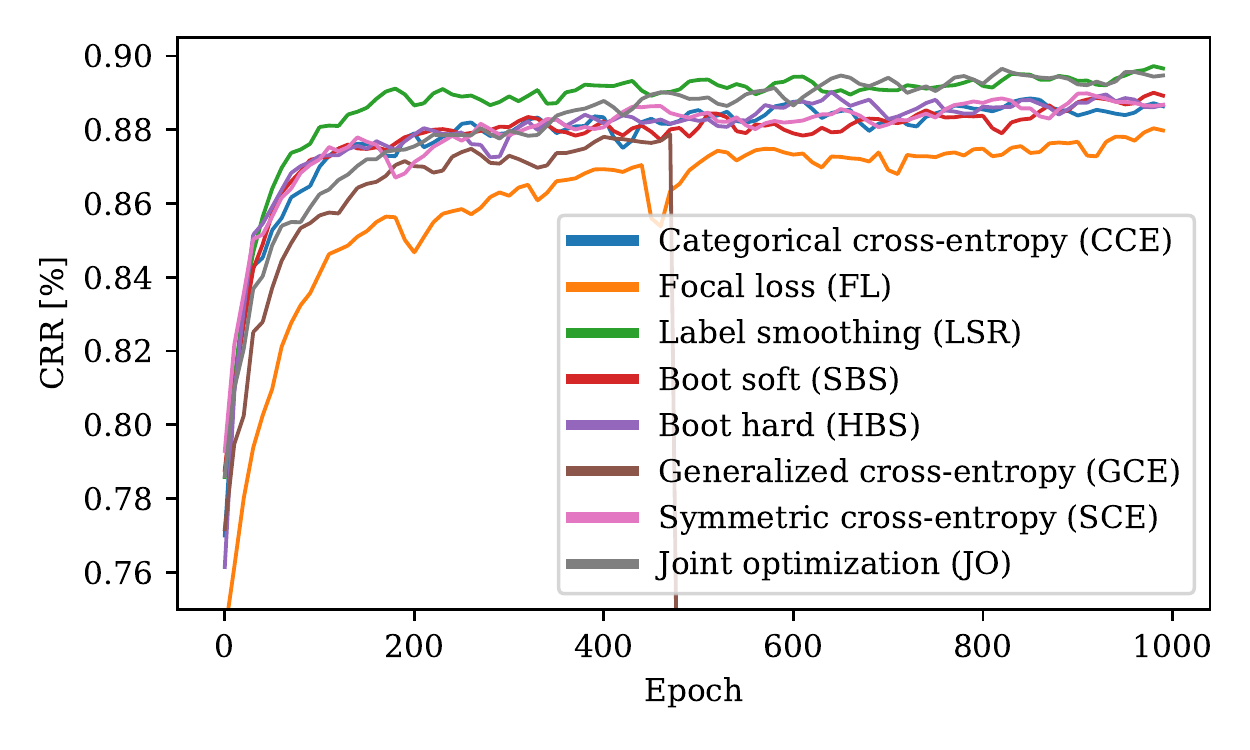}
        \vspace{-0.1cm}
    	\subcaption{OnHW-chars lower: WD.}
    	\label{image_chars_train5}
    \end{minipage}
    \hfill
	\begin{minipage}[b]{0.329\linewidth}
        \centering
    	\includegraphics[width=1.0\linewidth]{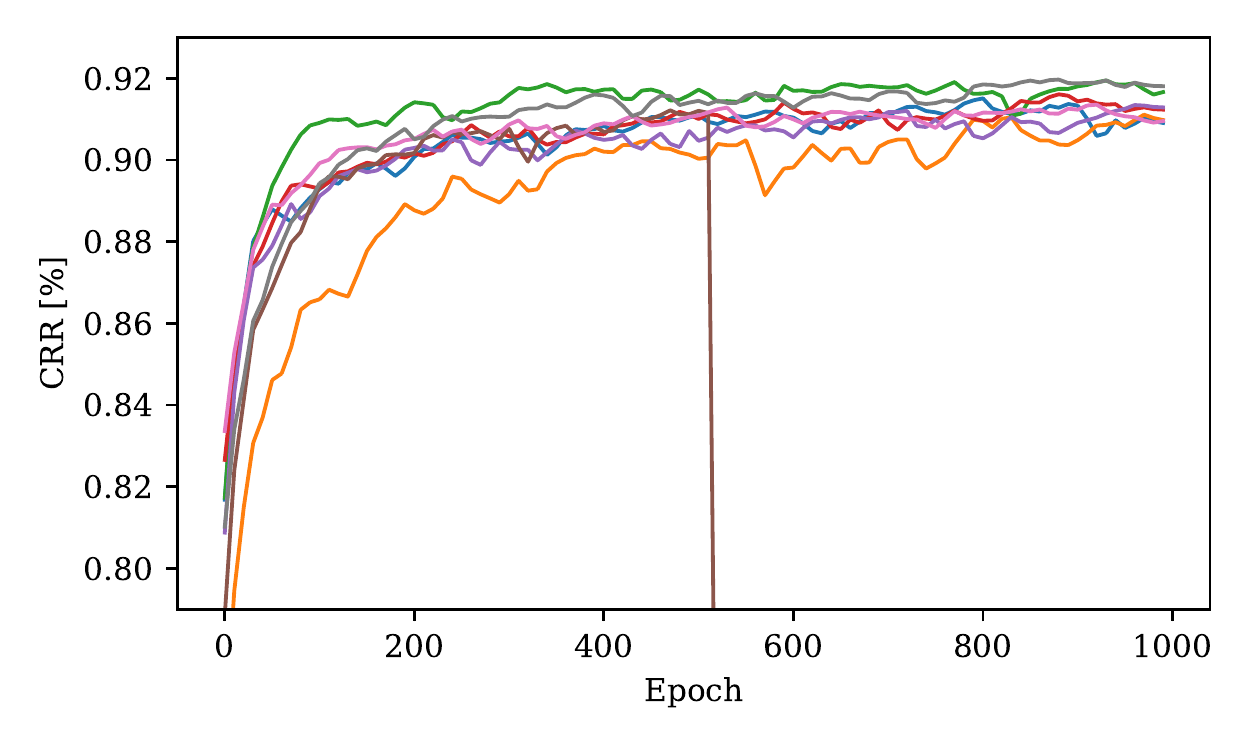}
        \vspace{-0.1cm}
    	\subcaption{OnHW-chars upper: WD.}
    	\label{image_chars_train6}
    \end{minipage}
    \hfill
	\begin{minipage}[b]{0.329\linewidth}
        \centering
    	\includegraphics[width=1.0\linewidth]{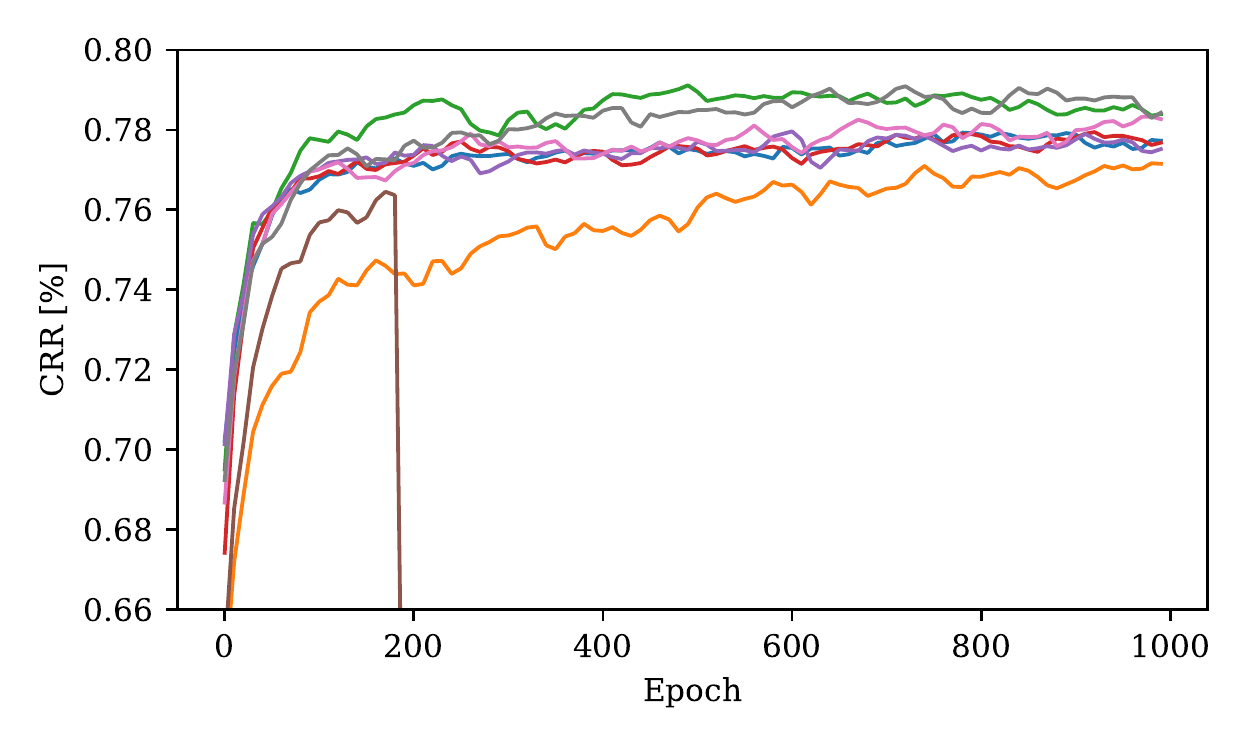}
        \vspace{-0.1cm}
    	\subcaption{OnHW-chars combined: WD.}
    	\label{image_chars_train7}
    \end{minipage}
	\begin{minipage}[b]{0.329\linewidth}
        \centering
    	\includegraphics[width=1.0\linewidth]{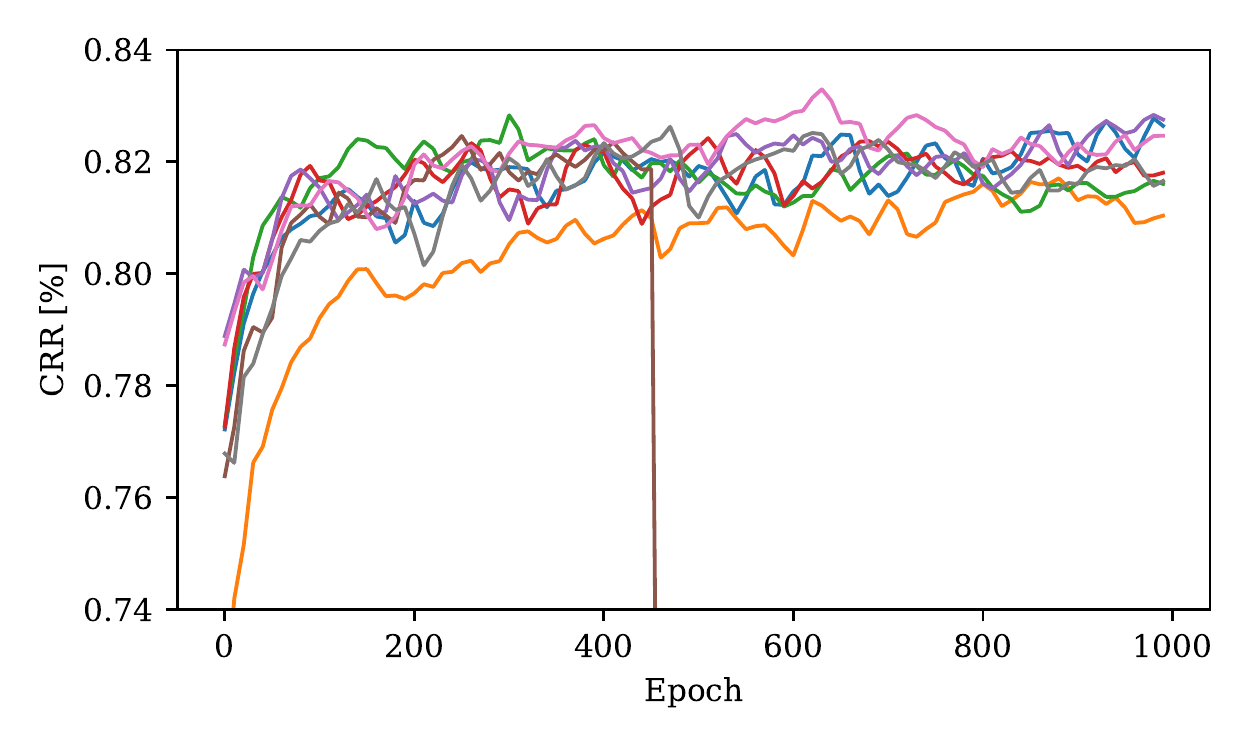}
        \vspace{-0.1cm}
    	\subcaption{OnHW-chars lower: WI.}
    	\label{image_chars_train8}
    \end{minipage}
    \hfill
	\begin{minipage}[b]{0.329\linewidth}
        \centering
    	\includegraphics[width=1.0\linewidth]{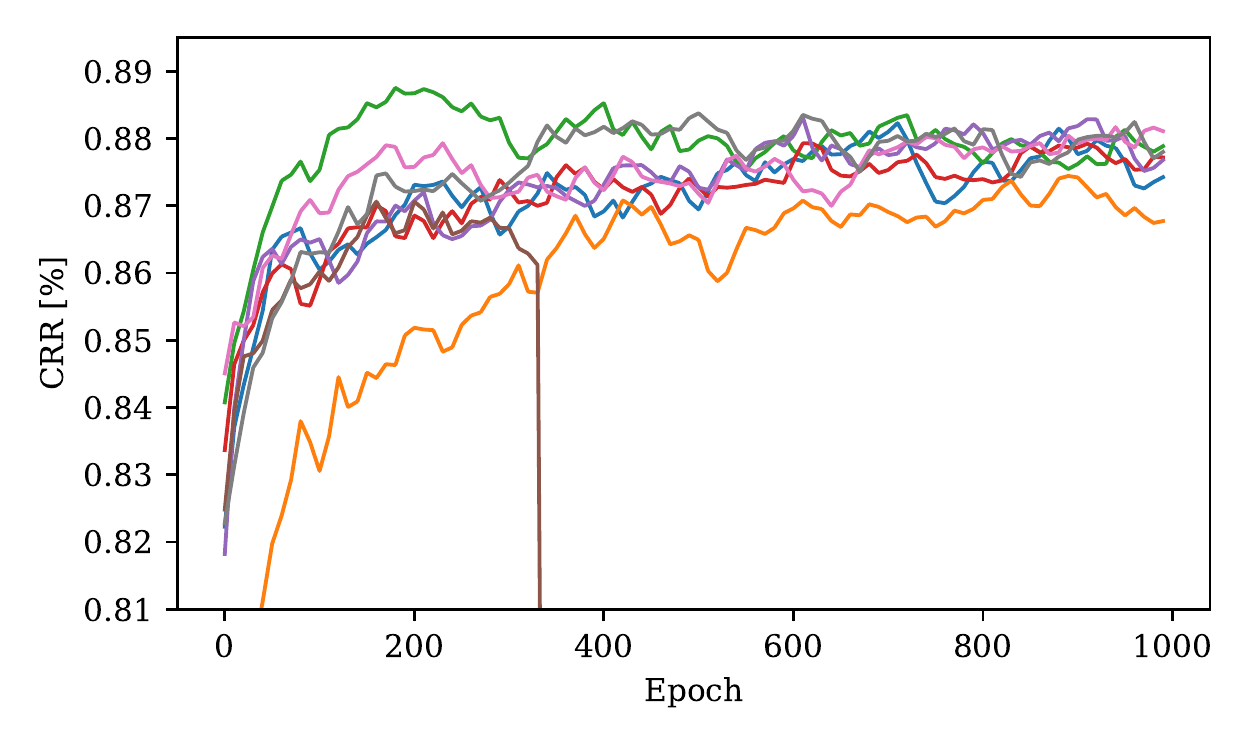}
        \vspace{-0.1cm}
    	\subcaption{OnHW-chars upper: WI.}
    	\label{image_chars_train9}
    \end{minipage}
    \hfill
	\begin{minipage}[b]{0.329\linewidth}
        \centering
    	\includegraphics[width=1.0\linewidth]{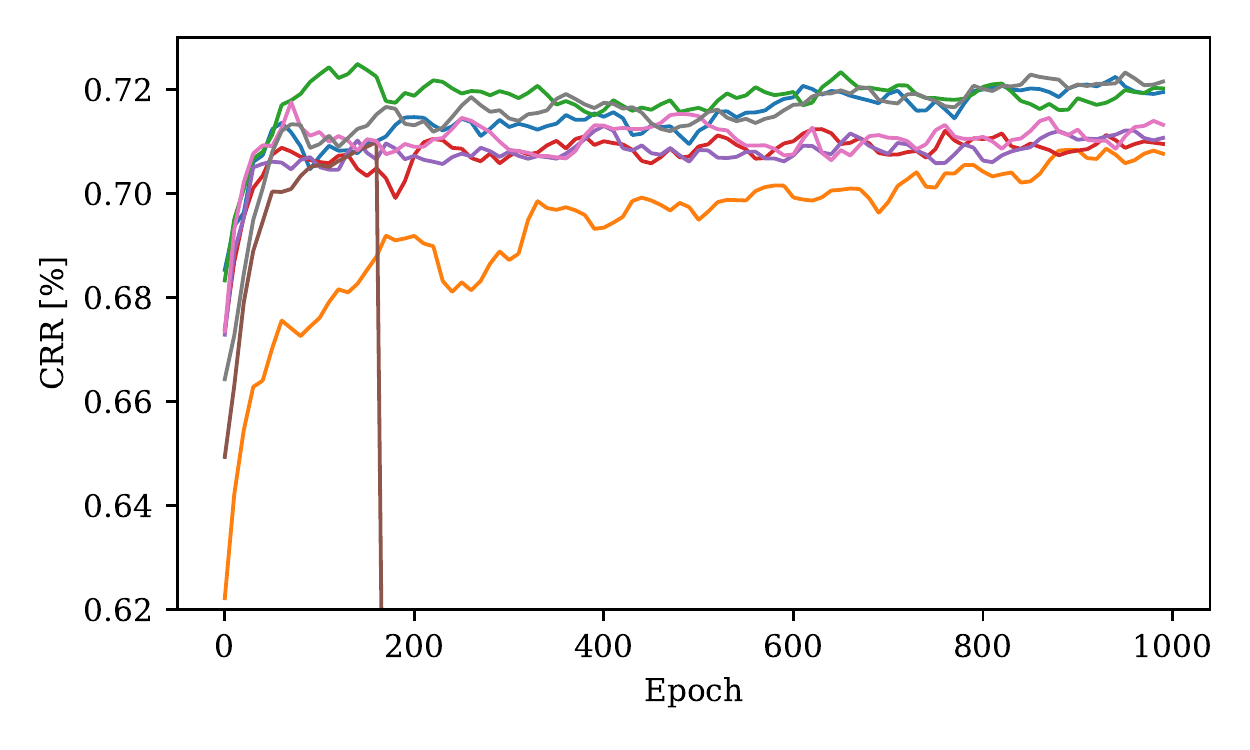}
        \vspace{-0.1cm}
    	\subcaption{OnHW-chars combined: WI.}
    	\label{image_chars_train10}
    \end{minipage}
    \vspace{-0.1cm}
    \caption{Overview of validation accuracies evaluated every $10^{th}$ training epoch for the WD and WI OnHW-symbols, split OnHW-equations and OnHW-chars~\cite{ott} datasets.}
    \label{image_training_single_chars}
\end{figure*}

\paragraph{Error and Accuracy Plots.} Figure~\ref{image_training_overview} shows an overview of error plots for all sequence-based datasets. While the training losses converges very fast (see Figure~\ref{image_training_1} and \ref{image_training_5}) and the models slightly overfit (see Figure~\ref{image_training_2} and \ref{image_training_6}), the WER (see Figure~\ref{image_training_3} and \ref{image_training_7}) and the CER (see Figure~\ref{image_training_4} and \ref{image_training_8}) are continuously decreasing. We propose the validation accuracies (CRR) while training in Figure~\ref{image_training_single_chars} for single-based datasets for the eight training losses. The generalized cross entropy (GCE) is often not robust, see Figure~\ref{image_chars_train3} to \ref{image_chars_train10}. The difference between loss function of the WD symbols and equations datasets is small (see Figure~\ref{image_chars_train1} and \ref{image_chars_train3}), but gets more important for the WI tasks (see Figure~\ref{image_chars_train2} and \ref{image_chars_train4}). From the OnHW-chars~\cite{ott} dataset, we can conclude that symmetric cross entropy (SCE), label smoothing (LSR) and joint optimization (JO) can improve the baseline categorical cross entropy (CCE) loss. The Focal loss (FL)~\cite{lin_goyal} converges slower, and boot soft (SBS) and boot hard (HBS) is similar to CCE.

\paragraph{Training Times.} Table~\ref{table_training_times} compares training times of all methods used for our benchmark for the lower OnHW-chars~\cite{ott} (WD) and our OnHW-equations (WD) datasets. For all trainings, we used Nvidia Tesla V100-SXM2 GPUs with 32 GB VRAM. TapNet~\cite{tapnet} has the fastest training time of $3.6s$, but we trained 3,000 epochs for convergence. While we train our CNN+LSTM model for 1,000 epochs with $8.0s$ each, the MLSTM-FCN~\cite{mlstm_fcn} trains slower, but converges faster (only 200 epochs). The training times per epoch of the Transformer variants~\cite{choromanski,jaegle,kitaev,tay_bahri,wang_li} are significantly higher (between $15.7s$ to $24.5s$), but the convergence is also significantly faster of less than 100 epochs. The Linformer~\cite{wang_li} ($8.8s$) is as fast as our CNN+LSTM model ($8.0s$). For seq2seq classification tasks, our attention-based model is the fastest with $27.5s$. The CNN+TCN model requires $43.5s$ and the CNN+LSTM model $62s$, and can emphasize the advantage of attention-based models. The CNN+BiLSTM model achieves the lowest error rates, but trains clearly slower with $131s$. In conclusion, Transformers train faster than classical methods, but our classical CNN and RNN models achieve the highest accuracies. Modules trained with the tsai toolbox have lower training times: InceptionTime ($2.0s$), XceptionTime ($3.8s$), ResCNN ($2.2s$) and ResNet ($2.9s$). The small model FCN is very fast at training ($1.5s$) that increases for added temporal units such as LSTM-FCN ($6.8s$) and MLSTM-FCN ($7.4s$). The training of InceptionTime increases from $2.0s$ for \textit{depth} 3 and \textit{nf} 16 up to $37.6s$ for \textit{depth} 12 and \textit{nf} 128. Added BiLSTM layers up to double the training times.

\begin{table}
\begin{center}
    \setlength{\tabcolsep}{3.2pt}
    \caption{Comparison of training times per epoch in seconds (\textit{s}).}
    \label{table_training_times}
    \small \begin{tabular}{ p{0.5cm} | p{0.5cm} | p{0.5cm} }
    \multicolumn{1}{c|}{\textbf{Method}} & \multicolumn{1}{c|}{\textbf{OnHW-}} & \multicolumn{1}{c}{\textbf{OnHW-}} \\
    & \multicolumn{1}{c|}{\textbf{chars}} & \multicolumn{1}{c}{\textbf{equations}} \\ \hline
    \multicolumn{1}{l|}{CNN+LSTM} & \multicolumn{1}{r|}{8.0} & \multicolumn{1}{r}{62} \\
    \multicolumn{1}{l|}{CNN+BiLSTM} & \multicolumn{1}{r|}{19.7} & \multicolumn{1}{r}{131} \\
    \multicolumn{1}{l|}{CNN+TCN} & \multicolumn{1}{r|}{7.3} & \multicolumn{1}{r}{43.5} \\
    \multicolumn{1}{l|}{Attention-based model} & \multicolumn{1}{c|}{-} & \multicolumn{1}{r}{27.5} \\
    \multicolumn{1}{l|}{Perceiver~\cite{jaegle}} & \multicolumn{1}{r|}{17.1} & \multicolumn{1}{c}{-} \\
    \multicolumn{1}{l|}{Sinkhorn~\cite{tay_bahri}} & \multicolumn{1}{r|}{16.1} & \multicolumn{1}{c}{-} \\
    \multicolumn{1}{l|}{Performer~\cite{choromanski}} & \multicolumn{1}{r|}{15.7} & \multicolumn{1}{c}{-} \\
    \multicolumn{1}{l|}{Reformer~\cite{kitaev}} & \multicolumn{1}{r|}{24.5} & \multicolumn{1}{c}{-} \\
    \multicolumn{1}{l|}{Linformer~\cite{wang_li}} & \multicolumn{1}{r|}{8.8} & \multicolumn{1}{c}{-} \\
    \multicolumn{1}{l|}{TapNet~\cite{tapnet}} & \multicolumn{1}{r|}{3.6} & \multicolumn{1}{c}{-} \\
    \multicolumn{1}{l|}{MLSTM-FCN~\cite{mlstm_fcn}} & \multicolumn{1}{r|}{12.0} & \multicolumn{1}{c}{-} \\
    \end{tabular}
\end{center}
\end{table}

\end{document}